\documentclass[11pt]{article}

\usepackage[final]{acl}

\usepackage{times}
\usepackage{latexsym}

\usepackage[T1]{fontenc}

\usepackage[utf8]{inputenc}

\usepackage{microtype}

\usepackage{inconsolata}

\usepackage{graphicx}

\usepackage{hyperref}
\usepackage{url}
\usepackage{xurl} 
\usepackage[utf8]{inputenc} 
\usepackage[T1]{fontenc}    
\usepackage{hyperref}       
\usepackage{url}            
\usepackage{booktabs}       
\usepackage{amssymb}
\usepackage{pifont}
\newcommand{\cmark}{\ding{51}}%
\newcommand{\xmark}{\ding{55}}%

\usepackage{subcaption}
\usepackage{siunitx} 

\usepackage{tocbasic}

\usepackage{microtype}      
\usepackage{xcolor}         
\usepackage{tikz}
\usepackage{amsmath}
\usepackage{fontawesome5}
\usepackage{graphicx}
\usepackage{paralist}
\usepackage{caption}
\usepackage{natbib}
\usepackage{listings}
\usepackage[table]{xcolor} 
\usepackage{wrapfig}  
\usepackage{colortbl} 
\usepackage{float}
\usepackage{xcolor}
\usepackage{colortbl}
\usepackage{multicol}
\usepackage{tabularx}
\usepackage{algorithm}
\usepackage{algorithmic}
\usepackage{multirow}
\usepackage{adjustbox}
\usepackage{enumitem}
\usepackage[dvipsnames]{xcolor}
\usepackage{comment}
\usepackage{enumitem}
\usepackage{subcaption}
\usepackage{makecell}
\usepackage{ragged2e}

\usepackage[toc,page,header]{appendix}
\usepackage{minitoc}

\usetikzlibrary{shapes.geometric, arrows.meta, positioning}

\hypersetup{
    colorlinks=true,
    citecolor=blue,
    linkcolor=black,
    urlcolor=black,
    pdfborder={0 0 0}
}

\usepackage[most]{tcolorbox}

\definecolor{darkblue}{rgb}{0, 0, 0.5}
\definecolor{dark-green}{HTML}{006400}
\hypersetup{colorlinks=true, citecolor=darkblue, linkcolor=darkblue, urlcolor=darkblue}

\title{CaTS-Bench: Can Language Models Describe Time Series?}

\author{
\textbf{Luca Zhou\footnotemark[1]\textsuperscript{\dag\ddag}},
\textbf{Pratham Yashwante\footnotemark[1]\textsuperscript{\ddag}},
 \textbf{Marshall Fisher\textsuperscript{\ddag}},
 \textbf{Alessio Sampieri\textsuperscript{\S}},
\\
 \textbf{Zihao Zhou\textsuperscript{\ddag}},
 \textbf{Fabio Galasso\textsuperscript{\dag}},
 \textbf{Rose Yu\textsuperscript{\ddag}}
\\
\\
 \textsuperscript{\dag}Sapienza University of Rome,
 \textsuperscript{\ddag}UC San Diego,
 \textsuperscript{\S}ItalAI Labs
\\
 \small{
   \textbf{Correspondence:} \href{mailto:luca.zhou@uniroma1.it}{luca.zhou@uniroma1.it}
 }
}

\begin{document}
\maketitle

\renewcommand{\thefootnote}{\fnsymbol{footnote}}
\footnotetext[1]{Equal contribution.}
\renewcommand{\thefootnote}{\arabic{footnote}}

\begin{abstract}
Time series captioning, the task of describing time series in natural language, requires numeric and temporal reasoning, trend interpretation, and contextual understanding. Existing benchmarks, however, often rely on fully synthetic or generic captions, and typically neglect metadata and visual representations. We introduce \textbf{CaTS-Bench}, a comprehensive benchmark for \textbf{C}ontext-\textbf{a}ware \textbf{T}ime \textbf{S}eries reasoning across $11$ diverse domains, centered on a gold-standard evaluation set of $1746$ human-rewritten captions that measure how effectively models translate numeric trends into immediately interpretable narratives. To address the scarcity of human-annotated data, we also propose a scalable pipeline for generating high-fidelity synthetic captions, the quality of which we validate. We evaluate leading Vision-Language Models on our benchmark, revealing that even proprietary models struggle to capture numeric nuances in temporal descriptions, while finetuning open-source models on synthetic data yields substantial performance gains. Finally, we release a diagnostic suite of $910$ multiple-choice questions and use tailored numeric metrics to gauge time-series-specific reasoning capabilities, establishing CaTS-Bench as a reliable foundation for grounded, multimodal text generation in numeric domains.

\end{abstract}

\begin{figure*}[t]
    \centering
    \includegraphics[width=\textwidth]{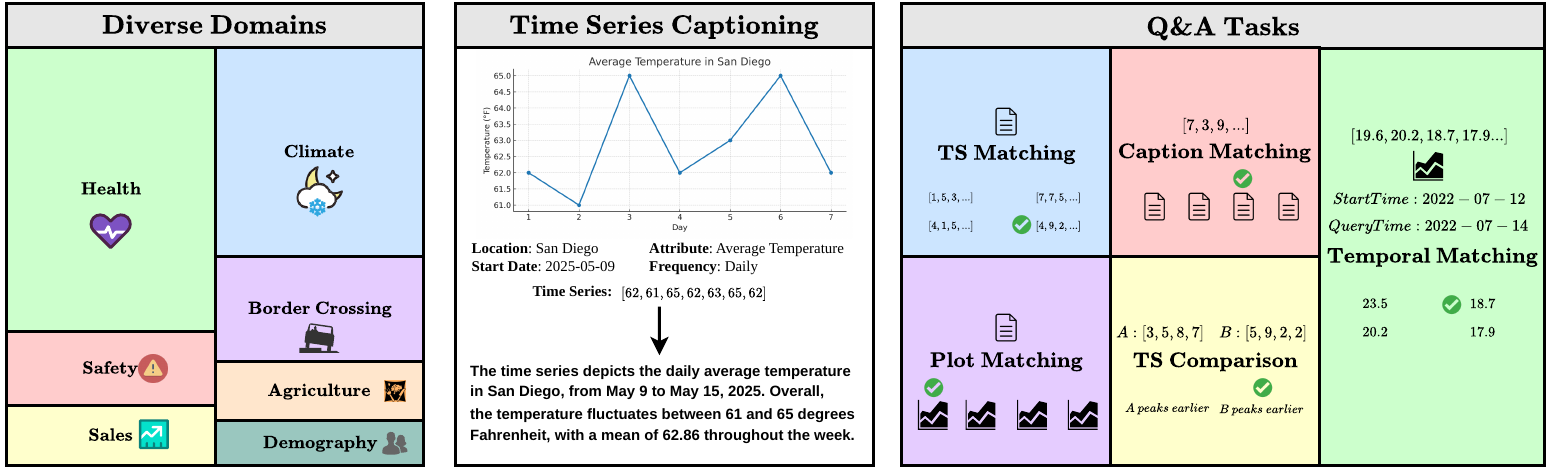}
    \caption{\textbf{Overview of CaTS-Bench.} It features $11$ domains organized into $7$ macro domains, provides human-rewritten and synthetic data, and formulates six challenging tasks, with time series captioning as the primary one.}
    \label{fig:teaser}
\end{figure*}

\section{Introduction}
Effective interpretation of time series data underpins decision-making in domains such as finance, healthcare, and climate analysis. Although raw numeric sequences are abundant, turning them into concise, human-readable summaries is labor-intensive and demands domain expertise and statistical literacy. In practice, analysts, policymakers, journalists, and practitioners rarely inspect raw sequences directly, instead relying on textual descriptions to quickly assess trends, anomalies, and changes before deciding whether deeper analysis is needed. Automating this process through \emph{time series captioning} (TSC) not only accelerates insight discovery but also democratizes access to complex temporal analytics, which enables non‑experts to ask natural‑language questions and receive explanations without writing code or inspecting raw charts. 

We further discuss the motivation and practical utility of TSC in Appendix~\ref{sec:motivation}, including: (i) its distinction from deterministic statistical analysis, (ii) its role in generating human-interpretable, context-aware narratives, and (iii) its use as a diagnostic task for evaluating free-form language generation.


Recent advances in Large Language Models (LLMs) and Vision-Language Models (VLMs) have transformed grounded language generation, yet these models show critical weaknesses in precise reasoning and factual grounding on numeric time series \citep{tang2025time, cao2024evaluation}. Progress in TSC is bottlenecked by the lack of high-quality, linguistically diverse datasets and benchmarks: existing resources are largely synthetic, template-based, or omit contextual metadata (e.g., units, domain-specific labels) that humans rely on, leading models to produce generic captions that miss the underlying narrative.

To address this, we introduce \textbf{CaTS-Bench}\footnote{The dataset is available at \url{https://huggingface.co/datasets/mhfisher/CaTSBench}}, a context-rich multimodal benchmark for \emph{context-aware} time series captioning. An overview of the benchmark is shown in Figure~\ref{fig:teaser}. Here, “context-aware” denotes synthesizing information from numeric series, visual plots, and contextual metadata. Rather than assuming that access to multiple modalities implies effective multimodal reasoning, we explicitly probe whether VLMs properly use these signals. Unlike previous efforts that rely solely on simplistic or synthetic data, our work centers on a new \textbf{gold-standard evaluation set} of $1746$ human-rewritten (HR) captions. These captions provide a rigorous baseline to measure linguistic richness, factual and numeric precision, and human-like reasoning in temporal descriptions.

Recognizing the prohibitive cost of large-scale human annotation, another key contribution of this work is a \textbf{scalable data augmentation pipeline} that uses an oracle LLM, grounded in factual metadata and validated via human verification studies, to generate high-fidelity synthetic training captions.  Our experiments show that zero-shot VLMs struggle with the complexity of HR descriptions, whereas finetuning on our synthetic data yields substantial gains, particularly in semantic and numeric metrics on the HR set, positioning the pipeline as an effective solution to data scarcity for TSC.

Beyond captioning, CaTS-Bench provides a diagnostic suite of $910$ multiple-choice questions (MCQs) targeting numeric precision and multimodal alignment, together with tailored evaluation metrics that prioritize numeric fidelity. Our results reveal that current VLMs often fail to exploit visual cues for TSC and struggle even on simple diagnostic MCQ tasks, such as correctly mapping values to time stamps and comparing basic statistical properties, highlighting persistent gaps in grounded multimodal reasoning. Overall, our contributions are:

\begin{enumerate}[leftmargin=*, itemsep=2pt, parsep=0pt, topsep=2pt] 

    \item \textbf{CaTS-Bench}: A multimodal, 11-domain benchmark featuring a 1746-sample human-rewritten evaluation set for TSC and $910$ MCQs designed to rigorously measure linguistically rich and factually grounded free-form text generation and temporal reasoning.

    \item \textbf{Scalable Augmentation Pipeline}: A framework for generating high-fidelity synthetic data. We empirically demonstrate that finetuning on this synthetic data improves model performance on our human-rewritten test set.
    
    \item \textbf{Diagnostic Analysis \& Metrics}: A suite of tailored numeric fidelity metrics that reveal systematic failures in visual grounding, temporal alignment, and multimodal reasoning.
    
\end{enumerate}

\section{Related Work} 

\begin{table*}[t]
\centering
\setlength{\tabcolsep}{3pt}
\renewcommand{\arraystretch}{1.1}
\caption{Comparison of TSC benchmarks.}
\resizebox{\textwidth}{!}{
\begin{tabular}{l c c c c c c}
\toprule
\rowcolor[gray]{0.95} \textbf{Dataset} & \textbf{Data Nature} & \textbf{Modalities} & \textbf{Domains} & \textbf{Context} & \textbf{Captions} & \textbf{Q\&A} \\ 
\midrule
TaxoSynth \citep{fons2024evaluating} & Synthetic & Numeric + Text & N/A & Minimal & Simplistic & \xmark \\
TADACap \citep{fons2024tadacap} & Partially Synthetic & Visual & 4 & Minimal & Simplistic  & \xmark \\
TRUCE \citep{jhamtani2021truth} & Partially Synthetic & Numeric & 2 & \xmark & Simplistic & \xmark \\
TACO \citep{dohi2025domain} & Mostly Synthetic & Numeric & 8 & \xmark & Expressive & \xmark \\
BEDTime \cite{sen2025bedtimeunifiedbenchmarkautomatically} & Partially Synthetic & Numeric + Text & 6 & \xmark & Simplistic & \cmark \\
\midrule
\textbf{CaTS-Bench} & HR + Synthetic & Numeric + Text + Visual & 11 & Rich & Expressive & \cmark \\
\bottomrule
\end{tabular}
}
\label{tab:benchmark_comparison}
\end{table*}

\paragraph{LLMs for Time Series Analysis.} LLMs are increasingly repurposed for temporal reasoning, with early efforts focused on forecasting and anomaly detection \citep{zhang2024large, liu2024large}. Methodological approaches include specialized prompt engineering \citep{liu2024large, chatzigeorgakidis2024multicast}, modality alignment \citep{liu2024timecma, suntest, liu2024taming, pan2024s, cai2024jolt}, numeric discretization \citep{ansari2024chronos, jin2023time}, and parameter-efficient finetuning \citep{zhou2023onefitsall, chang2023llm4ts}. While these studies demonstrate that models pretrained on massive text corpora can inherit temporal reasoning capabilities, they also highlight persistent "hallucinations" in numeric extrapolation and difficulty handling long-range dependencies \citep{tang2025time, merrill2024language, tan2024are, cao2024evaluation, zeng2023transformers}. We shift the focus from purely predictive tasks to generative description, exploring how these models translate numeric signals into language.

\paragraph{Related Benchmarks and TSC.} Traditional time series archives \citep{UCRArchive, bagnall2018uea, godahewa2021monash} focus on discriminative tasks, while prompt-based forecasting benchmarks like PISA \citep{xue2023promptcast} lack the linguistic pairings and metadata required for descriptive tasks. While recent TSC-specific frameworks have pioneered the area \citep{fons2024tadacap, jhamtani2021truth, fons2024evaluating, dohi2025domain, sen2025bedtimeunifiedbenchmarkautomatically}, they typically rely on simplistic trend patterns or template-based synthetic generation, failing to capture the descriptive nuance of human discourse (see Table~\ref{tab:benchmark_comparison}). Furthermore, although integrating auxiliary modalities, such as metadata and visual renderings, is vital for grounded interpretability \citep{zhou2025can, dong2024metadata, chen2024visionts, wang2024news, kim2024multi, williams2024contextkeybenchmarkforecasting, liu2025can, tang2023vistext}, a unified benchmark integrating numeric data, expressive human captions, and multimodal grounding has remained elusive. CaTS-Bench addresses this gap with a unified benchmark for grounded TSC that integrates numeric data, visual representations, contextual metadata, and human language.

\section{CaTS-Bench}
In this section, we describe the construction of CaTS-Bench, starting from raw numeric series to our dual-stream captioning framework. CaTS-Bench provides an HR gold standard for evaluation, alongside a large-scale synthetic corpus for data augmentation. Moreover, we further enrich the scope of CaTS-Bench by providing an additional suite of Q\&A tasks constructed from the same data, enabling a more fine-grained examination of time series and caption reasoning abilities.

\subsection{Base Data Curation}

We work with $11$ diverse real-world source datasets that span climate, public safety, border crossings, demography, health, sales, and agriculture. See Appendix~\ref{src_details} for details of the source datasets. This variety ensures that models must generalize across different temporal scales, units, and domains. 

\begin{figure*}[ht]
    \centering
    \includegraphics[width=0.9\textwidth, trim=10 15 10 15, clip]{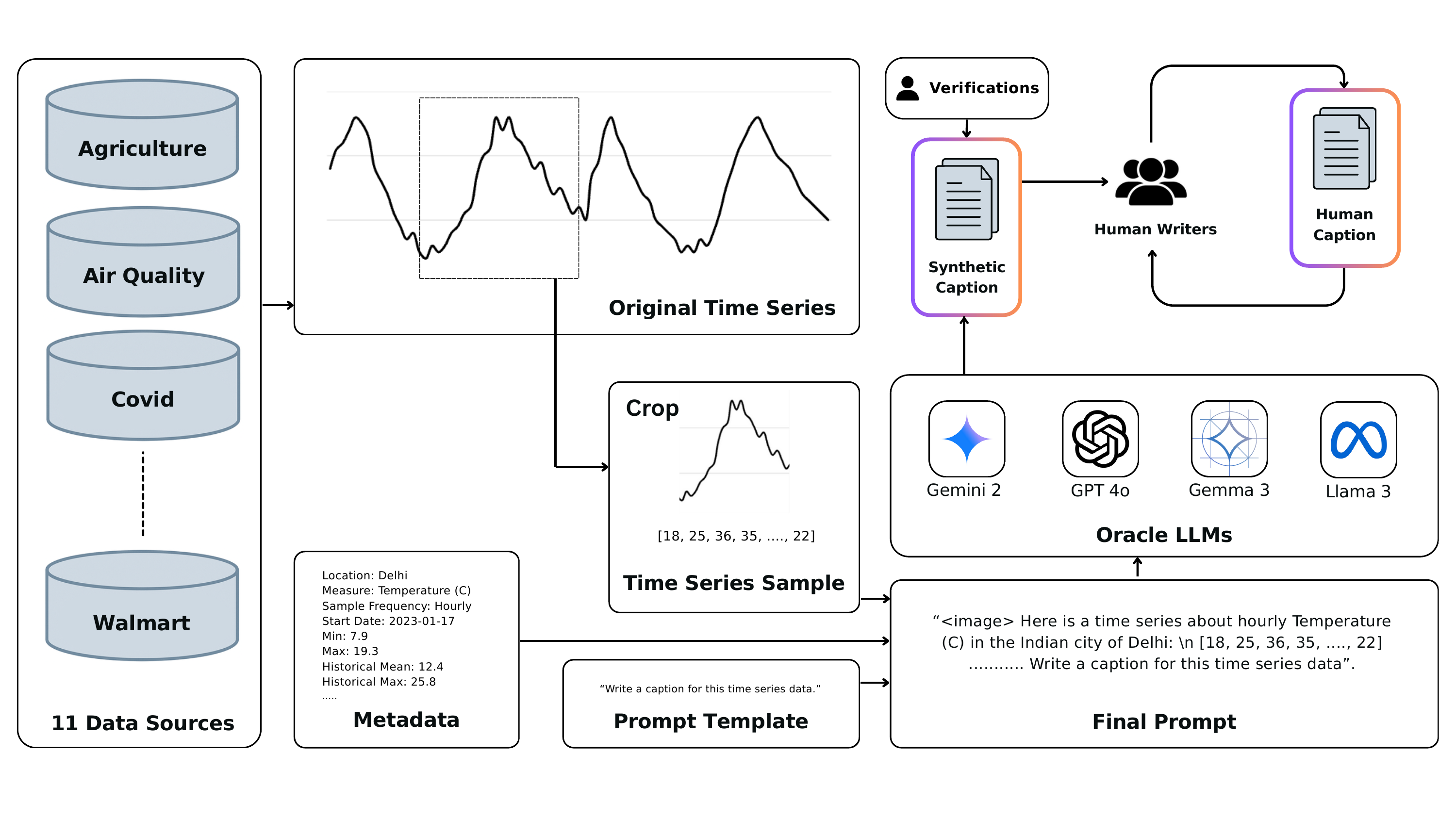}
    \caption{\textbf{Overview of CaTS-Bench's dual-stream data generation pipeline.} 
    A time series window is cropped, metadata is attached, and a human or LLM writes a reference caption. 
    See Appendix~\ref{example samples} for examples and Appendix~\ref{quality} for the quality verification protocol.}
    \label{fig:catbench}
\end{figure*}

The curation process begins by generating \textbf{base data triplets}. For each entity in a source dataset (e.g., a specific country or product), we apply a random window cropping strategy to extract segments of varying lengths (See Appendix \ref{cropping}). For each segment, we construct a triplet consisting of (i) the \textbf{raw numeric time series}, (ii) a structured \textbf{metadata JSON} containing domain-specific context (e.g., units, location, temporal bounds) and statistical primitives (mean, dispersion, minimum, and maximum), and (iii) a \textbf{line plot} rendering of the series with randomized visual styles (color, width, and size) to promote robustness to visual variation. The overview is shown in Figure~\ref{fig:catbench}. To prevent information leakage, we partition the data temporally: the first 80\% of each series is reserved for training/augmentation, while the final 20\% is used for the actual benchmark data. The sizes of data splits are reported in Table \ref{tab:dataset_statistics}. 

\begin{table*}[ht]
\caption{Dataset outline by domain. AQ: Air Quality, Border: Border Crossing, Demo: Demography, Injury: Road Injuries, Diet: Calories Consumption, Agri: Agriculture}
\centering
\resizebox{\textwidth}{!}{
\setlength{\tabcolsep}{3pt}      
\renewcommand{\arraystretch}{0.9} 
\begin{tabular}{l@{\hskip 6pt}|c@{\hskip 4pt}|c@{\hskip 4pt}c@{\hskip 4pt}c@{\hskip 4pt}c@{\hskip 4pt}c@{\hskip 4pt}c@{\hskip 4pt}c@{\hskip 4pt}c@{\hskip 4pt}c@{\hskip 4pt}c@{\hskip 4pt}c}
\toprule
\rowcolor[gray]{0.95} \textbf{Metric} & \textbf{Overall} & \textbf{AQ} & \textbf{Border} & \textbf{Crime} & \textbf{Demo} & \textbf{Injury} & \textbf{COVID} & \textbf{CO\textsubscript{2}} & \textbf{Diet} & \textbf{Walmart} & \textbf{Retail} & \textbf{Agri} \\
\midrule
\# Source Time Steps     & 287M  & 286M & 397k & 38k  & 14k  & 37k  & 720k & 34k  & 234k & 6k  & 7k  & 49k \\
\# Triplets Generated    & 20k   & 4.4k & 3.2k & 764  & 598  & 756  & 5.5k & 732  & 2.1k & 544 & 551 & 835 \\
\midrule
\# Human Test Samples   & 1746  & 0  & 323  & 153  & 120 & 152  & 300 & 100  & 211  & 109 & 111 & 167 \\
Average TS Length      & 26.8  & - & 21.5 & 76.9 & 5.0  & 3.6  & 71.8 & 8.8  & 5.5  & 11.8 & 8.1 & 7.5 \\
\midrule
\# Synth. Train Samples  & 16k & 3.5k & 2.6k & 611  & 478  & 604  & 4.4k & 585  & 1.7k & 435 & 440 & 668 \\
Average TS Length     & 29.1  & 65.3 & 21.2 & 76.8 & 11.6 & 5.9  & 75.8 & 9.5  & 12.2 & 12.2 & 22.4 & 7.3 \\
\midrule
\# Synth. Test Samples   & 4k  & 886  & 646  & 153  & 120  & 152  & 1.1k & 147  & 422  & 109 & 111 & 167 \\
Average TS Length      & 26.1  & 66.0 & 21.2 & 76.9 & 5.0  & 3.6  & 73.0 & 8.7  & 5.5  & 11.8 & 8.1 & 7.5 \\
\bottomrule
\end{tabular}
}
\label{tab:dataset_statistics}
\end{table*}

\subsection{Data Generation Streams}
Using the base data triplets as a foundation, we employ two distinct streams to generate natural language descriptions: one focused on HR evaluation and the other on scalable training.

\paragraph{Human-Rewritten Gold Standard.} 
To mitigate the potential oracle bias and stylistic homogeneity inherent in single-model benchmarks, we curated a primary evaluation set of $1746$ HR captions. To ensure broad linguistic coverage and minimize model-specific artifacts, we utilized a multi-stage curation process: we initially generated candidate drafts from four distinct LLMs (\texttt{Gemini 2.0 Flash}, \texttt{GPT-4o}, \texttt{Gemma 27B}, and \texttt{Llama 3 70B}). Using these diverse drafts as structural scaffolds, we performed extensive rewriting and verification (see Appendix~\ref{sec:interface} for details and results of the editing process and Appendix~\ref{sec:embedding-diversity} for diversity analysis). 

Across all domains, human rewriting resulted in substantial linguistic revision (average Word Error Rate = $0.60$) while maintaining strong numeric overlap (average Numeric Jaccard = $0.74$). This approach allowed us to move beyond simple post-editing, resulting in captions that are factually grounded, linguistically diverse, and enriched with the interpretive nuances that purely automated systems typically miss. These HR descriptions serve as the gold standard for CaTS-Bench, providing a rigorous test of a model's ability to generate high-fidelity, context-aware narratives. Beyond evaluation, these captions reflect how humans summarize time series patterns for rapid comprehension.

\paragraph{Synthetic Augmentation Pipeline.}
To facilitate large-scale training, we utilize a scalable pipeline to generate additional reference captions. We query an oracle LLM (\texttt{Gemini 2.0 Flash}) with a structured prompt comprising the full numeric sequence, enriched metadata with precomputed statistics, and the corresponding line plot. The oracle is strictly instructed to remain within the bounds of the provided data, ensuring that the generated captions are factually grounded (see Appendix~\ref{gt generation prompt} for a prompt example; see Appendix~\ref{metadata-motivation} for an ablation study on the role of metadata). We emphasize that the use of a single oracle model is intended as a proof of concept; different or multiple oracles can be used. While the synthetic data generation primarily serves as a large-scale training resource, we also utilize it to establish a dual-evaluation framework. 

We generate a synthetic corpus of approximately $20$k captions. Of these, $16$k are used for training, while the full synthetic test split contains roughly $4$k captions. For evaluation, however, we restrict attention to a subset of $1746$ synthetic captions that exactly correspond to the HR evaluation set in the underlying time series. This dual-track allows us to compare synthetic and HR captions on identical samples, revealing where model-generated descriptions diverge from human-rewritten ones.

\subsection{Synthetic Data Quality Validation}
\label{sec:quality-verification}

A key concern when using synthetic captions for training is whether they are factually accurate, linguistically diverse, and free from artifacts that hinder generalization to human-authored text. To validate our synthetic data pipeline, we conduct three complementary analyses, with full protocols and results reported in Appendix~\ref{quality}.

First, we perform a manual factual validation over approximately $2.9k$ synthetic captions, verifying statistical claims and trend descriptions against the underlying numeric series. This analysis shows that over $98.6\%$ of captions are factually correct, indicating minimal numeric hallucination. Second, we conduct a human indistinguishability study with $35$ participants, who are asked to distinguish synthetic captions from human-rewritten ones in a blind setting. The resulting detection accuracy is $41.1\%$, nearing random chance, suggesting that the synthetic captions are stylistically and semantically comparable to human-rewritten descriptions. Finally, we assess linguistic diversity to rule out template-driven generation. Embedding-based similarity analysis shows that near-duplicate captions (cosine similarity $>0.95$) occur in only $2.3\%$ of cases, and complementary N-gram and vocabulary analyses confirm broad lexical coverage comparable to human text (Appendix~\ref{sec:embedding-diversity}). These results show that synthetic captions provide high-fidelity training signals that closely approximate human descriptions.

\subsection{Task 1: Time Series Captioning}
The primary objective of TSC is the generation of a linguistically coherent narrative describing the salient temporal characteristics of a numeric series. During evaluation, models are presented with a multimodal prompt comprising four components: i) \textbf{numeric series} consisting of raw, time-indexed values that require direct numeric reasoning; ii) \textbf{contextual metadata} providing semantic background (e.g., measurement units and location), while explicitly excluding any statistical summaries of that series to ensure models perform genuine statistical inference; iii) a \textbf{visual representation} via a line-plot image that enables VLMs to utilize visual trend cues and pattern recognition; and iv) a standardized \textbf{instruction template} specifying the desired style and scope of the output (see Appendix~\ref{caption eval prompt}). Crucially, the task is not limited to identifying trends, patterns, or summary statistics, but evaluates whether models can synthesize numeric evidence, visual cues, and semantic context into a coherent \emph{free-form} natural language narrative that reflects not only what information is selected, but how it is composed and whether the resulting narrative is informative and correct, rather than producing templated or fragmentary observations.

\subsection{Task 2: Diagnostic Q\&A}
To complement the main captioning task, we introduce a suite of five multiple-choice Q\&A tasks designed to isolate and probe specific reasoning bottlenecks. These tasks are automatically derived from our base data triplets, with distractors designed to expose model failures in numeric precision and multimodal alignment. To ensure a high level of difficulty, we initially generated a pool of $4$k questions per category and filtered them by removing those correctly answered by \texttt{Qwen 2.5 Omni}, focusing the benchmark on "hard" reasoning cases (see Appendix~\ref{qwen filter} for an analysis of this filtering). The final diagnostic set consists of $910$ questions across the tasks below, with examples shown in Appendix \ref{qa-samples}:

\textbf{Time Series Matching.}
Models must identify the correct time series segment corresponding to a given caption from a set of distractors. These distractors are created via adversarial perturbations (shuffling, temporal reversal, Gaussian noise injection) to ensure that models cannot rely on simple matching and must instead understand the underlying temporal trend (Figure~\ref{fig:ts matching}; details in~\ref{artificial distractors}).

\textbf{Caption Matching.}
Given a numeric series, the model must select the correct caption from a set of semantic and numeric distractors. This isolates the model's \textit{understanding} of descriptive language from its ability to \textit{generate} fluent text, specifically targeting sensitivities to perturbed statistical claims (Figure~\ref{fig:caption matching}; prompt in~\ref{semantic perturbation prompt}).

\textbf{Plot Matching.}
The model is asked to select the correct line plot for a given caption and series. It specifically measures whether the visual modality provides a complementary signal or if the model remains biased toward textual priors (Figure~\ref{fig:plot matching}).

\textbf{Comparative Reasoning.}
Given a pair of time series, the model must select the correct comparative statement (e.g., “Series A exhibits higher variance than Series B”)  (Figures~\ref{fig:amplitude},~\ref{fig:peak},~\ref{fig:mean},~\ref{fig:variance}).  This probes the model’s ability to perform relative statistical assessment, a documented area of weakness for generative models \citep{merrill2024language} 

\textbf{Temporal Matching.} 
This task evaluates a model’s ability to perform exact temporal index alignment over time series data. In the first subtask (text-only), models are given a numeric time series as a sequence of values, along with a start timestamp and a query timestamp, and must select the correct value corresponding to the queried time. In the second subtask (image-based), the numeric series is replaced by a line plot where only the starting time is labeled, requiring the model to infer temporal offsets visually before retrieving the correct value. These subtasks isolate time indexing failures and test whether visual grounding helps or hinders temporal matching (Figures~\ref{fig:time-matching}, \ref{fig:time-matching-plot}).

\subsection{Evaluation Metrics}

We evaluate models using a multi-faceted framework to assess linguistic fluency, semantic alignment, and numeric grounding. Diagnostic Q\&A tasks are evaluated using standard accuracy. Below, we provide an overview of the metrics, with detailed definitions in Appendix \ref{eval-metrics}.

\textbf{Linguistic and Semantic Alignment.} We report \textsc{BLEU} \citep{papineni2002bleu}, \textsc{ROUGE-L} \citep{chin2004rouge}, and \textsc{METEOR} \citep{banerjee2005meteor} to measure surface-level lexical overlap. To better capture temporal and semantic nuances, we additionally use embedding-based similarity metrics, namely \textsc{DeBERTa Score} \citep{Zhang2020BERTScore} and \textsc{SimCSE} \citep{gao2021simcse}, which assess semantic consistency with human references independent of exact phrasing.

\textbf{Numeric Grounding and Hallucination.} A critical challenge in TSC is ensuring that language is accurately grounded in numeric reality. Standard NLP metrics are notoriously "number-blind"; a model can achieve a high BLEU score while hallucinating incorrect values. To address this, we introduce two tailored metrics:

\begin{enumerate}[leftmargin=*, itemsep=4pt, parsep=0pt, topsep=0pt]
    \item \textbf{Numeric Fidelity Score}: To measure the \textit{coverage} of numeric information, we extract all numeric values from the HR reference (excluding dates) and match them with numbers in the model's output. We calculate \textit{Accuracy} (precision of the matched numbers) and \textit{Recall} (the proportion of reference numbers successfully captured). The score is a weighted harmonic combination ($\lambda_A = 0.3, \lambda_R = 0.7$), reflecting our priority for recall; in a decision-making context, omitting a critical data point is often more detrimental than minor rounding differences.
    
    \item \textbf{Statistical Inference Accuracy}: This set of metrics evaluates a model's ability to correctly infer and verbalize key statistical references (mean, STD, min, max) from raw time series. We parse statistical claims from the generated text with an LLM and verify them against the true precomputed values (details in Appendix~\ref{parsing prompt}). We report the percentage of correctly identified statistics within a $5\%$ relative error to factor out rounding errors (details on the tolerance design are in Appendix~\ref{tolerance design}). Crucially, this is a \textit{precision-oriented} metric: models are not penalized for omitting a statistic but heavily penalized for reporting an incorrect one, thereby directly measuring hallucination. (see Appendix~\ref{sec:attempt-analysis} for details on attempt analysis)
\end{enumerate}

\section{Experiments}
We evaluate a number of VLMs on CaTS-Bench, covering both proprietary and open-source models. For TSC, we finetune some open-source models with the synthetic training data, and additionally consider a \textit{program-aided} (PAL) model \citep{gao2023PAL}. Finetuning details can be found in Appendix~\ref{finetune_config}. All models are prompted with the same template-based format to ensure fair comparison, avoiding task- or architecture-specific prompt engineering. Model details and PAL are in Appendix~\ref{model_details}; the human baseline is in Appendix~\ref{human baseline}.

\begin{table*}[ht]
\caption{Evaluation results of generated captions against HR and synthetic  (Synth) ground truths. The "Numeric Metrics" block comprises our tailored metrics (note: \textit{Numeric}: numeric fidelity score, \textit{Mean Inf.} = accuracy of detecting the time series mean). \textbf{Bolded} and \underline{underlined} scores denote first and second places.}
\centering
\scriptsize
\setlength{\tabcolsep}{3pt}
\renewcommand{\arraystretch}{1.05}

\resizebox{\textwidth}{!}{
\begin{tabular}{c|l|cc|cc|cc|cc|cc|cc|cc|cc|cc}
\toprule

\multirow{3}{*}{\textbf{}} &
\multirow{3}{*}{\textbf{Model}} &
\multicolumn{10}{c|}{\textbf{Lexical and Semantic Metrics}} &
\multicolumn{8}{c}{\textbf{Numeric Metrics}} \\
\cmidrule(lr){3-12}\cmidrule(lr){13-20}

& &
\multicolumn{2}{c}{\textbf{DeBERTa F1}} &
\multicolumn{2}{c}{\textbf{SimCSE}} &
\multicolumn{2}{c}{\textbf{BLEU}} &
\multicolumn{2}{c}{\textbf{ROUGE-L}} &
\multicolumn{2}{c|}{\textbf{METEOR}} &
\multicolumn{2}{c}{\textbf{Numeric}} &
\multicolumn{2}{c}{\textbf{Mean Inf.}} &
\multicolumn{2}{c}{\textbf{Max Inf.}} &
\multicolumn{2}{c}{\textbf{Min Inf.}} \\

\cmidrule(lr){3-4}\cmidrule(lr){5-6}\cmidrule(lr){7-8}
\cmidrule(lr){9-10}\cmidrule(lr){11-12}\cmidrule(lr){13-14}
\cmidrule(lr){15-16}\cmidrule(lr){17-18}\cmidrule(lr){19-20}

& &
HR & Synth & HR & Synth & HR & Synth & HR & Synth & HR & Synth & HR & Synth & HR & Synth & HR & Synth & HR & Synth \\

\midrule

\multirow{3}{*}{\rotatebox{90}{Propr.}}
 & Gemini 2.0 Flash
 & \underline{0.694} & \underline{0.717}
 & 0.884 & \underline{0.907}
 & \underline{0.113} & \underline{0.173}
 & \underline{0.283} & \underline{0.340}
 & 0.304 & \underline{0.378}
 & \textbf{0.757} & \textbf{0.795}
 & 0.894 & 0.894
 & 0.991 & 0.991
 & 0.985 & 0.985 \\


 & Claude 3 Haiku
 & 0.676 & 0.697
 & 0.856 & 0.873
 & 0.086 & 0.137
 & 0.261 & 0.305
 & 0.286 & 0.349
 & 0.669 & 0.712
 & 0.842 & 0.841
 & 0.983 & 0.983
 & 0.981 & 0.980 \\

 & GPT-4o
 & 0.685 & 0.702
 & \textbf{0.886} & 0.903
 & 0.090 & 0.128
 & 0.259 & 0.300
 & \textbf{0.314} & 0.374
 & 0.739 & 0.772
 & \underline{0.921} & \underline{0.921}
 & \textbf{0.998} & \textbf{0.998}
 & \textbf{0.995} & \textbf{0.995} \\


\midrule

\multirow{11}{*}{\rotatebox{90}{Pretrained}}
 & InternVL 2.5 38B
 & 0.682 & 0.704 & 0.867 & 0.889 & 0.085 & 0.144
 & 0.262 & 0.314 & 0.294 & 0.367 & 0.740 & 0.790
 & 0.865 & 0.865 & 0.983 & 0.982 & 0.978 & 0.978 \\

 & LLaVA v1.6
 & 0.641 & 0.655 & 0.796 & 0.814 & 0.064 & 0.088
 & 0.234 & 0.267 & 0.248 & 0.290 & 0.597 & 0.618
 & 0.488 & 0.489 & 0.792 & 0.791 & 0.614 & 0.614 \\

 & Idefics 2
 & 0.615 & 0.633 & 0.760 & 0.787 & 0.035 & 0.062
 & 0.213 & 0.255 & 0.169 & 0.206 & 0.514 & 0.528
 & 0.460 & 0.460 & 0.689 & 0.688 & 0.639 & 0.639 \\

 & SmolVLM
 & 0.572 & 0.586 & 0.691 & 0.709 & 0.029 & 0.053
 & 0.195 & 0.230 & 0.169 & 0.203 & 0.508 & 0.540
 & 0.476 & 0.494 & 0.860 & 0.860 & 0.647 & 0.647 \\

 & QwenVL
 & 0.638 & 0.656 & 0.809 & 0.832 & 0.060 & 0.095
 & 0.229 & 0.268 & 0.239 & 0.282 & 0.544 & 0.564
 & 0.496 & 0.496 & 0.788 & 0.789 & 0.601 & 0.601 \\

 & QwenVL PAL
 & 0.643 & 0.660 & 0.814 & 0.833 & 0.060 & 0.108
 & 0.241 & 0.287 & 0.224 & 0.273 & 0.637 & 0.656
 & \textbf{0.985} & \textbf{0.985}
 & \underline{0.992} & \underline{0.992}
 & \underline{0.993} & \underline{0.993} \\

 & Llama 3.2 Vision
 & 0.569 & 0.574 & 0.758 & 0.767 & 0.039 & 0.054
 & 0.200 & 0.222 & 0.226 & 0.256 & 0.665 & 0.701
 & 0.554 & 0.551 & 0.890 & 0.889 & 0.781 & 0.781 \\

 & Gemma 3 27B
 & 0.680 & 0.696 & 0.883 & 0.898 & 0.089 & 0.111
 & 0.255 & 0.289 & \underline{0.307} & 0.356 & 0.745 & 0.778
 & 0.816 & 0.816 & 0.984 & 0.983 & 0.973 & 0.973 \\

 & Phi-4 Multimodal Instruct
 & 0.650 & 0.666 & 0.828 & 0.849 & 0.071 & 0.113
 & 0.247 & 0.291 & 0.259 & 0.310 & 0.652 & 0.681
 & 0.617 & 0.618 & 0.875 & 0.876 & 0.824 & 0.825 \\

 & InternVL 2.5 8B
 & 0.666 & 0.686 & 0.856 & 0.875 & 0.078 & 0.123
 & 0.247 & 0.293 & 0.264 & 0.319 & 0.705 & 0.732
 & 0.731 & 0.731 & 0.985 & 0.985 & 0.933 & 0.934 \\

 & Gemma 3 12B
 & 0.677 & 0.695 & 0.871 & 0.888 & 0.093 & 0.110
 & 0.257 & 0.289 & 0.294 & 0.338 & 0.715 & 0.748
 & 0.767 & 0.765 & 0.967 & 0.966 & 0.949 & 0.949 \\

\midrule

\multirow{6}{*}{\rotatebox{90}{Finetuned}}
 & LLaVA v1.6
 & 0.644 & 0.659 & 0.802 & 0.819 & 0.064 & 0.090
 & 0.234 & 0.269 & 0.250 & 0.292 & 0.589 & 0.613
 & 0.500 & 0.498 & 0.827 & 0.826 & 0.615 & 0.615 \\

 & Idefics 2
 & \textbf{0.713} & \textbf{0.752}
 & \underline{0.885} & \textbf{0.914}
 & \textbf{0.131} & \textbf{0.278}
 & \textbf{0.325} & \textbf{0.442}
 & 0.303 & \textbf{0.427}
 & \underline{0.748} & \underline{0.791}
 & 0.859 & 0.859
 & 0.972 & 0.972
 & 0.937 & 0.937 \\

 & QwenVL
 & 0.678 & 0.705 & 0.863 & 0.888 & 0.090 & 0.166
 & 0.257 & 0.324 & 0.272 & 0.357 & 0.669 & 0.707
 & 0.744 & 0.745 & 0.977 & 0.978 & 0.942 & 0.943 \\

 & SmolVLM
 & 0.601 & 0.616 & 0.751 & 0.771 & 0.056 & 0.105
 & 0.224 & 0.273 & 0.197 & 0.243 & 0.604 & 0.632
 & 0.433 & 0.437 & 0.796 & 0.796 & 0.606 & 0.606 \\

 & Llama 3.2 Vision
 & 0.672 & 0.686 & 0.851 & 0.867 & 0.087 & 0.130
 & 0.266 & 0.309 & 0.295 & 0.352 & 0.740 & 0.761
 & 0.711 & 0.710 & 0.942 & 0.942 & 0.877 & 0.876 \\

 & Phi-4 Multimodal Instruct
 & 0.664 & 0.682 & 0.862 & 0.880 & 0.078 & 0.117
 & 0.243 & 0.285 & 0.290 & 0.349 & 0.703 & 0.743
 & 0.651 & 0.651 & 0.853 & 0.855 & 0.817 & 0.817 \\

\bottomrule
\end{tabular}
}

\label{tab:combined-eval}
\end{table*}

\subsection{Results on Time Series Captioning}
To ensure fair comparison across the domains, we report macro-averaged scores for each metric, mitigating sample size imbalances, as some domains contain more data, and preventing any domain from disproportionately influencing the results.

Our TSC results, shown in Table~\ref{tab:combined-eval}, reveal several key insights.
First, \textbf{proprietary models} consistently outperform pretrained open-source counterparts, with newer models generally establishing a stronger baseline.
However, \textbf{finetuned models} demonstrate substantial improvements across nearly all metrics, with several open-source models narrowing the gap to proprietary baselines, showcasing the advantage of finetuning on our quality-verified synthetic data.
Most notably, \textbf{Idefics 2} stands out as the most effective learner: upon finetuning, it achieves the strongest performance among open-source models on several linguistic metrics and substantially narrows the gap to proprietary baselines, achieving, for instance, a DeBERTa $F_1$ score of $0.713$ and a ROUGE-L of $0.325$ against human ground truth. Although its gains in statistical inference are the most pronounced among finetuned models, it still lags behind the strongest proprietary systems in the \textit{Numeric} category.
In contrast, \textbf{GPT-4o} exhibits exceptional proficiency in statistical inference, followed closely by \textbf{QwenVL PAL}.
The superior performance of the PAL variant over standard \textit{QwenVL} highlights the substantial benefit of program-aided reasoning and code execution for accurate numerical tasks.
Additionally, these results confirm that larger open-source models generally outperform their smaller counterparts.

Crucially, lexical and semantic scores are consistently lower when evaluated against HR ground truth than against synthetic ground truth. This consistent performance gap indicates that HR captions encode richer and less templatic reasoning patterns, and that the gains from finetuning primarily stem from exposure to higher-quality, more human-like data rather than increased data scale. Additionally, we evaluate some newer models only on the human-rewritten set in Section~\ref{sec:additional-models} and provide an information coverage analysis in Section~\ref{sec:info-cov-analysis}, showing how finetuning improves the inclusion of core time series reasoning elements.



\begin{figure*}[ht]
    \centering
    \begin{minipage}{0.35\linewidth}
        \centering
        \includegraphics[width=\linewidth]{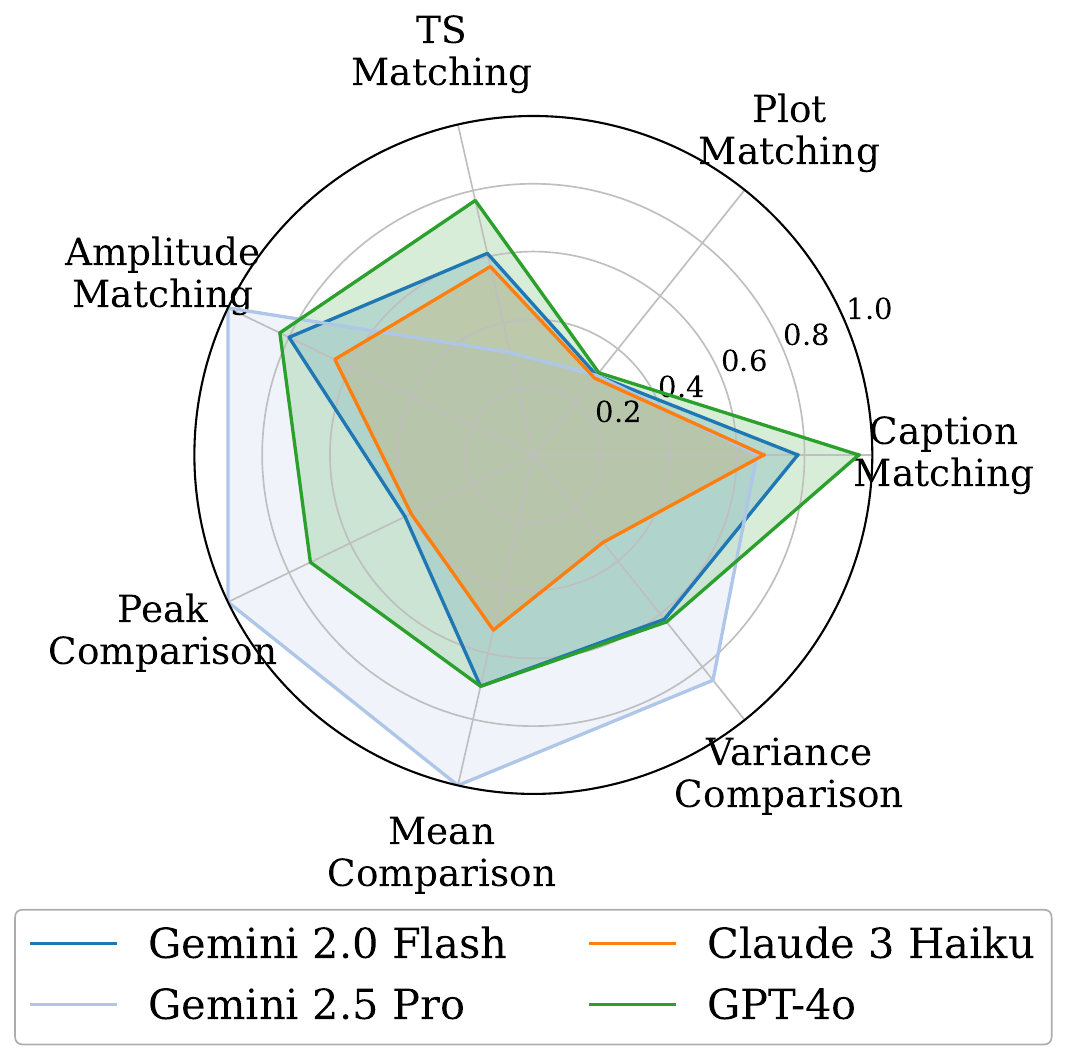}
        \caption*{(a) Proprietary VLMs}
    \end{minipage}
    \hspace{0.1\linewidth}
    \begin{minipage}{0.35\linewidth}
        \centering
        \includegraphics[width=\linewidth]{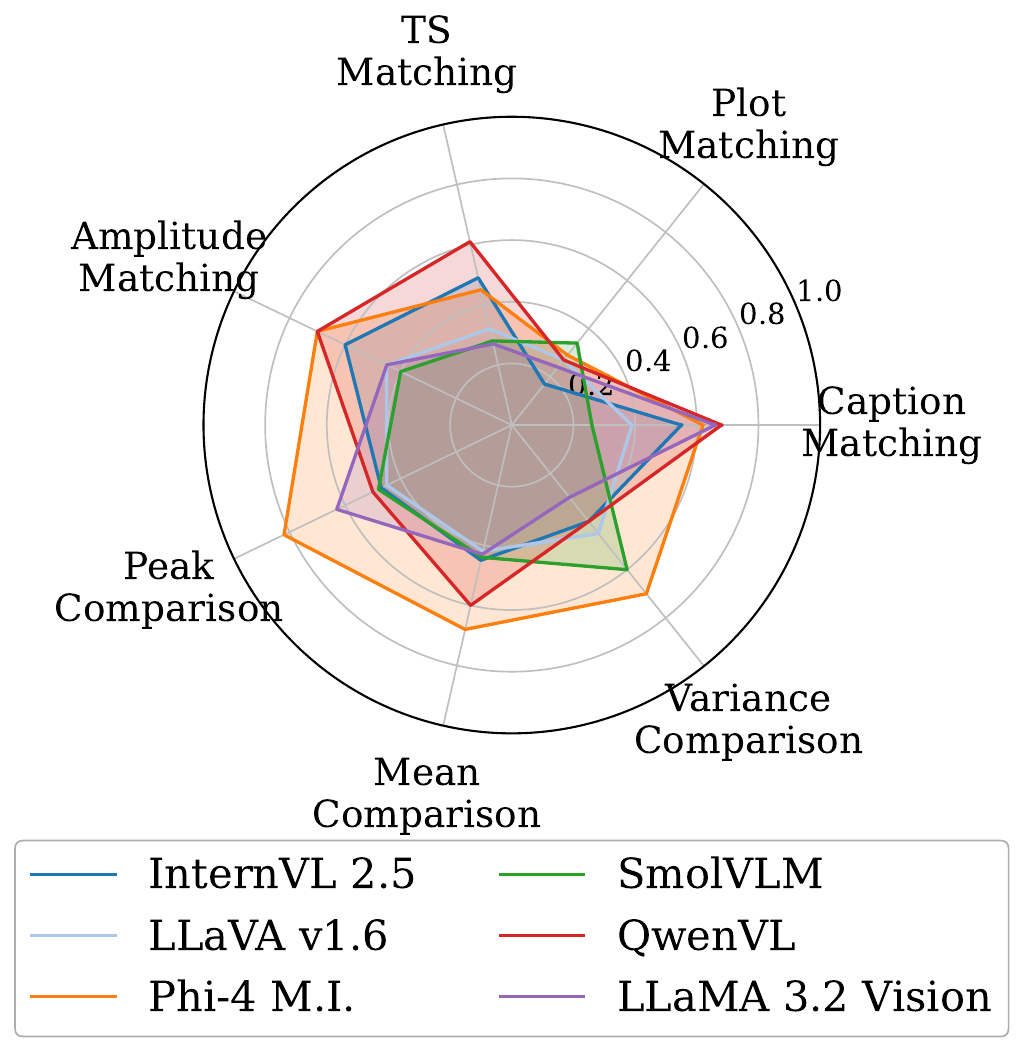}
        \caption*{(b) Open-source VLMs}
    \end{minipage}
    \caption{Model accuracy across Q\&A sub-tasks (except Temporal Matching). Proprietary models perform best, open-source models lag behind in all tasks.}
    \label{fig:radar}
\end{figure*}


\subsection{Results on Q\&A Tasks}
Figure~\ref{fig:radar} summarizes model performance on our Q\&A tasks (see Table~\ref{tab:qa_eval} for detailed results).

Models show no consistent winner across tasks. Binary time series comparisons are easier, likely due to the narrower decision space, while caption matching is harder, and plot matching is hardest. Proprietary models lead, and \texttt{Phi-4 M.I.} is the strongest open-source model overall. Humans achieve the highest overall scores and are nearly perfect on plot matching, where all models remain near random. Despite the tasks' apparent simplicity, they reveal fundamental limitations in VLMs' temporal reasoning capabilities, which suggests the need to address basic time series understanding before tackling more complex applications.

\begin{figure}[ht]
    \centering
    \includegraphics[width=\linewidth]{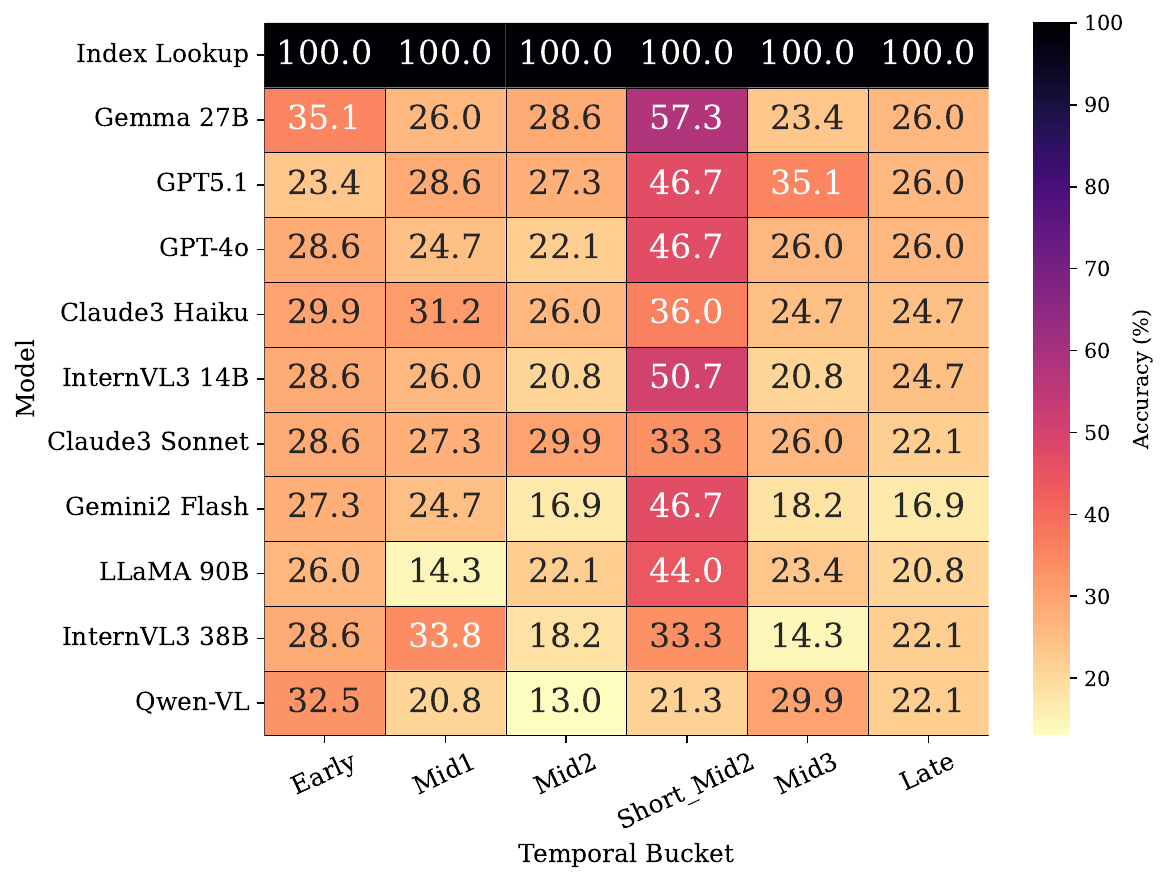}

\caption{MCQ accuracy (\%) for exact temporal value lookup at different query positions. \textit{Short\_Mid2} denotes sequences of length~6; other columns use long time series with queries at different temporal offsets. Results are averaged over 75 questions per temporal bucket.}

    \label{fig:datematch-heatmap}
\end{figure}

Temporal matching also unveils significant weaknesses, with results shown in Figure~\ref{fig:datematch-heatmap}. Models perform near chance on long sequences ($\approx 20$--$35\%$), with accuracy degrading as the queried time moves further from the series start in most models, while short sequences (length $=6$) provide higher accuracy ($\approx 50$--$57\%$). These results indicate that current language models struggle with exact index-based temporal reasoning from text, and increased model scale does not reliably mitigate this limitation. See Appendix~\ref{sec:temp-match-appendix} for more information about temporal buckets, baseline, and detailed results.

\begin{table}[ht]
\centering
\small
\caption{MCQ accuracy on \textit{Short-Mid2} questions when the time series is provided as numeric text, as a plotted image, or as both image and text.  Vision input improves performance for most models, but does not resolve failures in exact temporal indexing.}
\resizebox{\linewidth}{!}{
\begin{tabular}{lccc}
\toprule
\rowcolor[gray]{0.95}
\textbf{Model} & \textbf{Text Only} & \textbf{Image Only} & \textbf{Image+Text} \\
\midrule
Gemini 2.0 Flash         & 46.7 & 57.3 & 34.7 \\
Claude 3 Haiku      & 33.3 & 42.7 & 38.7 \\
GPT-4o               & 46.7 & 60.0 & 57.3\\
GPT-5.1 & 46.7 & 58.7 & 54.7 \\
InternVL3 14B & 50.7 & 60.0 & \underline{74.7} \\
InternVL3 38B         & 33.3 & 46.7 & \textbf{76.0} \\
LLaMA 90B            & 44.0 & 44.0 & 38.7 \\
Gemma 27B              & 57.3 & 58.7 & 58.7 \\
\bottomrule
\end{tabular}
}

\label{tab:vision_vs_text}
\end{table}

\paragraph{Modality Ablations for Temporal Matching.} Table~\ref{tab:vision_vs_text} compares model accuracy on \textit{mid2} Temporal Matching questions when the same underlying time series is provided as numeric text, as a plotted image, or as both image and text. Image-only input yields higher accuracy than text-only input for most models and often reaches the highest performance for a given model, suggesting that visual structure strongly supports coarse temporal localization. 

However, adding numeric text alongside the image frequently degrades performance, indicating that many models struggle to jointly integrate visual and textual representations of the same time series. This multimodal interference is evident for models such as \textit{Gemini 2.0 Flash}, \textit{GPT-4o}, and \textit{LLaMA-90B}, where image-only accuracy exceeds image-and-text performance. 

Notably, only the newer \textit{InternVL3} \citep{zhu2025internvl3exploringadvancedtraining} models, particularly the larger InternVL variants, consistently benefit from multimodal input and achieve the highest accuracies in the combined setting. These results suggest that while vision alone can aid approximate temporal reasoning, effective fusion of redundant visual and numeric signals remains a challenge for most models, with successful multimodal integration emerging only in more recent architectures.

\subsection{Role of the Visual Modality}

\paragraph{Visual Modality Ablation.} We perform a modality removal experiment by stripping away the time series plot and providing only the associated textual metadata and the numeric values of the time series. This quantifies the contribution of the visual channel and enables a better understanding of the model's captioning performance. We evaluate a selected subset of pretrained baselines to assess their intrinsic reliance on vision. 

\begin{figure}[t]
    \centering
    \includegraphics[width=\columnwidth]{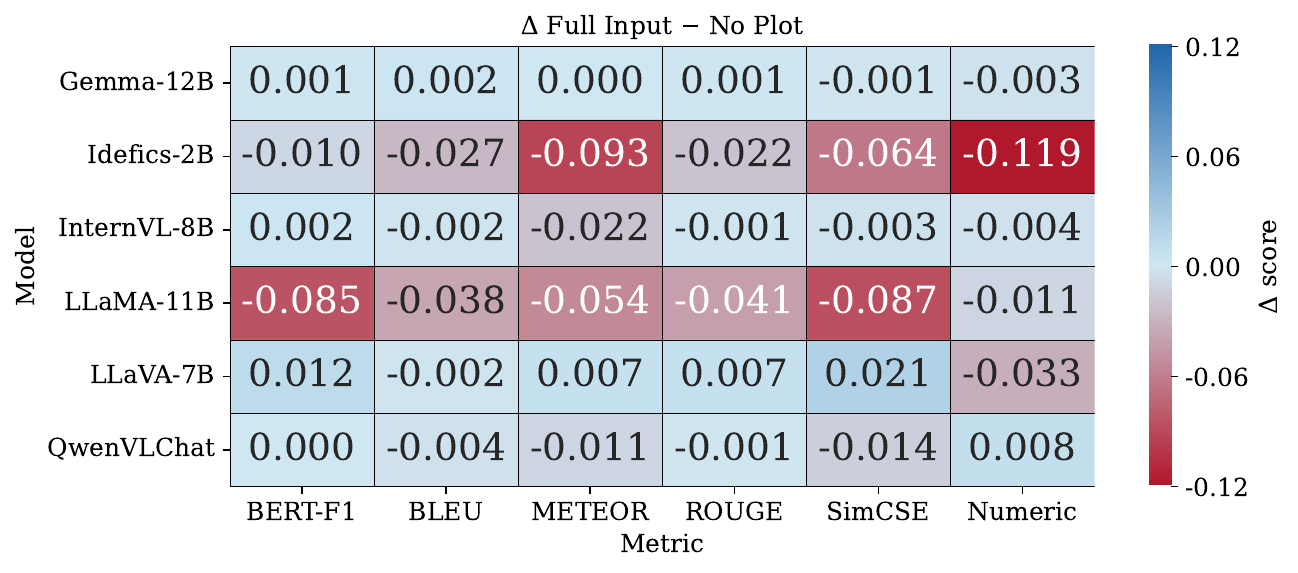}
 \caption{Performance change when adding visual input. Blue indicates improvement; red indicates degradation.}
    \label{fig:hmap}
\end{figure}

Our experiments show that visual input contributes little to caption quality for most models. As Figure~\ref{fig:hmap} illustrates, removing the plot image produces only minor changes in both semantic and numeric performance, with numeric accuracy degrading modestly and sometimes even improving without the plot. This suggests that models rely mainly on textual priors rather than grounded interpretation of the plotted signal, so caption generation is driven primarily by language-level analysis of typical time series behaviors.

The weak sensitivity to image removal indicates that visual input is only weakly used during generation, for both semantic and numeric metrics, implying that vision plays a limited role in numeric grounding. As discussed in Section~\ref{sec:visual-encodings}, this phenomenon is not restricted to line plots: more expressive encodings such as Gramian Angular Fields and recurrence plots likewise fail to elicit strong visual reasoning from current VLMs in TSC.

\begin{figure}[ht]
    \centering
    \includegraphics[width=\columnwidth]{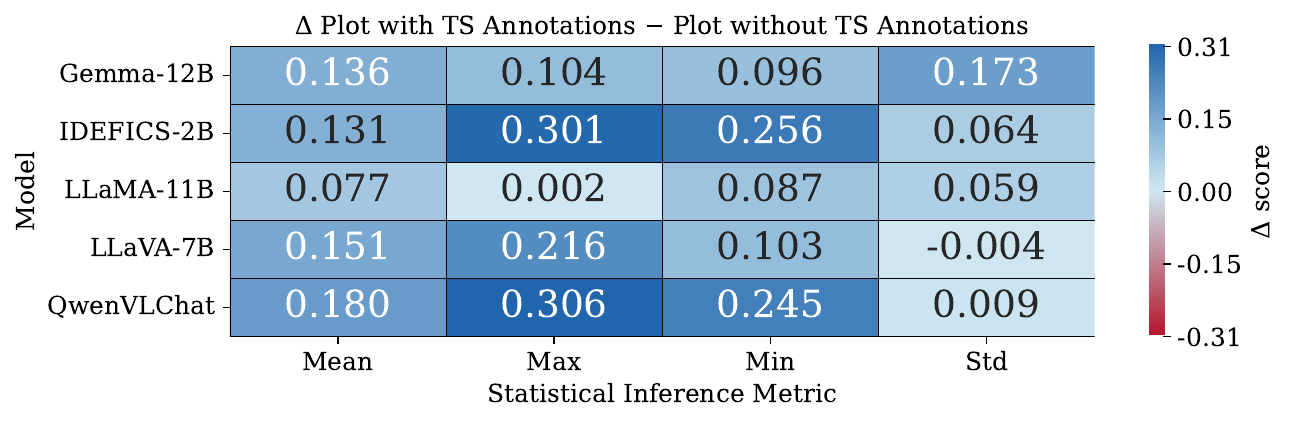}
\caption{Performance deltas on statistical inference with and without annotations. Blue indicates improvement; red degradation.}
    \label{fig:hmap-annotations}
\end{figure}

Figure~\ref{fig:hmap-annotations} isolates the effect of numeric annotations on the plot, while keeping the visual modality fixed. In contrast to the full-versus-no-plot comparison, the impact here is substantially larger. When time series values are explicitly annotated, models show clear improvements in fine-grained numeric metrics, particularly $\texttt{Max}$, $\texttt{Min}$, and $\texttt{Mean}$. Removing these annotations while retaining the plotted curve leads to pronounced drops in numeric fidelity. This pattern indicates that models benefit primarily from explicit numeric text embedded in the image, rather than from interpreting the visual shape of the time series curve. The relatively small and inconsistent changes in $\texttt{Std}$ further suggest that dispersion is not reliably inferred from visual structure alone.

\paragraph{Visual Attention Analysis.} To better understand how VLMs use plots during caption generation, we examine their attention maps and find minimal visual grounding: models focus mainly on textual elements (e.g., axis labels and titles) with limited attention to the actual line trends. Attention to visual patterns is weak and inconsistent, suggesting that learned parameters largely ignore visual cues in favor of textual priors, highlighting a gap between nominal multimodal input and actual integration (full results in~\ref{visual-attention-analysis} and Figure~\ref{fig:attention}).

These findings indicate that current VLM pretraining paradigms are ill-suited for structured visual reasoning over time series plots: despite strong performance on natural images, their objectives align poorly with abstract, low-level visual encodings that require relational, temporal, and quantitative interpretation rather than object recognition. Consequently, models default to textual priors and explicit numeric cues, largely ignoring visual structure, even though CaTS-Bench explicitly provides both plots and rich metadata to support multimodal reasoning. That most models fail to leverage visual signals exposes a critical gap in current VLM capabilities and motivates future work on architectures that more effectively integrate plot-based information with textual and numeric cues.

\section{Conclusion}
We presented \textbf{CaTS-Bench}, a multimodal benchmark for TSC and time series reasoning that emphasizes linguistic expressiveness, numeric fidelity, and grounded multimodal alignment. By combining a human-rewritten gold standard with a scalable synthetic augmentation pipeline, CaTS-Bench enables rigorous evaluation of free-form narrative generation over time series while addressing data scarcity. Our results show that proprietary VLMs provide strong zero-shot baselines, while finetuning open-source models on synthetic data yields substantial gains in TSC. However, zero-shot VLMs exhibit systematic failures such as hallucinated statistics in free-form captioning. Diagnostic Q\&A tasks further reveal persistent weaknesses in exact temporal indexing and multimodal integration, indicating that the current VLM pretraining paradigm is ill-designed for time series reasoning. Overall, CaTS-Bench exposes key limitations in grounded time series understanding and provides a foundation for developing more reliable multimodal generative models in numeric domains.

\section*{Limitations}
While CaTS-Bench represents a significant step toward multimodal, context-aware time series understanding, it also has limitations that suggest clear avenues for future work.

First, the training captions are synthetic and generated by a single oracle model (\texttt{Gemini 2.0 Flash}). Our extensive validation studies confirm their factual reliability and linguistic diversity; however, reliance on a single oracle may still introduce subtle modeling biases during finetuning. As discussed earlier, this design choice is intended as a proof of concept to demonstrate the feasibility and effectiveness of scalable augmentation. Extending the pipeline to multiple oracles is straightforward but beyond the scope of this work. Although we release a small subset of paraphrases generated by different LLMs, its limited scale highlights opportunities for future extensions.

Second, although we provide a human-rewritten test set, the scale of expert-verified content remains limited. We acknowledge that captions written by domain experts can offer deeper insights, and future work could incorporate expert-authored annotations across domains such as economics, healthcare, or climate data. At the same time, human authors introduce their own stylistic biases, which should be considered when designing and evaluating benchmarks.

Overall, we view CaTS-Bench as a scalable foundation rather than a fixed resource, with ample room to grow through richer human input, multi-oracle augmentation, and extended coverage of temporal reasoning tasks.

\section*{Ethical Considerations} \label{ethical}
The development of CaTS-Bench was guided by a commitment to ethical research practices. All datasets used in this work are publicly available and do not contain personally identifiable information (PII). The domains, such as climate, public health, and agriculture, were chosen for their public relevance and data accessibility. Our use of an oracle LLM for data augmentation was a deliberate design choice to ensure scalability. We have taken extensive measures to validate the quality of the synthetic captions, gauging their factual accuracy and diversity, as detailed in Section \ref{sec:quality-verification} and Appendix \ref{quality}. Our human-rewritten test set is also an attempt to further ensure evaluation reliability. For our human Q\&A evaluation studies, all participation was voluntary. We obtained informed consent from all participants, who were university students. The study's purpose was clearly communicated, and all responses were collected anonymously to protect participant privacy, as shown in an example of a consent form in Appendix \ref{human baseline}. 

\section*{LLM Usage Statement}
LLMs played a central role in multiple stages of this work. 

\begin{enumerate}[leftmargin=*]
    \item LLMs were employed as \textbf{data generators}, producing synthetic captions that mainly serve as training data.
    \item LLMs were employed as \textbf{data extractors}, for example to parse statistical claims from captions during our evaluation analyses. \item LLMs and VLMs served as \textbf{baselines} in our experiments as captioning models for evaluation.
    \item LLMs were employed as a \textbf{writing assist tool} to polish the presentation of the paper. 
\end{enumerate}
LLMs did not contribute to research ideation or decision-making. All factual claims, analyses, and conclusions are the responsibility of the authors.

\section*{Acknowledgments}
This work was supported in part by the U.S. Army Research Office
under Army-ECASE award W911NF-07-R-0003-03, the U.S. Department Of Energy, Office of Science, IARPA HAYSTAC Program, and NSF Grants \#2205093, \#2146343, \#2134274, \#2441832, CDC-RFA-FT-23-0069, DARPA AIE FoundSci and DARPA YFA. We also thank the anonymous reviewers for their constructive feedback, which significantly improved the clarity and presentation of this work.

\bibliography{custom}

\appendix


\label{sec:appendix}

\lstset{
  basicstyle=\ttfamily\small,
  breaklines=true,
  breakatwhitespace=true,
  breakindent=0pt,
  columns=fullflexible,
  keepspaces=true,
  showstringspaces=false,
  frame=single,
  backgroundcolor=\color{gray!10}
}

\newpage
\appendix
\section*{Appendix}

\section{Motivation and Utilities of TSC}\label{sec:motivation}

An interesting question to think about is whether TSC is a meaningful task at all, given that many properties of a time series (e.g., means, extrema, trends) can be computed exactly using deterministic algorithms or statistical tools. While such tools are indeed precise, they fundamentally address a different problem. TSC is not concerned with computing statistics, but with generating \emph{free-form natural language descriptions} that summarize, contextualize, and communicate temporal data in a human-interpretable way.

In many real-world settings, users do not interact directly with raw statistics or code outputs, but instead rely on narrative descriptions. Business reports, financial dashboards, scientific summaries, ECG reports, policy briefs, and automated analytics systems routinely present trends and patterns through textual explanations accompanying plots. In these contexts, the goal is not merely to compute correct values, but to decide \emph{what is salient}, \emph{how to describe it}, and \emph{how to express uncertainty, comparison, or change over time} in natural language. These decisions are inherently under-specified and context-dependent, making them unsuitable for purely deterministic pipelines.

LLMs are, therefore, a natural candidate for this task, as they are designed to produce fluent, context-aware descriptions. However, our results demonstrate that current models struggle precisely at this interface between numeric data and free-form generation. We observe frequent hallucination of numeric values, incorrect temporal alignment between values and timestamps, and failures in simple comparative reasoning. Moreover, when strong textual cues are present, models often ignore the underlying plot, generating plausible but weakly grounded descriptions driven by language priors rather than data. These failure modes are not exposed by deterministic statistical tools, which are guaranteed to be correct by construction, but are critical in generative settings where outputs are unconstrained and must be trusted by downstream users. The fact that calculators or statistical algorithms can compute accurate summaries does not obviate the need to evaluate how models \emph{describe} those summaries in natural language. Instead, it highlights the importance of studying TSC as a distinct problem, one that probes a model’s ability to integrate numeric reasoning, temporal structure, and linguistic generation.

From this perspective, TSC serves as a diagnostic task that reveals the limits of current LLMs and VLMs in grounded free-form generation. Our findings do not argue against deterministic methods, but rather show that replacing human-written narrative descriptions with model-generated captions with direct prompting remains challenging. Understanding these limitations is essential for developing reliable systems that can safely and effectively communicate temporal data in real-world settings.

\section{Source Datasets}\label{src_details}


\begin{enumerate}[leftmargin=*]
\item \textbf{Air Quality} – Hourly air pollution data from 453 Indian cities (2010–2023), covering 30+ parameters including PM\textsubscript{2.5}, NO\textsubscript{x}, CO, and SO\textsubscript{2}, compiled from CPCB \cite{jha2023airquality}.
    \item \textbf{Border Crossing} – Monthly inbound border crossing counts at U.S.-Mexico and U.S.-Canada ports, disaggregated by transport mode and collected by U.S. Customs and Border Protection \cite{bts_bordercrossing}.
\item \textbf{Crime} – Incident-level crime reports in Los Angeles from 2020 onward, provided by LAPD OpenData and updated biweekly, including NIBRS-compliant records \cite{lapd_crime_2020}.
\item \textbf{Demography} – Annual global indicators from the UN and World Bank (2000--2021) covering population growth, fertility, life expectancy, death rates, and median age to assess patterns of demographic change and collapse \cite{aziz1985population}. 
    \item \textbf{Injury} – Annual counts of fatal and severe road traffic injuries in California (2002–2010), disaggregated by transport mode and geography, from CDPH's Healthy Communities Indicators \cite{cdph_roadtrafficinjuries}.
\item \textbf{COVID} – Global daily COVID-19 case and death counts (2020), compiled by ECDC, covering over 200 countries with population-adjusted metrics \cite{ecdc_covid_data}.
    \item \textbf{CO\textsubscript{2}} – National-level per capita CO\textsubscript{2} emissions and GDP trends from Our World in Data, adjusted for trade (consumption-based), spanning 1990–2023 \cite{owid-co2-gdp-decoupling}.
    \item \textbf{Diet} – Food supply and caloric intake patterns from FAO Food Balance Sheets \cite{fao_diet_fbs}.
\item \textbf{Walmart} – Weekly sales data from 45 Walmart stores (2010–2012), enriched with features like temperature, fuel price, CPI, unemployment rate, and holiday flags \cite{kaggle_walmart}.
\item \textbf{Retail} – Transactional records from a UK-based online gift retailer (2010–2011), capturing item-level purchases, cancellations, and customer behavior \cite{online_retail_352}.
    \item \textbf{Agriculture} – Annual agricultural total factor productivity (TFP) indices from USDA for 1961--2022, covering outputs and inputs like land, labor, capital, and materials across countries \cite{usda_agriculture_productivity}.
\end{enumerate}

\section{Time Series Segment Cropping} \label{cropping}
 Our cropping strategy balances diversity with consistency across datasets. Many source time series (e.g., $50$ years of hourly CO2 emissions) are too long to process directly, so we sample random windows of variable lengths. Each window length is drawn from a dataset-specific range $[min, max]$, with the maximum based on the original series length. This ensures that cropped windows preserve the scale and structure of the data while introducing sufficient variability for training and evaluation. We summarize these rules in Table~\ref{tab:seq_lengths} below.


\begin{table}[ht]
\caption{Minimum and maximum segment lengths for each dataset.}
\label{tab:seq_lengths}
\centering
\small
\setlength{\tabcolsep}{6pt}
\renewcommand{\arraystretch}{1.15}

\resizebox{\columnwidth}{!}{
\begin{tabular}{c c c}
\toprule
\rowcolor[gray]{0.95}
\textbf{Source Dataset} & \textbf{Min Length} & \textbf{Max Length} \\
\midrule
\parbox[c]{0.35\columnwidth}{\centering
Air Quality, Crime,\\
Border Crossing, CO$_2$,\\
Walmart, Agriculture}
 & 5
 & $\min(150,\; 5 + \lfloor \tfrac{\text{original\_length}}{8} \rfloor)$ \\
\midrule
Demography, Online Retail
 & 5 & original\_length \\
\midrule
Road Injuries
 & 3 & $\min(3,\; 0.2 \times \text{original\_length})$ \\
\midrule
COVID
 & 5 & $\min(150,\; 5 + \lfloor \tfrac{\text{original\_length}}{5} \rfloor)$ \\
\midrule
Calories
 & 5 & $\min(6,\; 0.2 \times \text{original\_length})$ \\
\bottomrule
\end{tabular}
}
\end{table}

\section{Why Metadata?}\label{metadata-motivation}

A central challenge in synthetic data generation is ensuring factual fidelity. Free-form synthetic captions produced by unconstrained LLMs are often unreliable as they may hallucinate numeric values, introduce unsupported context, or mischaracterize extrema and temporal trends, thereby corrupting both supervision and evaluation. Such issues are especially problematic in time series captioning, where even minor numeric inaccuracies can significantly distort semantic interpretation.

To address this, our synthetic captioning pipeline is not based on open-ended prompting. Instead, the oracle is strictly grounded in three sources: (i) the underlying numeric time series, (ii) the corresponding plot, and (iii) enriched metadata capturing domain context and structural attributes. This grounding constrains generation and substantially reduces hallucination.

We validate the importance of metadata grounding through a controlled ablation study. When captions are generated without metadata and manually evaluated for statistical correctness, temporal consistency, and hallucinated values, factual accuracy drops to 10\% (5/50). In contrast, captions generated with metadata grounding achieve 98\% accuracy (49/50). This performance gap highlights that metadata is not merely auxiliary, but a critical component for ensuring high-fidelity synthetic supervision.

\section{Hardware and Settings}\label{finetune_config}

All experiments were conducted on a high-performance computing node featuring two \textit{AMD EPYC 7453} processors, providing a total of $56$ logical CPUs, and $125$ GB of RAM (with over $117$ GB available during runtime). For GPU acceleration, the system includes eight \textit{NVIDIA A100} GPUs - six PCIe $80$ GB models and two PCIe $40$ GB models - along with an ASPEED graphics controller used for display purposes. This configuration offers ample computational and memory resources suitable for mid- to large-scale deep learning training and inference. The models we finetune range in size from $2$ billion to $11$ billion parameters, with finetuning times spanning from a few hours to a day.

\begin{table}[ht]
\centering
\setlength{\tabcolsep}{1pt}
\renewcommand{\arraystretch}{0.9}
\caption{Finetuning configuration details}
\begin{tabular}{lc}
\toprule
\rowcolor[gray]{0.95} \textbf{Hyperparameter} & \textbf{Value} \\
\midrule
Batch size & 4 \\
Gradient Accumulation Steps & 12 \\
Training Epochs & 3 \\
Learning Rate & $2 \times 10^{-5}$ \\
Learning Rate Scheduler & Cosine \\
Optimizer & AdamW  \\
Numerical Precision & bfloat16 \\
LoRA Rank & 8 or 16 \\
Dropout Probability & 0.05 \\
Input Image Resolution & 224--560 \\
\bottomrule
\end{tabular}
\label{tab:finetune-hparams}
\end{table}

For finetuning, we adopt a unified training strategy guided by best practices in instruction tuning for multimodal inputs (see Table~\ref{tab:finetune-hparams}). All models are trained using the AdamW optimizer with a cosine learning rate scheduler and a base learning rate of $2 \times 10^{-5}$. We apply gradient accumulation to simulate a larger batch size. Mixed precision training and gradient checkpointing are enabled for memory efficiency. Low Rank Adaptation (LoRA) is used to adapt large models by tuning a small subset of parameters, while keeping the rest of the model frozen or partially frozen. To ensure deterministic and focused generation, we use a temperature of $0.3$ during inference across all evaluated models. Each model is finetuned using a structured JSONL dataset comprising time series plot images and corresponding image-grounded chat-style conversations. We preprocess data with each model’s native processor and apply minimal resizing to maintain fidelity in the visual input. Special care is taken to exclude padding and <image> tokens from loss computation by assigning them an ignore index.


\section{Baseline Models}\label{model_details}

For TSC, we evaluate \texttt{Gemini 2.0 Flash} \citep{geminiteam2025geminifamilyhighlycapable}, \texttt{Claude 3 Haiku} \citep{TheC3}, \texttt{GPT-4o} \citep{openai2024gpt4technicalreport}, \texttt{GPT-5.1} \citep{openai2025gpt51}, \texttt{InternVL 2.5 (8b \& 38b)} \citep{chen2025expandingperformanceboundariesopensource}, \texttt{LLaVA v1.6 Mistral 7b} (default) and \texttt{34b} \citep{liu2023llava}, \texttt{Phi-4 Multimodal Instruct 5.6b} \citep{abdin2024phi4technicalreport}, \texttt{Idefics 2 (8b)} \citep{laurenccon2024matters}, \texttt{SmolVLM (2b)} \citep{marafioti2025smolvlm}, \texttt{QwenVL (7b)} \citep{Qwen-VL}, \texttt{Llama 3.2 Vision (11b)} \citep{grattafiori2024llama3herdmodels}, and \texttt{Gemma 3 (12b \& 27b)} \citep{gemmateam2025gemma3technicalreport} for both TSC and Q\&A tasks.

TSC requires precise numeric reasoning alongside text generation, making it suitable for program-aided language (PAL) models \citep{gao2023PAL}. Hence, we also evaluate \texttt{QwenVL 32b} by prompting it to generate a Python program that outputs the full time series caption. The program is executed in Python, and its return value is taken as the caption. Most (~90\%) generated programs succeed on the first attempt; if a program fails, we increase the token limit and regenerate until successful. The full prompt example can be found in Appendix~\ref{PAL_prompt}.

\section{Evaluation Metrics}\label{eval-metrics}

\subsection{Linguistic Metrics}\label{ling-eval-metrics}

\paragraph{DeBERTa Score}
The \textsc{DeBERTa Score} is a contextual similarity metric based on cosine similarity between contextual embeddings of tokens in the candidate ($c$) and reference ($r$) captions. Given token embeddings from the DeBERTa encoder, the metric computes token-level precision, recall, and F1:

\begin{equation}
\begin{aligned}
\text{F1}_{\text{DeBERTa}} &= \frac{2PR}{P + R}, \\
\text{where}\quad
P &= \frac{1}{|c|} \sum_{i \in c} \max_{j \in r} \cos(\mathbf{e}_i, \mathbf{e}_j), \\
R &= \frac{1}{|r|} \sum_{j \in r} \max_{i \in c} \cos(\mathbf{e}_j, \mathbf{e}_i).
\end{aligned}
\end{equation}

where $\mathbf{e}_i$ and $\mathbf{e}_j$ are the contextual embeddings of candidate and reference tokens, respectively.

\paragraph{BLEU}
\textsc{BLEU} evaluates n-gram overlap between a candidate caption and reference using precision with a brevity penalty to discourage short outputs:

\begin{equation}
\begin{aligned}
\text{BLEU} &= \text{BP} \cdot \exp\!\left( \sum_{n=1}^N w_n \log p_n \right), \\
\text{where}\quad
\text{BP} &=
\begin{cases}
1, & \text{if } c > r, \\
e^{\,1 - r/c}, & \text{if } c \le r,
\end{cases}
\end{aligned}
\end{equation}

where $p_n$ is the modified precision for $n$-grams, $w_n$ are weights (usually uniform), $c$ is candidate length, and $r$ is reference length.

\paragraph{ROUGE-L}
\textsc{ROUGE-L} measures fluency via the length of the longest common subsequence (LCS) between candidate and reference:

\begin{equation}
\begin{aligned}
\text{ROUGE-L}_{\text{F1}} &= \frac{(1 + \beta^2)\cdot \text{LCS}}{r + c}
\end{aligned}
\end{equation}

where $\beta$ balances recall and precision (often $\beta = 1$), and $r$ and $c$ are the reference and candidate lengths. 

\paragraph{METEOR}
\textsc{METEOR} aligns unigrams using exact matches, stems, synonyms, and paraphrases. It then computes an F-score and applies a fragmentation penalty:

\begin{equation}
\begin{aligned}
\text{METEOR} &= F_{\text{mean}} \cdot (1 - \text{Pen}), \\
\text{where}\quad
F_{\text{mean}} &= \frac{10 \cdot P \cdot R}{R + 9P}, \\
\text{Pen} &= 0.5 \left( \frac{\text{chunks}}{\text{matches}} \right)^3.
\end{aligned}
\end{equation}

where $P$ and $R$ are unigram precision and recall, and chunks refers to non-contiguous matched segments.

\paragraph{SimCSE}
\textsc{SimCSE} computes semantic similarity at the sentence level using cosine similarity between sentence embeddings:

\begin{equation}
\text{SimCSE}(c, r) = \cos\left( \mathbf{h}_c, \mathbf{h}_r \right) = 
\frac{\mathbf{h}_c \cdot \mathbf{h}_r}{\|\mathbf{h}_c\| \|\mathbf{h}_r\|}
\end{equation}

where $\mathbf{h}_c$ and $\mathbf{h}_r$ are candidate and reference sentence embeddings, produced by a contrastively trained RoBERTa encoder.

\subsection{Tolerance Design in Numeric Metrics} \label{tolerance design}
We adopt a $5\%$ relative tolerance for both numeric metrics, as it is a widely accepted threshold in numeric evaluation across data science and time series literature. This value balances sensitivity and robustness: it is tight enough to catch meaningful deviations from the true value, ensuring that significant errors are penalized, yet lenient enough to accommodate minor variations due to rounding, numeric precision, or natural approximations in model-generated captions. By using this standard threshold, our evaluation aligns with common practice while focusing on practically relevant numeric accuracy.

\section{Quality Validation}\label{quality}


\subsection{Manual Verification of Synthetic Captions}
\label{sec:manual-verification}

To assess the reliability of synthetic captions used as ground truth, we manually verified statistical and trend claims in around $2.9k$ captions of the test set under a three-tier scoring system: \emph{exact} (within $\pm0.05$ of the true value), \emph{near} (within $\pm0.1$), and \emph{incorrect}.

\begin{table}[ht]
\centering
\footnotesize
\setlength{\tabcolsep}{4pt}
\renewcommand{\arraystretch}{0.95}
\vspace{-1em}
\caption{Manual verification of SS captions. Accuracy remains high across all categories.}
\begin{tabular}{lll r}
\toprule
\rowcolor[gray]{0.95}
\textbf{Task} & \textbf{Feature/Type} & \textbf{Occurences} & \textbf{Acc.} \\
\midrule
\multirow{4}{*}{Statistical}
 & mean / average         & 570 & 0.980 \\
 & minimum / dip          & 310 & 0.994 \\
 & standard deviation     & 123 & 1.000 \\
 & maximum / peak         & 217 & 0.994 \\
\midrule
\multirow{4}{*}{Trend}
 & Upward                 & 467  & 0.980 \\
 & Downward               & 328 & 0.997 \\
 & Stability              & 45  & 0.970 \\
 & Fluctuation            & 87  & 0.974 \\
\midrule
\multirow{5}{*}{\shortstack{Historical \\ (Trends + Stats)}}
 & mean / average      & 435 & 0.980 \\
 & standard deviation  & 26  & 1.000 \\
 & maximum / peak      & 95  & 0.994 \\
 & minimum / dip       & 12  & 1.000 \\
 & norms               & 164 & 0.980 \\
\midrule
\textbf{Total / Avg.} &  & \textbf{2879} & \textbf{0.986} \\
\bottomrule
\end{tabular}
\label{tab:verification}
\end{table}

\paragraph{Method} Each claim was extracted and checked against the underlying time series metadata. We extracted statistical claims and trend patterns from captions using structured keyword clustering across selected statistical categories (mean/average, min/max, standard deviation) and trend keywords (increasing, decreasing, plateau, fluctuation). We (1) identified terms using keyword matching, (2) extracted ground truth values therein, and (3) created verification sheets comparing claims against actual metadata for manual verification. Then, we scored each claim by comparing statements against ground truth data and verifying that trend claims matched actual trajectories. 

Accuracy across both statistical descriptors and trend types is consistently high as shown in Table~\ref{tab:verification}, confirming that oracle-generated captions provide factually reliable reference annotations.

\subsection{Human Study on Detectability}
\label{sec:human-detectability}

We conducted a blind study with $35$ participants, each reviewing $11$ captions (roughly half written by humans, half by Gemini, using the same context information). Participants labeled each as human or AI-written but achieved only $41.1\%$ accuracy, essentially random, suggesting \textbf{Gemini’s captions were indistinguishable from human ones}. Participation form with guidelines is in Appendix~\ref{human baseline}.

\subsection{Robustness of Evaluation} \label{robsutness of evaluation}
A legitimate concern when using a single LLM to generate reference captions is the potential for evaluation bias towards the specific linguistic style of that model. To ensure that CaTS-Bench evaluates generalizable time series understanding capabilities rather than the tendency to mimic Gemini's linguistic style, we conducted a robustness experiment.

\begin{table*}[htb]
\caption{Evaluation of generated captions across paraphrased/original ground truths.}
\centering
\scriptsize
\setlength{\tabcolsep}{4pt}
\renewcommand{\arraystretch}{1.1}
\resizebox{\textwidth}{!}{
\begin{tabular}{l|l|cc|ccc|c|ccc}
\toprule
\rowcolor[gray]{0.95}
\textbf{Model} & \textbf{GT Style} 
& \multicolumn{2}{c|}{\textbf{Embedding}} 
& \multicolumn{3}{c|}{\textbf{N-gram}} 
& \multicolumn{1}{c|}{\textbf{Numeric}} 
& \multicolumn{3}{c}{\textbf{Stat. Inference}} \\
\cmidrule(lr){3-4} \cmidrule(lr){5-7} \cmidrule(lr){8-8} \cmidrule(lr){9-11}
 &  & SimCSE & DeBERTa 
 & METEOR & ROUGE-L & BLEU 
 & Numeric 
 & Mean & Max & Min \\
\midrule
\multirow{4}{*}{\makecell{Gemini \\ 2.0 Flash}} 
 & GPT-4o Phr.   & 0.8707 & 0.6803 & 0.2313 & 0.2605 & 0.0820 & 0.6715 & 0.3333 & 0.9823 & 0.9377 \\
 & Gemma Phr.    & 0.8635 & 0.6748 & 0.2068 & 0.2443 & 0.0605 & 0
 .
 6578 & 0.5357 & 0.9823 & 0.9374 \\
 & Llama Phr.    & 0.8680 & 0.6715 & 0.2080 & 0.2445 & 0.0710 & 0.6745 & 0.3750 & 0.9807 & 0.9351 \\
 & Original      & 0.8716 & 0.6860 & 0.2720 & 0.3068 & 0.1315 & 0.6802 & 0.3750 & 0.9823 & 0.9377 \\
\midrule
\multirow{4}{*}{GPT-4o} 
 & GPT-4o Phr.   & 0.8752 & 0.6740 & 0.2583 & 0.2488 & 0.0845 & 0.6773 & 0.8000 & 0.9921 & 0.9393 \\
 & Gemma Phr.    & 0.8645 & 0.6665 & 0.2250 & 0.2268 & 0.0578 & 0.6673 & 0.8167 & 0.9921 & 0.9393 \\
 & Llama Phr.    & 0.8726 & 0.6678 & 0.2295 & 0.2380 & 0.0725 & 0.6673 & 0.8000 & 0.9921 & 0.9379 \\
 & Original      & 0.8773 & 0.6785 & 0.2880 & 0.2785 & 0.1048 & 0.6558 & 0.8000 & 0.9921 & 0.9379 \\
\midrule
\multirow{4}{*}{\makecell{Claude \\ 3 Haiku}} 
 & GPT-4o Phr.   & 0.8636 & 0.6720 & 0.2497 & 0.2480 & 0.0693 & 0.6383 & 0.7500 & 0.9797 & 0.9338 \\
 & Gemma Phr.    & 0.8563 & 0.6665 & 0.2263 & 0.2295 & 0.0495 & 0.6223 & 0.8000 & 0.9781 & 0.9321 \\
 & Llama Phr.    & 0.8598 & 0.6678 & 0.2320 & 0.2473 & 0.0605 & 0.6285 & 0.8333 & 0.9797 & 0.9339 \\
 & Original      & 0.8683 & 0.6795 & 0.2873 & 0.2870 & 0.1038 & 0.6333 & 0.7500 & 0.9797 & 0.9338 \\
\midrule
\multirow{4}{*}{Idefics 2} 
 & GPT-4o Phr.   & 0.7952 & 0.6113 & 0.1463 & 0.2035 & 0.0243 & 0.3893 & 0.8056 & 0.8908 & 0.8377 \\
 & Gemma Phr.    & 0.7897 & 0.6058 & 0.1368 & 0.1915 & 0.0158 & 0.4005 & 0.8056 & 0.8908 & 0.8377 \\
 & Llama Phr.    & 0.7850 & 0.6045 & 0.1335 & 0.1950 & 0.0183 & 0.4198 & 0.8056 & 0.8908 & 0.8377 \\
 & Original      & 0.7962 & 0.6178 & 0.1623 & 0.2250 & 0.0380 & 0.4580 & 0.8056 & 0.8908 & 0.8377 \\
\midrule
\multirow{4}{*}{QwenVL} 
 & GPT-4o Phr.   & 0.8347 & 0.6323 & 0.2250 & 0.2205 & 0.0483 & 0.4098 & 0.4375 & 0.7948 & 0.6793 \\
 & Gemma Phr.    & 0.8262 & 0.6278 & 0.2043 & 0.2025 & 0.0350 & 0.4160 & 0.4375 & 0.7926 & 0.6793 \\
 & Llama Phr.    & 0.8270 & 0.6303 & 0.2113 & 0.2213 & 0.0453 & 0.4470 & 0.4375 & 0.7948 & 0.6776 \\
 & Original      & 0.8427 & 0.6405 & 0.2548 & 0.2488 & 0.0798 & 0.4895 & 0.4375 & 0.7926 & 0.6776 \\
\midrule
\multirow{4}{*}{\makecell{Llama \\ 3.2 Vision}} 
 & GPT-4o Phr.   & 0.8663 & 0.6625 & 0.2663 & 0.2473 & 0.0795 & 0.6890 & 0.4667 & 0.9562 & 0.8937 \\
 & Gemma Phr.    & 0.8591 & 0.6575 & 0.2393 & 0.2315 & 0.0572 & 0.6810 & 0.4722 & 0.9562 & 0.8952 \\
 & Llama Phr.    & 0.8647 & 0.6613 & 0.2588 & 0.2545 & 0.0793 & 0.6915 & 0.4722 & 0.9546 & 0.8961 \\
 & Original      & 0.8704 & 0.6680 & 0.2990 & 0.2843 & 0.1060 & 0.6853 & 0.4722 & 0.9562 & 0.8937 \\
\bottomrule
\end{tabular}
}
\label{tab:paraphrase_results}
\end{table*}

We systematically paraphrased a representative subset of our ground truth captions (agriculture, crime, demography, Walmart) using architecturally distinct LLMs (GPT-4o, Gemma 27B, and Llama 90B), resulting in three additional sets of ground truth captions. The paraphrasing prompt was designed to instruct the model to thoroughly alter sentence structure, syntax, and word choice while preserving all factual information, numeric values, and statistical details with absolute fidelity. The prompt used for paraphrasing is shown in~\ref{caption paraphrasing prompt}. We define this paraphrased caption set as CaTS-Bench-Paraphrased, which contains captions that are semantically equivalent to the original ground truth but differ only in linguistic style.

We then re-evaluated the outputs of all six representative pretrained models against CaTS-Bench-Paraphrased using the full suite of metrics. Results of this analysis are presented in the Table~\ref {tab:paraphrase_results}, one for each ground-truth generator's linguistic style. Values represent the average across select domains. We also report the results obtained with our original Gemini captions as ground truth.

\begin{figure*}[htb]
    \centering
    \includegraphics[width=\linewidth]{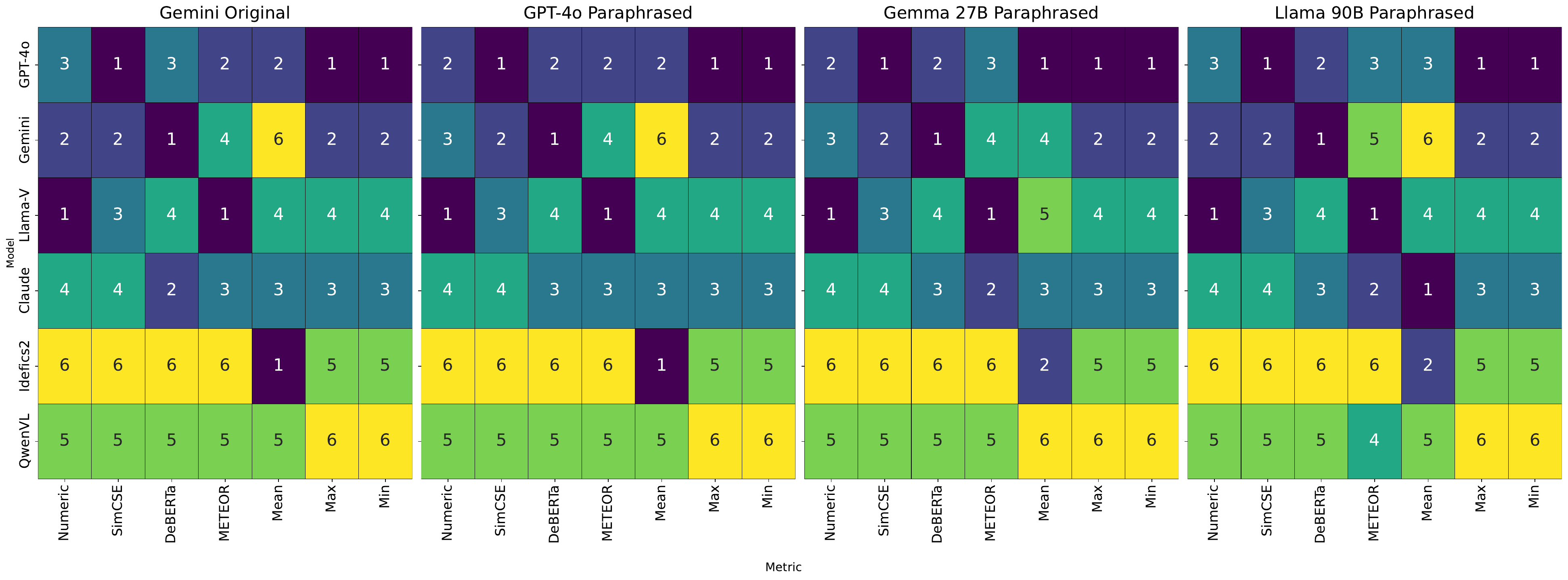}
    \caption{Model ranking heatmaps across metrics  
    under four reference styles. 
    Rankings: 1 (highest) to 6 (lowest). Model mappings: Gemini (Gemini 2.0 Flash), Claude (Claude 3 Haiku), Llama-V (Llama 3.2 Vision).}
    \label{fig:appendix_ranks}
\end{figure*}

Next, we provide an analysis of the rank correlation of model performances between the original and paraphrased evaluation settings. A high correlation in model rankings would indicate that our benchmark is robust to linguistic variation; the metrics would be consistently measuring the underlying semantic content and numeric accuracy of the captions, not their surface-level similarity to a specific writing style. A low correlation would suggest a non-trivial dependence on the oracle's particular linguistic patterns. For each linguistic evaluation metric, we provide the model ranking across the four linguistic styles in Figure~\ref{fig:appendix_ranks}.

Qualitatively, we observe a negligible impact of model-specific linguistic style on the model rankings, suggesting that our evaluations are robust to particular linguistic styles. For each linguistic metric, we measure the average Spearman correlation between the ranking according to Gemini's style and the ranking according to the other three styles (see Table~\ref{tab:spearman_correlations}).  

\begin{table}[ht]
\caption{Spearman correlation between Gemini-based rankings and rankings induced by different metrics. A value of $1$ indicates identical rankings.}
\centering
\small
\setlength{\tabcolsep}{6pt}
\renewcommand{\arraystretch}{1.15}
\begin{tabular}{lcc}
\toprule
Metric & Correlation & $p$-value \\
\midrule
DeBERTa F1 & 0.971 & 0.0018 \\
SimCSE & 1.000 & 0.0000 \\
BLEU & 0.814 & 0.0557 \\
ROUGE-L & 0.905 & 0.0257 \\
METEOR & 0.943 & 0.0079 \\
\midrule
\textbf{Average} & \textbf{0.927} & \textbf{0.0182} \\
\bottomrule
\end{tabular}
\label{tab:spearman_correlations}
\end{table}

In summary, this experiment demonstrates that our evaluation framework is robust to variations in the linguistic style of the reference captions. This conclusion is quantitatively supported by the high Spearman correlation coefficients observed in model rankings between the original and paraphrased benchmark sets. These results indicate that \textbf{our evaluation framework captures the semantic fidelity and factual quality of the generated content, rather than rewarding models for merely mimicking the stylistic patterns of our oracle model, Gemini 2.0 Flash}. Consequently, the benchmark evaluates fundamental capabilities in time series understanding and description, not superficial stylistic alignment. Furthermore, Gemini 2.0 Flash maintains its superior rank in most metrics regardless of the linguistic style of the ground truth. This consistent dominance validates its selection as a highly capable oracle, reinforcing that its utility stems from its intrinsic ability to generate high-quality descriptions rather than from any benchmark-specific bias.

\subsection{Diversity Analysis}
\label{sec:embedding-diversity}

\subsubsection{Content in the Captions} 

We analyzed all the $4005$ Gemini-generated test set captions using a structured keyword-based approach (see Table~\ref{tab:stat-trend-coverage}). Captions across 11 domains were scanned for statistical descriptors (e.g., \emph{mean}, \emph{average}, \emph{standard deviation}, \emph{maximum}, \emph{minimum}, \emph{range}) and trend-related expressions (e.g., \emph{increase}, \emph{decrease}, \emph{stability}, \emph{volatility}, \emph{seasonality}). These terms were grouped into clusters such as \emph{Central Tendency}, \emph{Dispersion}, \emph{Extremes}, \emph{Increasing/Decreasing Trends}, \emph{Stability and Volatility}, and \emph{Comparative Trends}. The results in Table~\ref{tab:caption-content-coverage} show that \textbf{captions consistently draw from a diverse mix of descriptors, spanning both statistical features and temporal patterns}. While some categories (e.g., percentiles, distribution-shape terms) were rare, coverage of core descriptors was broad, and every caption included at least one statistical or trend-related element.

We apply the same keyword-based analysis to the HR caption set to examine the types of statistical and temporal descriptors used.
Across domains, human captions draw from a wide range of descriptor clusters, including measures of central tendency, extrema, dispersion, and multiple forms of trend characterization. In addition to core statistical terms, human captions frequently include compound trend expressions, contextual qualifiers, and multi-stage temporal descriptions that span increases, decreases, stabilization, and recovery phases within a single caption.

Coverage extends beyond surface-level descriptors to include higher-order patterns such as structural transitions, regime changes, and shape-based trends, which appear with greater regularity in the human set. 
While all human captions contain at least one statistical or trend-related element, many incorporate several distinct clusters simultaneously, reflecting a richer and more varied composition of quantitative and temporal content.

\begin{table*}[ht]
\centering
\footnotesize
\setlength{\tabcolsep}{8pt}
\renewcommand{\arraystretch}{1.05}
\caption{Coverage of statistical and temporal descriptor clusters across human captions. For each cluster, representative surface-form keywords are shown (non-exhaustive).}
\begin{tabular}{l | l | r}
\toprule
\rowcolor[gray]{0.95}
\textbf{Category} & \textbf{Cluster (Representative Keywords)} & \textbf{Captions} \\
\midrule
\multirow{7}{*}{Statistical}
 & Central Tendency (\emph{average, mean, median, mode, count, etc.})               & 3184 \\
 & Maximum Values (\emph{max, maximum, peak, highest value, record high, etc.})     & 1487 \\
 & Minimum Values (\emph{min, minimum, dip, trough, lowest value, etc.})            & 1530 \\
 & Dispersion (\emph{standard deviation, deviation, variance, std, iqr, etc.})     & 684  \\
 & Variability (\emph{variability, variation, fluctuate, oscillation, noise, etc.})& 665  \\
 & Range / Spread (\emph{range, spread, difference, gap, extent, etc.})             & 341  \\
 & Extremes (\emph{peak, spike, trough, extreme, record high, etc.})                & 2966 \\
\midrule
\multirow{10}{*}{Trend and Patterns}
 & Increasing Trends (\emph{increase, rising, growing, upward, gain, etc.})         & 1207 \\
 & Decreasing Trends (\emph{decrease, declining, falling, drop, reduction, etc.})   & 1308 \\
 & Stability / Flat Trends (\emph{stable, steady, flat, constant, plateau, etc.})   & 1082 \\
 & Sudden Changes (\emph{spike, jump, shock, abrupt, rapid increase, etc.})          & 219  \\
 & Volatility (\emph{fluctuate, volatile, oscillate, erratic, instability, etc.})   & 1210 \\
 & Comparative Relationships (\emph{higher than, lower than, exceeds, below, gap, etc.}) & 1433 \\
 & Structural Changes (\emph{inflection, shift, transition, regime change, pivot, etc.}) & 190  \\
 & Recovery / Correction (\emph{recovery, rebound, normalize, correction, bounce back, etc.}) & 362 \\
 & Cyclical Patterns (\emph{seasonal, cyclical, periodic, recurring, repeats, etc.})& 135  \\
 & Shape-based Patterns (\emph{v-shaped, u-shaped, w-shaped, rise-and-fall, hill-shaped, etc.}) & 75 \\
\bottomrule
\end{tabular}
\label{tab:stat-trend-coverage}
\end{table*}

\begin{table}[ht]
\centering
\footnotesize
\caption{Coverage of statistical and trend descriptor clusters across synthetic captions. Captions consistently include diverse descriptors capturing both quantitative and temporal aspects of the data.}
\resizebox{\columnwidth}{!}{%
\begin{tabular}{c | l | r}
\toprule
\rowcolor[gray]{0.95}
\textbf{Category} & \textbf{Cluster and Keywords in the Cluster} & \textbf{Captions} \\
\midrule
\multirow{6}{*}{\rotatebox{90}{\textbf{Statistical}}}
 & Central Tendency (mean, average, median, mode) & 3930 \\
 & Minimum Values (min, minimum, lowest value)    & 1528 \\
 & Maximum Values (max, maximum, highest value)   & 1487 \\
 & Dispersion (std, variance, deviation, iqr)     & 1376 \\
 & Range / Spread (range, spread)                 & 492  \\
 & Extremes (peak, spike, dip, trough)            & 2966 \\
\midrule
\multirow{5}{*}{\rotatebox{90}{\parbox{2cm}{\centering\textbf{Trend \& \\ Patterns}}}}
 & Peaks and Valleys (peak, dip, spike, trough)   & 3013 \\
 & Increasing Trends (increase, rising, growing)  & 1987 \\
 & Decreasing Trends (decrease, drop, falling)    & 2197 \\
 & Comparative Trends (higher/lower, difference)  & 2217 \\
 & Stability and Volatility (stable, fluctuating) & 2276 \\
\bottomrule
\end{tabular}
}
\label{tab:caption-content-coverage}
\end{table}

\subsubsection{N-Gram Diversity} \label{ngram}
We analyze n-gram diversity using the type–token ratio (TTR) to assess whether the generated captions exhibit repetitive or templated phrasing. For the synthetic corpus, computed over approximately $411$k tokens, TTR increases steadily with n-gram order, rising from $0.0288$ for unigrams to $0.6050$ for 5-grams. This monotonic growth indicates rapid diversification at the phrase level rather than reliance on fixed templates. In particular, the presence of over $250$k unique 5-grams out of $411$k tokens suggests substantially higher-order lexical variability in the synthetic captions.

We perform the same analysis on the human-rewritten gold captions to provide a reference point. As shown in Table~\ref{tab:ttr-only}, HR captions exhibit consistently higher TTR across all n-gram orders, reflecting the greater compositional variability expected from human-authored text. Importantly, the synthetic captions follow a similar upward trajectory with increasing n, differing primarily in magnitude rather than qualitative behavior. This pattern suggests that while synthetic captions are less diverse than their human-revisited counterparts, they remain far from being rigidly templated and capture a broad range of phrase structures.

\begin{table}[ht]
\centering
\small
\caption{Type–token ratio (TTR) for synthetic and human-rewritten captions across n-gram orders. TTR increases monotonically with n for both corpora, indicating growing phrase-level diversity. Human-rewritten captions exhibit higher TTR across all n, while synthetic captions remain highly non-templated at higher n-grams.}
\begin{tabular}{lcc}
\toprule
\textbf{N-gram} & \textbf{Synthetic TTR} & \textbf{Human TTR} \\
\midrule
1-gram & 0.0288 & 0.0472 \\
2-gram & 0.1288 & 0.2320 \\
3-gram & 0.2971 & 0.5011 \\
4-gram & 0.4638 & 0.7066 \\
5-gram & 0.6050 & 0.8344 \\
\bottomrule
\end{tabular}

\label{tab:ttr-only}
\end{table}

\begin{table*}[ht]
\centering
\caption{Embedding similarity statistics for human-rewritten and synthetic captions across encoder models.}
\scriptsize
\setlength{\tabcolsep}{4pt}
\renewcommand{\arraystretch}{1.15}
\resizebox{\textwidth}{!}{
\begin{tabular}{l r | r r r r r r | r r r r r r}
\toprule
\rowcolor[gray]{0.95}
\textbf{Encoder} & \textbf{Dim} &
\multicolumn{6}{c|}{\textbf{Human Captions}} &
\multicolumn{6}{c}{\textbf{Synthetic Captions}} \\
\cmidrule(lr){3-8} \cmidrule(lr){9-14}
 &  &
Mean & Med & STD & Intra & Inter & \%$>0.95$ &
Mean & Med & STD & Intra & Inter & \%$>0.95$ \\
\midrule
MiniLM-L12-v2  & 384  & 0.2755 & 0.2437 & 0.1560 & 0.6090 & 0.2308 & 0.17 &
                       0.2932 & 0.2500 & 0.1771 & 0.6202 & 0.2269 & 5.26 \\
MiniLM-L6-v2   & 384  & 0.3367 & 0.3113 & 0.1429 & 0.6291 & 0.2975 & 0.12 &
                       0.3339 & 0.3013 & 0.1510 & 0.5999 & 0.2800 & 1.47 \\
mpnet-base-v2  & 768  & 0.3260 & 0.2859 & 0.1679 & 0.6866 & 0.2776 & 0.44 &
                       0.3278 & 0.2791 & 0.1775 & 0.6631 & 0.2599 & 8.29 \\
bge-large-v1.5 & 1024 & 0.5921 & 0.5753 & 0.0955 & 0.7903 & 0.5655 & 0.23 &
                       0.5807 & 0.5605 & 0.0986 & 0.7600 & 0.5443 & 1.67 \\
mxbai-large-v1 & 1024 & 0.5344 & 0.5111 & 0.1096 & 0.7691 & 0.5030 & 0.24 &
                       0.5293 & 0.5032 & 0.1116 & 0.7370 & 0.4871 & 2.15 \\
Qwen3-4B       & 2560 & 0.3833 & 0.3613 & 0.1195 & 0.6344 & 0.3496 & 0.02 &
                       0.3816 & 0.3530 & 0.1273 & 0.6098 & 0.3354 & 0.20 \\
Qwen3-8B       & 4096 & 0.3800 & 0.3571 & 0.1188 & 0.6306 & 0.3465 & 0.00 &
                       0.3736 & 0.3465 & 0.1225 & 0.5947 & 0.3288 & 0.32 \\
e5-mistral-7b  & 4096 & 0.6469 & 0.6354 & 0.0670 & 0.7839 & 0.6285 & 0.01 &
                       0.6507 & 0.6358 & 0.0697 & 0.7770 & 0.6251 & 1.33 \\
\midrule
\textbf{Average} &  &
\textbf{0.4344} & \textbf{0.4101} & \textbf{0.1222} & \textbf{0.6910} & \textbf{0.3999} & \textbf{0.15} &
\textbf{0.4386} & \textbf{0.4092} & \textbf{0.1256} & \textbf{0.6678} & \textbf{0.3922} & \textbf{2.30} \\
\bottomrule
\end{tabular}
}
\label{tab:embedding-comparison}
\end{table*}

\begin{figure*}[ht]
    \centering
    \includegraphics[width=0.9\linewidth]{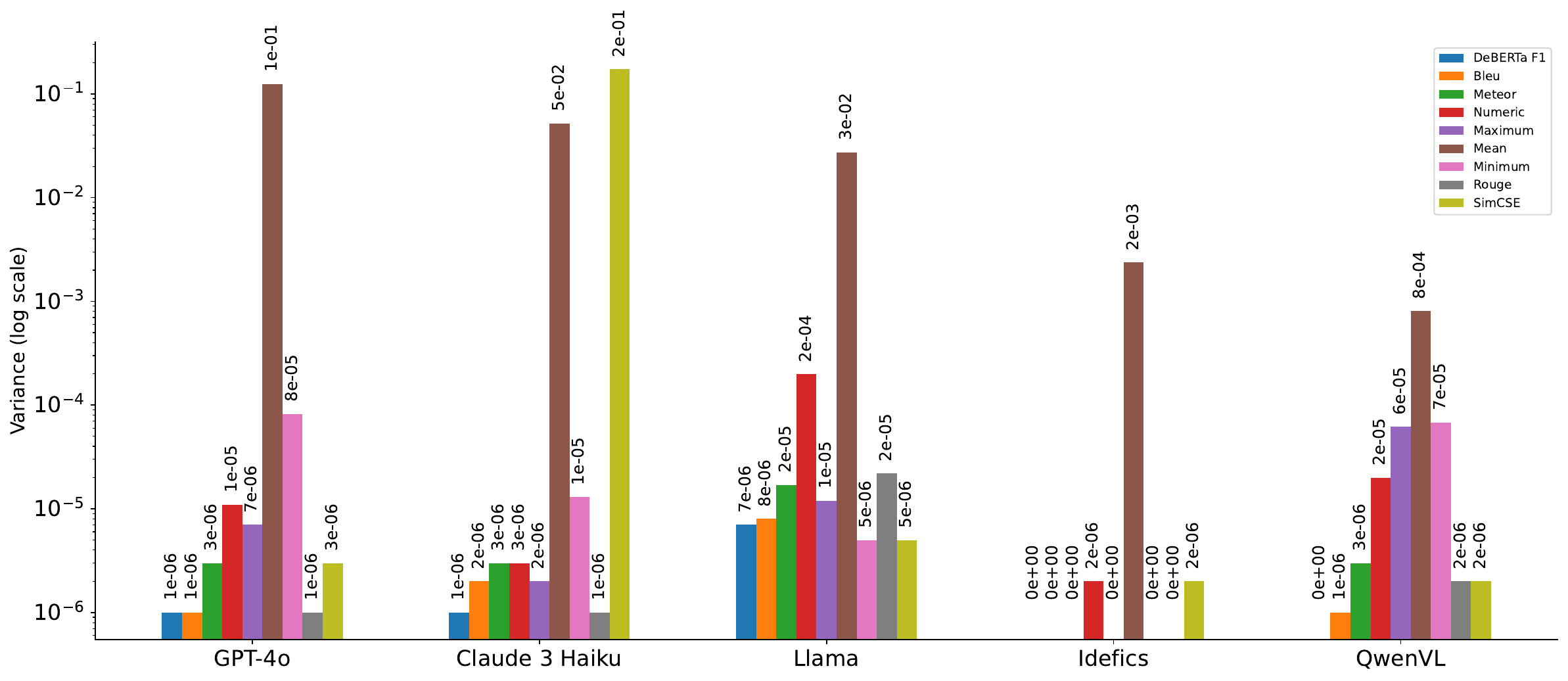}
    \caption{Variance across three independent runs (approximately 500 samples) 
    for each model–metric pair. The logarithmic scale highlights both very small 
    variances ($10^{-6}$) and moderately larger values.}
    \label{fig:variance_plot}
    \vspace{-10pt} 
\end{figure*}

\subsubsection{Latent Similarities} \label{latent sims}

We further assessed whether Gemini-generated captions introduce systematic stylistic or linguistic bias. 
Results across multiple embedding models with different dimensions (Dim) show minimal template reliance and high diversity, consistent with prior work on linguistic convergence between AI and humans~\citep{wilkenfeld2022ai}
We perform $\sim8M$ pairwise comparisons in the $4005$ captions using nine embedding models (Table~\ref{tab:embedding-comparison}). Intra-domain similarity was consistently higher ($0.59-0.78$) than inter-domain similarity ($0.23-0.54$), with large effect sizes (Cohen’s $d > 3.26$). Near-duplicates (cosine $>0.95$) were rare, with an average of $2.3\%$ of pairs. Even within domains, similarity showed non-trivial variance, indicating that \textbf{captions are not rigid templates but semantically varied}. 


\begin{figure*}[ht]
    \centering
    \includegraphics[width=\textwidth, trim=0 0 0 0, clip]{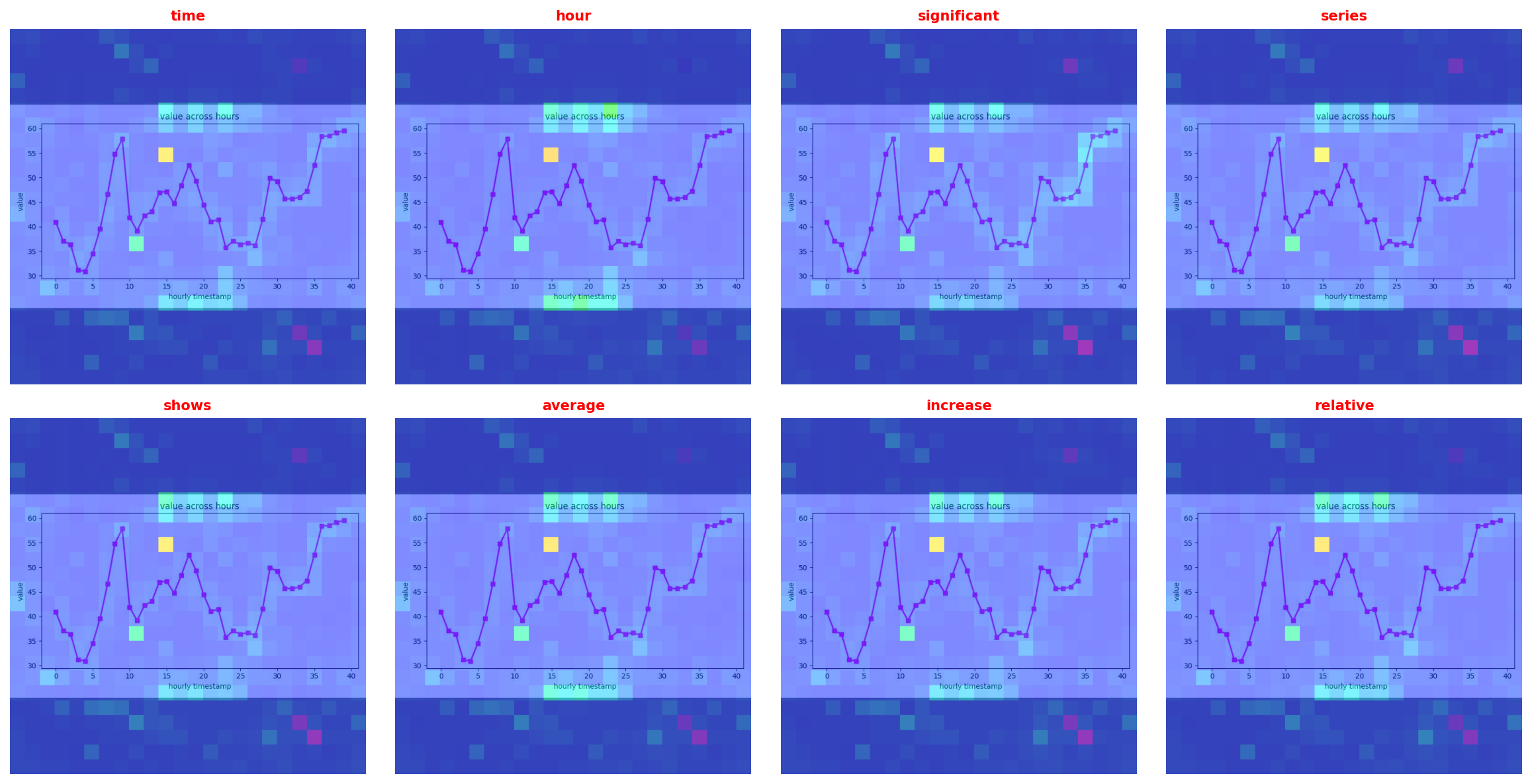}
    \caption{Word-level attention maps for the top 8 tokens from \texttt{LLaVA 1.6 Mistral} overlaid on a time series plot. Despite expectations of alignment with visual trends, attention remains largely diffuse, offering only weak evidence of localized visual grounding in caption generation.
    }
    \label{fig:attention}
\end{figure*}

\subsubsection{Human vs.\ AI Embedding Similarity} \label{ai vs human}

We measure cosine similarity between captions authored by humans, GPT-4o, and Gemini using Qwen-8B embeddings to assess whether AI-generated captions diverge stylistically from human-rewritten references. 
Across both CaTS-Bench and the STOCK dataset, we observe only modest gaps between human-AI and human-human similarity in embedding space, indicating overlap in semantic space despite substantial surface-level differences. This suggests that while AI captions differ lexically and structurally from human captions, they do not collapse to a distinct semantic mode. For CaTS-Bench, we report results on a balanced subset of $500$ human and $500$ AI captions, averaged over five random draws. For STOCK, we evaluate $3k$ human and $3k$ AI captions following prior work. 
Full results are shown in Table~\ref{tab:embedding-similarity}.

\begin{table}[ht]
\caption{Cosine similarities between human, GPT-4o, and Gemini captions using Qwen-8B embeddings.
Human-AI similarity is consistently lower than human-human similarity, but remains semantically comparable.}
\label{tab:embedding-similarity}
\centering
\small
\setlength{\tabcolsep}{12pt}
\renewcommand{\arraystretch}{1.15}

\begin{tabular}{lccc}
\toprule
\rowcolor[gray]{0.95}
Comparison & Human & GPT & Gemini \\
\midrule

\multicolumn{4}{l}{\textbf{CaTS-Bench (500 vs 500)}} \\
\midrule
Human  & 0.389 & 0.341 & 0.336 \\
GPT    &       & 0.414 & 0.321 \\
Gemini &       &       & 0.372 \\
\midrule

\multicolumn{4}{l}{\textbf{STOCK (3k vs 3k)}} \\
\midrule
Human  & 0.658 & 0.621 & 0.603 \\
GPT    &       & 0.692 & 0.588 \\
Gemini &       &       & 0.601 \\
\bottomrule
\end{tabular}
\end{table}


\subsection{Reproducibility Validation}
\label{sec:reproducibility}

We re-ran our evaluation three times on $\sim$500 randomly sampled test examples from the synthetic data across five representative models as generators (GPT-4o, Claude 3 Haiku, LLaMA, Idefics, Qwen-VL). 
Figure~\ref{fig:variance_plot} shows a log-scale visualization. 
\textbf{Across nearly all metrics, the variance is vanishingly small, often on the order of $10^{-6}$, which confirms stability and supports single-run robustness}.

\section{Role of Vision}\label{vision-role}

\subsection{Visual Attention Analysis}\label{visual-attention-analysis}

Interpreting visual grounding in large multimodal models is non-trivial, as not all of them expose interpretable cross-modal attention mechanisms. We attempt this using the LLaVA model, which provides access to decoder-level cross-attention weights over vision tokens. We adapt the approach in \cite{vlm_visualizer} for the \texttt{LLaVA 1.6} model. We visualize per-token visual grounding via the following steps. For each generated token, we extract the decoder cross-attention matrix $\mathbf{A}_{\text{llm}} \in \mathbb{R}^{T \times V}$, where $T$ is the number of generated tokens and $V$ is the number of vision tokens.

Next, we zero out the attention to the beginning-of-sequence token and normalize each row:
\begin{equation}
\tilde{\mathbf{A}}_{\text{llm}}[t, v] = 
\begin{cases}
0, & \text{if } v = \texttt{<bos>} \\
\frac{\mathbf{A}_{\text{llm}}[t, v]}{\sum_{v'} \mathbf{A}_{\text{llm}}[t, v']}, & \text{otherwise}
\end{cases}
\end{equation}

From the CLIP style vision encoder, we extract attention matrices $\mathbf{A}_{\text{vit}}^{(l)} \in \mathbb{R}^{V \times V}$ from multiple layers and average them:

\begin{equation}
\bar{\mathbf{A}}_{\text{vit}} = \frac{1}{L} \sum_{l=1}^{L} \mathbf{A}_{\text{vit}}^{(l)}
\end{equation}

For each token $t$, we compute its attention-weighted vision token distribution and project it back to the 2D image grid, and the projected map $\hat{\mathbf{H}}_t$ is rendered as a heatmap and overlaid on the original image. This allows inspection of which visual regions contribute to each generated token.

\begin{equation}
\hat{\mathbf{H}}_t = \texttt{Upsample}\left(\text{reshape}(\mathbf{H}_t, \text{grid})\right)
\end{equation}

\subsection{Testing Alternative Visual Encodings}
\label{sec:visual-encodings}

\tcbset{
  colback=white, 
  boxrule=0.4pt, 
  arc=1mm,
  fonttitle=\bfseries\color{black}, 
}

\begin{figure}[ht]
  \centering

\begin{tcolorbox}[width=\columnwidth, colback=teal!5, colframe=teal!40, title=Air Quality]
    \begin{subfigure}{0.4\textwidth}
      \includegraphics[width=\linewidth,height=4cm,keepaspectratio]{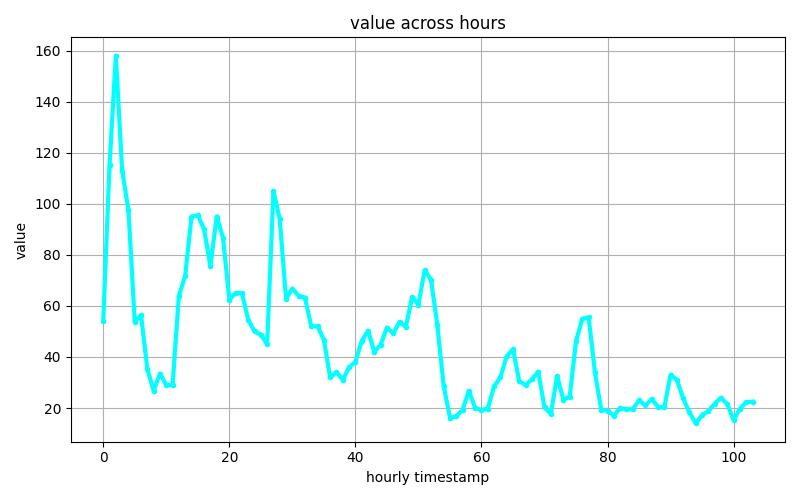}
      \caption*{(a) Line}
    \end{subfigure}%
    \hfill
    \begin{subfigure}{0.275\textwidth}
      \includegraphics[width=\linewidth,height=4.5cm,keepaspectratio]{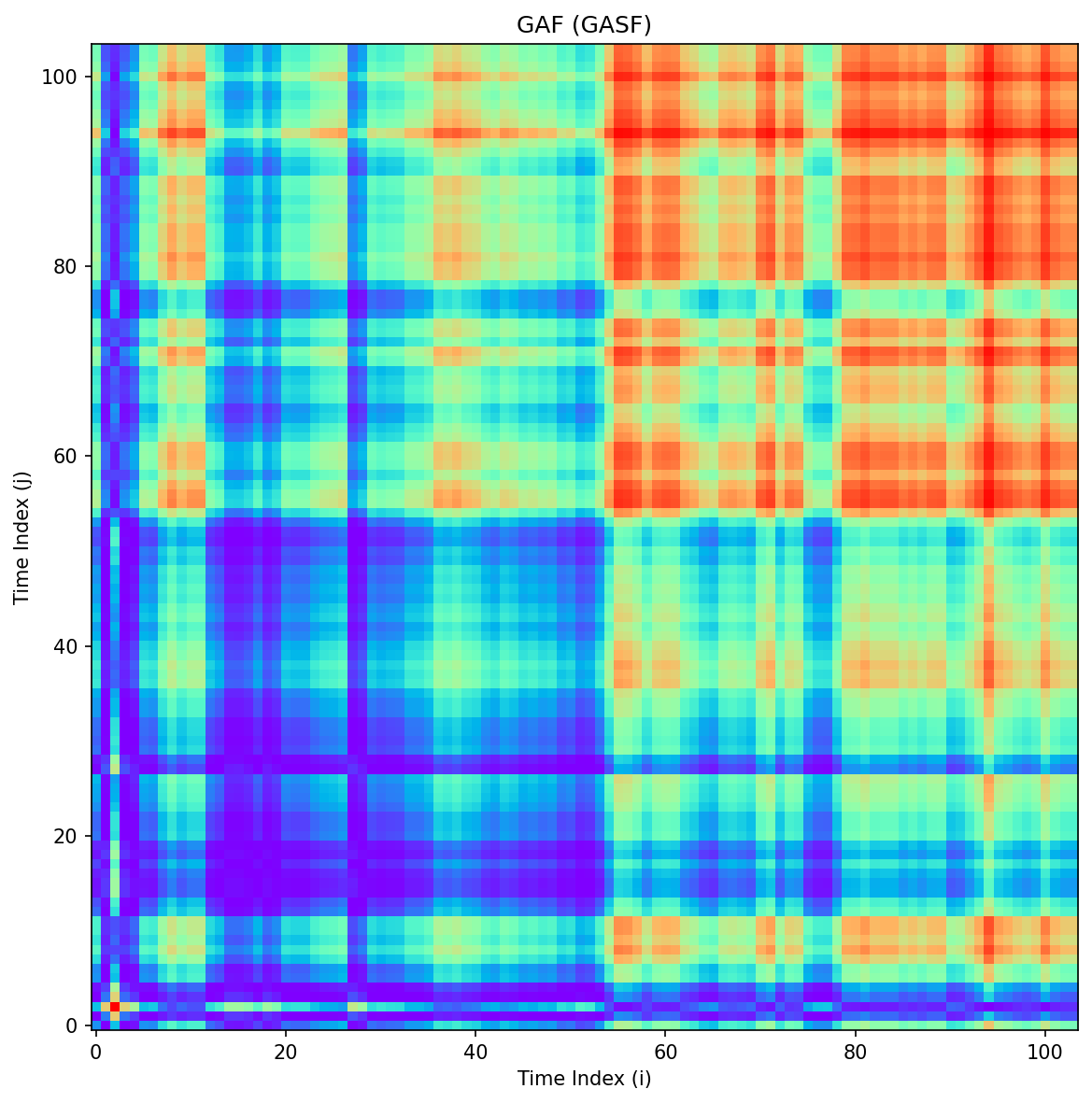}
      \caption*{(b) GAF}
    \end{subfigure}%
    \hfill
    \begin{subfigure}{0.275\textwidth}
      \includegraphics[width=\linewidth,height=4.5cm,keepaspectratio]{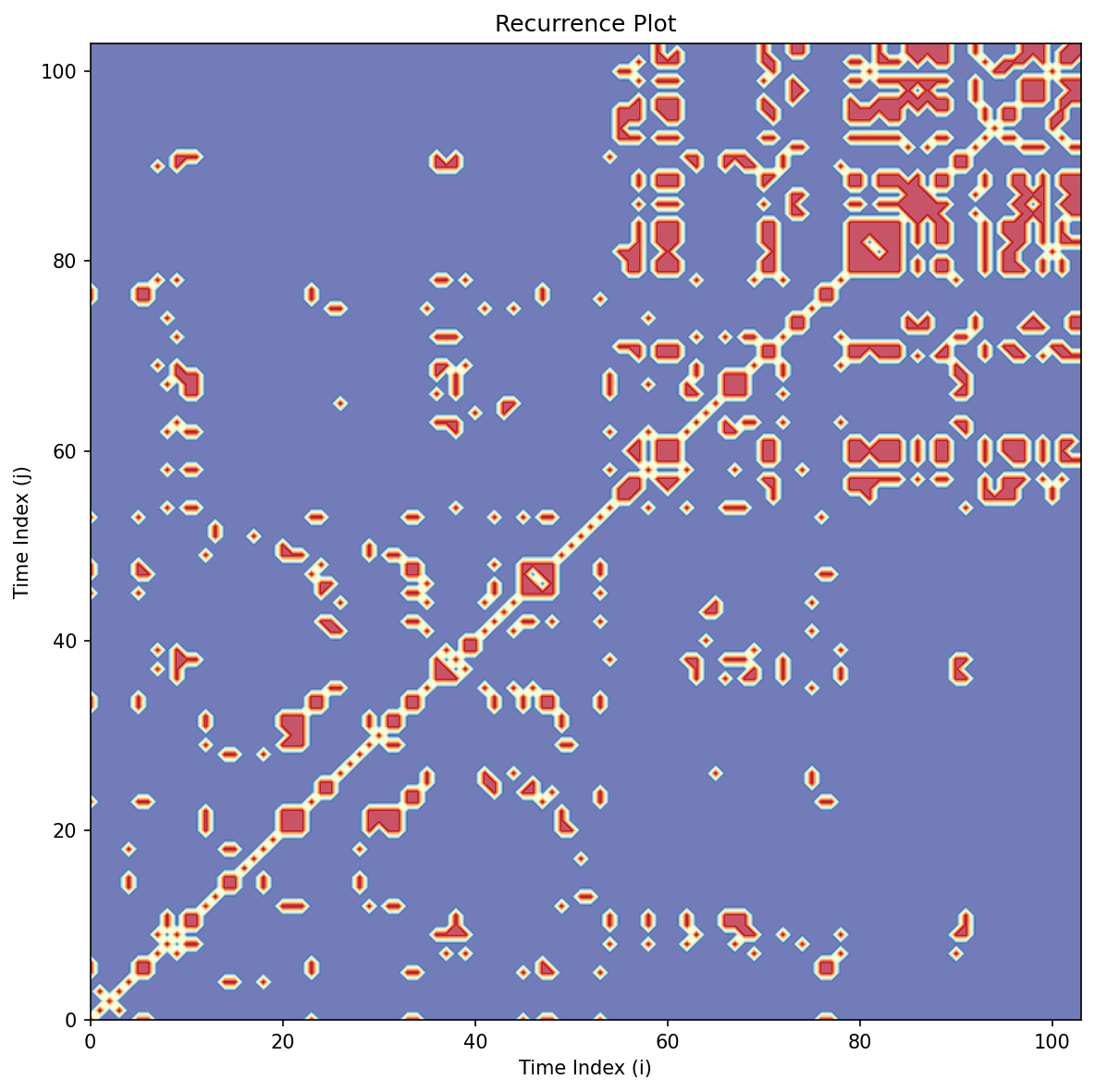}
      \caption*{(c) RP}
    \end{subfigure}
  \end{tcolorbox}

  \begin{tcolorbox}[width=\columnwidth, colback=blue!5, colframe=blue!40, title=Crime]
    \begin{subfigure}{0.4\textwidth}
      \includegraphics[width=\linewidth,height=4cm,keepaspectratio]{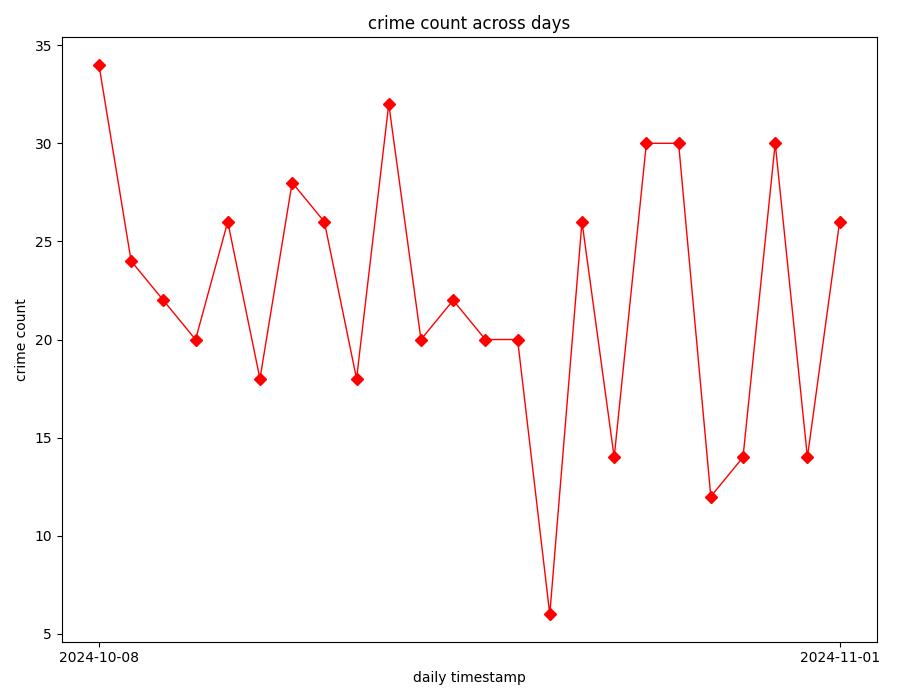}
      \caption*{(a) Line}
    \end{subfigure}%
    \hfill
    \begin{subfigure}{0.275\textwidth}
      \includegraphics[width=\linewidth,height=4.5cm,keepaspectratio]{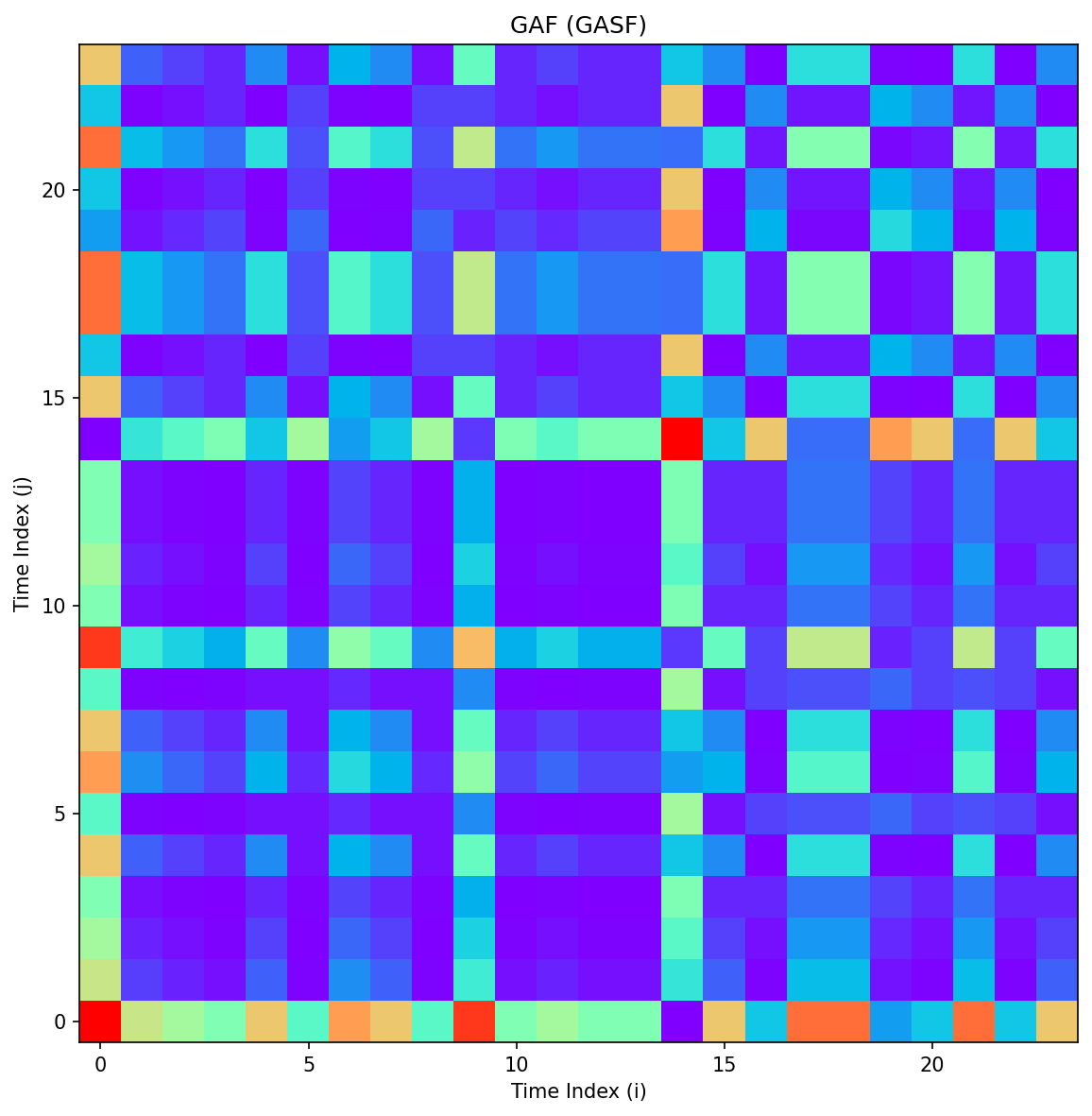}
      \caption*{(b) GAF}
    \end{subfigure}%
    \hfill
    \begin{subfigure}{0.275\textwidth}
      \includegraphics[width=\linewidth,height=4.5cm,keepaspectratio]{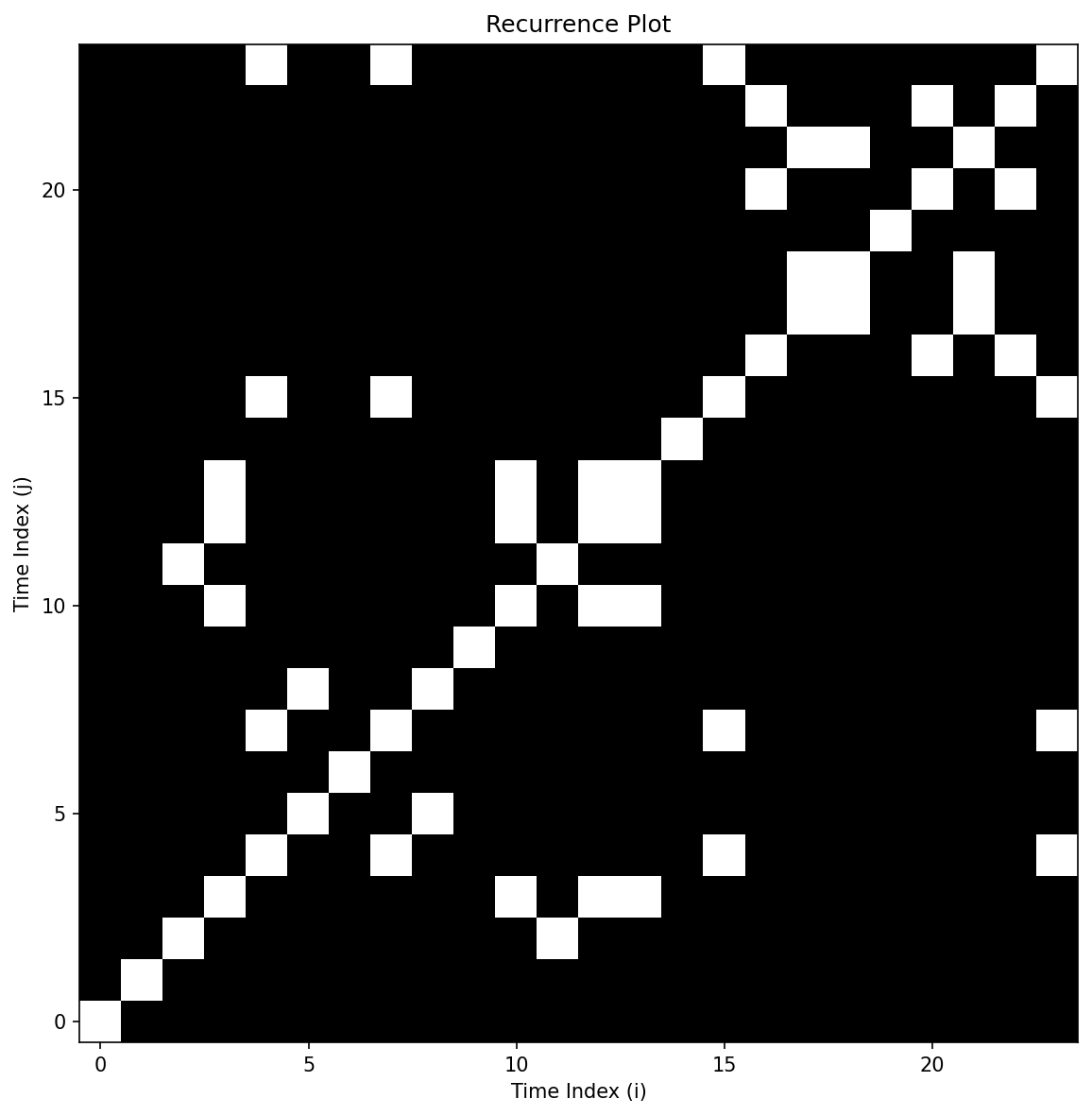}
      \caption*{(c) RP}
    \end{subfigure}
  \end{tcolorbox}

  \begin{tcolorbox}[width=\columnwidth, colback=red!5, colframe=red!40, title=Demography]
    \begin{subfigure}{0.4\textwidth}
      \includegraphics[width=\linewidth,height=4cm,keepaspectratio]{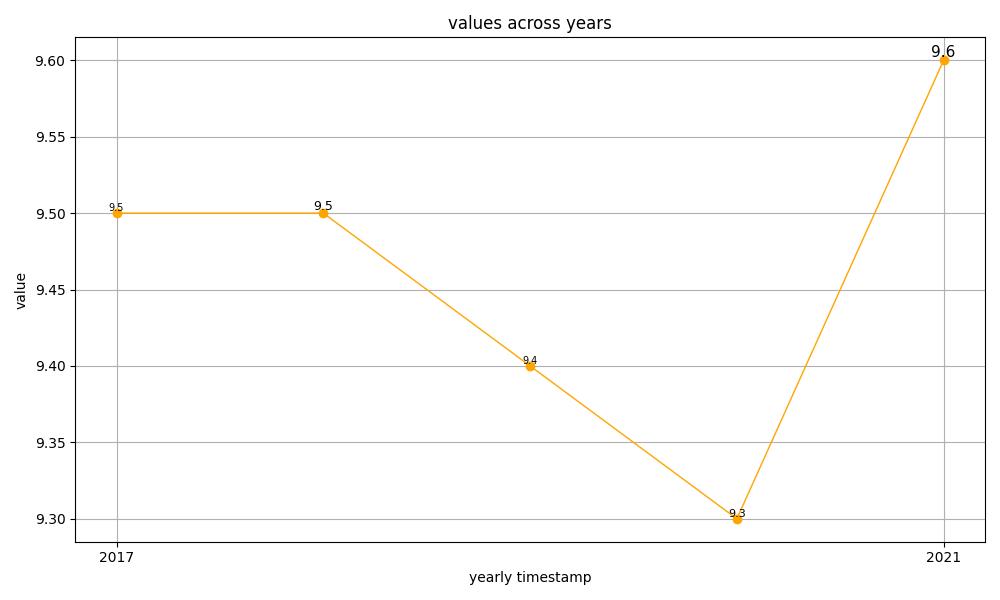}
      \caption*{(a) Line}
    \end{subfigure}%
    \hfill
    \begin{subfigure}{0.275\textwidth}
      \includegraphics[width=\linewidth,height=4.5cm,keepaspectratio]{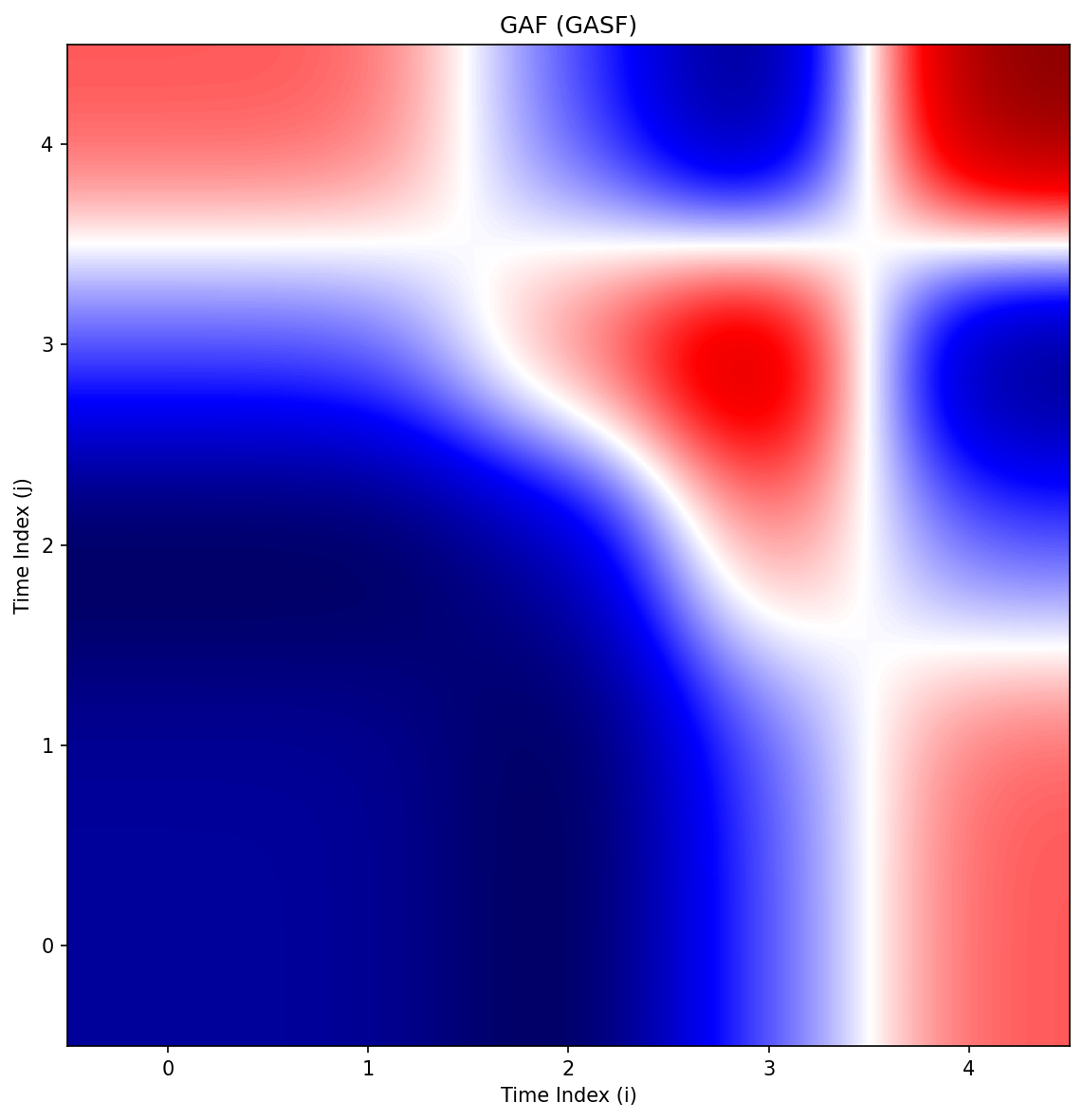}
      \caption*{(b) GAF}
    \end{subfigure}%
    \hfill
    \begin{subfigure}{0.275\textwidth}
      \includegraphics[width=\linewidth,height=4.5cm,keepaspectratio]{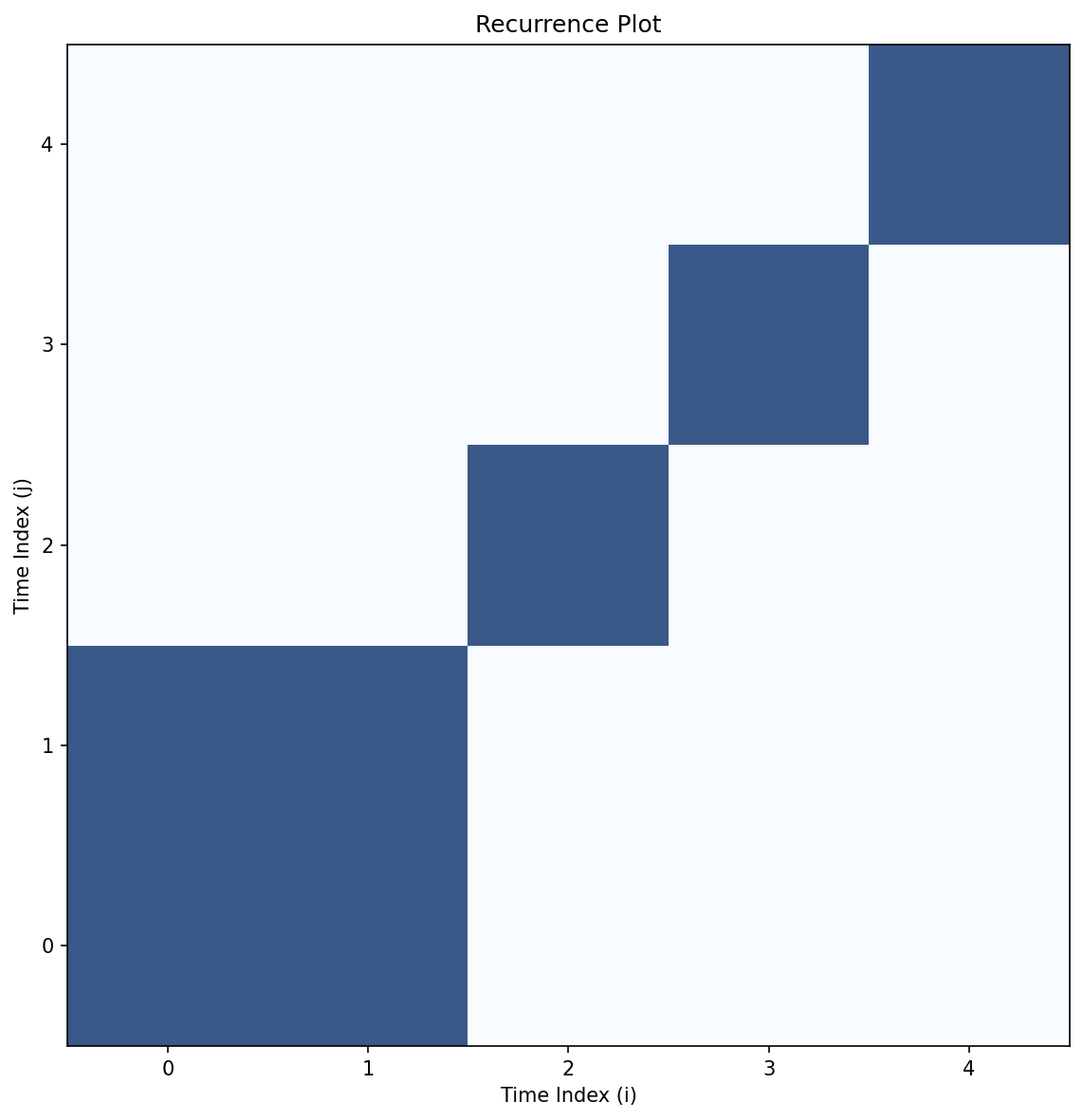}
      \caption*{(c) RP}
    \end{subfigure}
  \end{tcolorbox}

  \caption{Examples of visual encodings across three domains across three sample domains (Air Quality, Crime, Demography) (a) Line, (b) GAF, and (c) RP.}
  \label{fig:visual-encodings-examples}
\end{figure}

To investigate whether more complex visualizations could help in the modality collapse issue, we experimented with Gramian Angular Fields (GAFs) and recurrence plots (RPs) in addition to standard line plots (Figure~\ref{fig:visual-encodings-examples}). Each was generated from the same univariate time series windows, using first-order deltas as input. There are also other encodings, such as multi-series overlays or confidence intervals, but we find these are less applicable in our strictly univariate setting.

\definecolor{lightgreen}{RGB}{220, 255, 220}
\definecolor{lightred}{RGB}{255, 220, 220}

\begin{table}[ht]
\caption{Effect of alternative visual encodings on captioning performance on a subset of synthetic captions. Values show the baseline score for \emph{No Plot (TS+Text)} and relative differences ($\Delta$) for line plots, Gramian Angular Fields (GAFs), and recurrence plots (RPs).}
\label{tab:visual_encodings}
\centering
\small
\setlength{\tabcolsep}{3pt}      
\renewcommand{\arraystretch}{1.0} 

\resizebox{\columnwidth}{!}{
\begin{tabular}{l|l|r|r|r|r}
\toprule
\rowcolor[gray]{0.95}
\textbf{Model} & \textbf{Metric} & \textbf{No Plot} & \textbf{Line ($\Delta$)} & \textbf{GAF ($\Delta$)} & \textbf{RP ($\Delta$)} \\
\midrule
\multirow{11}{*}{\rotatebox{90}{Idefics2 (8B)}}
 & DeBERTa F1 & 0.6255 & \cellcolor{lightred}{-0.0780} & \cellcolor{lightred}{-0.0525} & \cellcolor{lightred}{-0.0525} \\
 & BLEU & 0.0690 & \cellcolor{lightred}{-0.0405} & \cellcolor{lightred}{-0.0460} & \cellcolor{lightred}{-0.0430} \\
 & METEOR & 0.2275 & \cellcolor{lightred}{-0.0925} & \cellcolor{lightred}{-0.0905} & \cellcolor{lightred}{-0.0815} \\
 & NUMERIC & 0.4455 & \cellcolor{lightred}{-0.1665} & \cellcolor{lightred}{-0.1715} & \cellcolor{lightred}{-0.1285} \\
 & MAXIMUM & 0.8080 & \cellcolor{lightgreen}{+0.0730} & \cellcolor{lightred}{-0.1520} & \cellcolor{lightred}{-0.1300} \\
 & MEAN & 0.4195 & \cellcolor{lightgreen}{+0.1285} & \cellcolor{lightgreen}{+0.0945} & \cellcolor{lightgreen}{+0.2995} \\
 & MINIMUM & 0.7590 & \cellcolor{lightred}{-0.0845} & \cellcolor{lightred}{-0.2510} & \cellcolor{lightred}{-0.3940} \\
 & STD & 0.5500 & \cellcolor{lightred}{-0.1835} & \cellcolor{lightred}{-0.1060} & \cellcolor{lightred}{-0.3360} \\
 & ROUGE & 0.2375 & \cellcolor{lightred}{-0.0460} & \cellcolor{lightred}{-0.0465} & \cellcolor{lightred}{-0.0415} \\
 & SIMCSE & 0.7915 & \cellcolor{lightred}{-0.1735} & \cellcolor{lightred}{-0.1935} & \cellcolor{lightred}{-0.1705} \\
\midrule
\multirow{11}{*}{\rotatebox{90}{LLaMA 3.2 Vision}}
 & DeBERTa F1 & 0.6550 & \cellcolor{lightred}{-0.0020} & \cellcolor{lightred}{-0.0030} & \cellcolor{lightred}{-0.0080} \\
 & BLEU & 0.0840 & \cellcolor{lightgreen}{+0.0005} & \cellcolor{lightred}{-0.0090} & \cellcolor{lightred}{-0.0100} \\
 & METEOR & 0.2865 & \cellcolor{lightred}{-0.0080} & \cellcolor{lightred}{-0.0195} & \cellcolor{lightred}{-0.0315} \\
 & NUMERIC & 0.5070 & \cellcolor{lightgreen}{+0.0315} & \cellcolor{lightgreen}{+0.0120} & \cellcolor{lightgreen}{+0.0170} \\
 & MAXIMUM & 0.8611 & \cellcolor{lightgreen}{+0.0584} & \cellcolor{lightred}{-0.0381} & \cellcolor{lightred}{-0.0151} \\
 & MEAN & 0.5395 & \cellcolor{lightgreen}{+0.0690} & \cellcolor{lightgreen}{+0.0755} & \cellcolor{lightgreen}{+0.1215} \\
 & MINIMUM & 0.6670 & \cellcolor{lightgreen}{+0.0820} & \cellcolor{lightgreen}{+0.0740} & \cellcolor{lightgreen}{+0.0670} \\
 & STD & 0.8750 & \cellcolor{lightred}{-0.3270} & \cellcolor{lightred}{-0.2310} & \cellcolor{lightred}{-0.1250} \\
 & ROUGE & 0.2570 & \cellcolor{lightgreen}{+0.0040} & 0.0000 & \cellcolor{lightred}{-0.0040} \\
 & SIMCSE & 0.8265 & \cellcolor{lightred}{-0.0020} & \cellcolor{lightred}{-0.0085} & \cellcolor{lightred}{-0.0235} \\
\midrule
\multirow{11}{*}{\rotatebox{90}{Qwen-VL}}
 & DeBERTa F1 & 0.6315 & \cellcolor{lightred}{-0.0045} & \cellcolor{lightred}{-0.0045} & \cellcolor{lightred}{-0.0095} \\
 & BLEU & 0.0645 & \cellcolor{lightred}{-0.0040} & \cellcolor{lightgreen}{+0.0025} & \cellcolor{lightred}{-0.0015} \\
 & METEOR & 0.2385 & \cellcolor{lightred}{-0.0050} & \cellcolor{lightgreen}{+0.0025} & \cellcolor{lightred}{-0.0025} \\
 & NUMERIC & 0.3555 & \cellcolor{lightgreen}{+0.0560} & \cellcolor{lightred}{-0.0675} & \cellcolor{lightred}{-0.0515} \\
 & MAXIMUM & 0.6610 & \cellcolor{lightgreen}{+0.1135} & \cellcolor{lightgreen}{+0.0200} & \cellcolor{lightgreen}{+0.0430} \\
 & MEAN & 0.4510 & \cellcolor{lightred}{-0.0110} & \cellcolor{lightred}{-0.2150} & \cellcolor{lightred}{-0.2570} \\
 & MINIMUM & 0.5350 & \cellcolor{lightred}{-0.0055} & \cellcolor{lightred}{-0.0650} & \cellcolor{lightred}{-0.0620} \\
 & STD & 0.1070 & \cellcolor{lightgreen}{+0.2095} & \cellcolor{lightred}{-0.0620} & \cellcolor{lightred}{-0.0070} \\
 & ROUGE & 0.2310 & \cellcolor{lightred}{-0.0045} & \cellcolor{lightgreen}{+0.0130} & \cellcolor{lightgreen}{+0.0030} \\
 & SIMCSE & 0.7835 & \cellcolor{lightgreen}{+0.0020} & \cellcolor{lightred}{-0.0085} & \cellcolor{lightred}{-0.0090} \\
\bottomrule
\end{tabular}
}
\end{table}

Table~\ref{tab:visual_encodings} reports results for three representative models (Idefics2-8B, LLaMA 3.2 Vision, Qwen-VL). Values show the baseline metric with no plot (TS+Text only), followed by relative gains or losses for each visualization type. Across models and metrics, neither GAFs nor recurrence plots significantly improved performance; in many cases, they degraded results relative to line plots or even no visual input.  These findings suggest that the bottleneck lies not in the choice of visualization but in the models’ inability to effectively integrate visual cues. With this, we motivate the development of models and encodings that better exploit the structured information available in visual time series.

\section{Bonus TSC Evaluations on Newer Models}
\label{sec:additional-models}
We additionally evaluate some newer proprietary models (\texttt{GPT-5.1}, \texttt{Gemini-3 Flash}) as well as specialized time series language models (\texttt{ChatTS 14B}, \texttt{TimeOmni-1 7B}) on the human-rewritten set of CaTS-Bench. Their results are provided in Table \ref{tab:new_model_metrics}. Comparing them to the models in the main paper (Table \ref{tab:combined-eval}), the overall takeaways are: i) the newest proprietary models do not consistently outperform their older generation counterparts but remain competitive, and ii) time-series-specialized models perform poorly compared to generalist language models, since TSC is still fundamentally a free-form text generation task, though with a taste of numeric and temporal reasoning.

\begin{table*}[t]
\centering
\small
\caption{Performance across metrics of additional models on the HR set.}
\begin{tabular}{l|cccccc|cccc}
\toprule
Model & BERT F1 & SimCSE & BLEU & ROUGE & METEOR & Numeric & Mean & Max & Min & STD \\
\midrule
Gemini 3 Flash  & 0.690 & 0.885 & 0.083 & 0.259 & 0.287 & 0.783 & 0.967 & 0.995 & 0.992 & 0.718 \\
GPT-5.1         & 0.657 & 0.875 & 0.058 & 0.227 & 0.298 & 0.820 & 0.926 & 1.000 & 0.999 & 0.523 \\
ChatTS 14B      & 0.605 & 0.667 & 0.040 & 0.186 & 0.215 & 0.587 & 0.929 & 0.701 & 0.667 & 0.044 \\
TimeOmni-1 7B   & 0.613 & 0.798 & 0.046 & 0.205 & 0.225 & 0.717 & 0.763 & 0.974 & 0.953 & 0.130 \\

\bottomrule
\end{tabular}

\label{tab:new_model_metrics}
\end{table*}

\section{Information Coverage Analysis}
\label{sec:info-cov-analysis}

To better understand the gains from finetuning, we conduct an information coverage analysis comparing pretrained and finetuned \textsc{Idefics2} on $\sim$1.3k generated captions. Using \textsc{GPT-4o} as an automatic annotator (with manual spot-checking), we measure whether captions include core time series reasoning elements such as central tendency, extrema, trends, and comparative reasoning.

\begin{table}[ht]
\centering
\small
\setlength{\tabcolsep}{6pt}
\renewcommand{\arraystretch}{1.1}
\caption{Information coverage of key time series reasoning elements in captions generated by pretrained vs. finetuned models.}
\resizebox{\linewidth}{!}{
\begin{tabular}{l|c|c|c}
\toprule
\rowcolor[gray]{0.95}
\textbf{Category} & \textbf{Pretrained} & \textbf{Finetuned} & \textbf{Gain} \\
\midrule
Central Tendency        & 14.0\% & 95.5\% & +81.6\% \\
Peak/Maxima             & 55.5\% & 84.7\% & +29.2\% \\
Trough/Minima           & 51.7\% & 64.1\% & +12.4\% \\
Trend Direction         & 56.0\% & 74.6\% & +18.6\% \\
Temporal References     & 76.1\% & 100.0\% & +23.9\% \\
Comparative Reasoning   & 6.5\%  & 95.6\% & +89.1\% \\
Variability/Dispersion  & 31.6\% & 88.1\% & +56.6\% \\
Pattern/Shape           & 15.3\% & 67.3\% & +52.0\% \\
\bottomrule
\end{tabular}
}
\label{tab:info_coverage}
\end{table}

As shown in Table~\ref{tab:info_coverage}, finetuning substantially increases the inclusion of essential reasoning components, particularly for central tendency and comparative reasoning. These improvements help explain the gains observed across evaluation metrics, which suggests that finetuning teaches the model to produce more structurally complete and informative time series descriptions.

\section{Inference Examples} \label{inference-samples}
\subsection{Numeric Time Series Ablation}

\begin{figure}[ht]
  \begin{tcolorbox}[
      colback=white,
      colframe=black!40,
      boxrule=0.4pt,
      arc=1mm,
      left=1pt,
      right=1pt,
      top=2pt,
      bottom=2pt,
      width=\columnwidth
    ]
    \scriptsize
    \centering

    \begin{tcolorbox}[
        colback=blue!5,
        colframe=blue!40,
        arc=1mm,
        boxrule=0.3pt,
        width=\textwidth,
        valign=top,
        top=10pt,
        bottom=10pt
      ]
      \includegraphics[width=\textwidth]{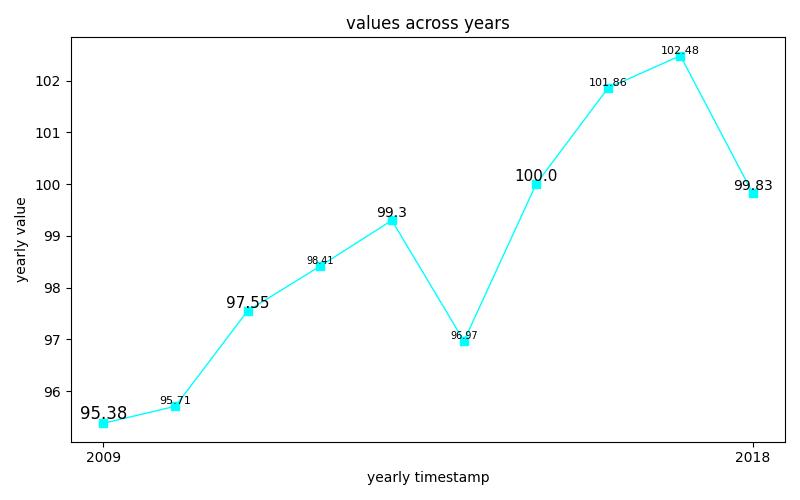}
    \end{tcolorbox}

    \begin{tcolorbox}[
        colback=green!5,
        colframe=yellow!50!black,
        arc=1mm,
        boxrule=0.3pt,
        width=\textwidth,
        valign=top,
        top=2pt,
        bottom=2pt
      ]
      \textbf{Generated Caption (With Numeric Time Series)}\\[0.3em]
      The time series plot shows the yearly aggregated input index (2015=100) in El Salvador from 2009 to 2018. The index comprises land, labor, capital, and materials. The plot shows a general upward trend, with a slight dip in 2016, reaching a peak of 102.48 in 2018.
    \end{tcolorbox}
    
    \begin{tcolorbox}[
        colback=green!5,
        colframe=yellow!50!black,
        arc=1mm,
        boxrule=0.3pt,
        width=\textwidth,
        valign=top,
        top=2pt,
        bottom=2pt
      ]
      \textbf{Generated Caption (Without Numeric Time Series)}\\[0.3em]
      The plot shows the yearly aggregated input index (2015=100) in El Salvador from 2009 to 2018. It displays a general upward trend, rising from 90.5 in 2009 to around 100 in 2010, peaking in 2014, then returning closer to 100.
    \end{tcolorbox}

  \end{tcolorbox}
  \caption{Comparison of generated captions with and without numeric time series input. Model evaluated is \texttt{LLaVA 1.6}. }
  \label{fig:gen_caption_numerics}
\end{figure}

We compare the generated captions with and without explicitly providing the numeric time series data as input to the model in the prompt. As shown in Figure~\ref{fig:gen_caption_numerics}, when a numeric time series is included, the model produces some factual and interpretative errors. Notably, it describes the trend as “increased steadily,” despite the clear dip in 2014 and a decline post-2017. It also incorrectly identifies 2018 as the year of the peak value 102.48, while the actual peak occurs in 2017. Similarly, the slight dip is misattributed to 2016 instead of the correct year 2014. 

In contrast, when the numeric time series is removed from the input, the generated caption becomes significantly more erroneous. The model fabricates plausible-sounding but incorrect values, for example, claiming the index reached 90.5 in 2009 and spiked to 105 in 2014, neither of which is present in the actual plot. This suggests that the absence of explicit numeric context forces the model to hallucinate plausible trajectories based solely on the shape of the line plot. While both versions demonstrate limitations in temporal precision, the numeric-aware caption is more grounded and less prone to hallucinating specific values, producing some factual errors but still outperforming the numeric-agnostic version, which fabricates values entirely.

\subsection{Statistical Inference Failure \& Success}
In the main paper, we mentioned that finetuned models often become overconfident when inferring statistical properties such as means and standard deviations, despite lacking the capability to compute them accurately. In this section, we present two concrete cases that illustrate this overconfidence in practice, and one case where the finetuned model actually successfully inferred the statistics.

\paragraph{Case 1: Failure} The following error case shows the finetuned \texttt{LLaVA v1.6 Mistral} erroneously attempting to infer the time series mean, while its pretrained version does not. We compare the captions generated by both versions of the model against the numerically grounded reference.

\begin{figure}[ht]
  \centering

  \begin{tcolorbox}[
      colback=white,
      colframe=black!40,
      boxrule=0.4pt,
      arc=1mm,
      left=2pt,
      right=2pt,
      top=2pt,
      bottom=1pt,
      width=\columnwidth
    ]
    
    \fontsize{8pt}{9pt}\selectfont
    \centering

    \begin{tcolorbox}[
        colback=green!5,
        colframe=yellow!50!black,
        arc=1mm,
        boxrule=0.3pt,
        width=\textwidth,
        valign=top,
        top=2pt,
        bottom=2pt
      ]
      \textbf{Ground-Truth Caption}\\[0.3em]
      The Total Factor Productivity (TFP) index for Malawi from 2008 to 2017 reveals notable fluctuations, starting at 85.99 in 2008 and peaking at 110.47 in 2017, with 2015 set at 100.0. There's an initial rise from 2008 to 2009, followed by some volatility before a significant jump to 107.82 in 2013, indicating a period of strong productivity growth, and the mean of \textbf{98.93} for this period is significantly higher than the historical mean of \textbf{61.88}, suggesting a period of above-average productivity compared to Malawi's longer-term performance. 

    \end{tcolorbox}

    \begin{tcolorbox}[
        colback=blue!5,
        colframe=blue!40,
        arc=1mm,
        boxrule=0.3pt,
        width=\textwidth,
        valign=top,
        top=2pt,
        bottom=2pt
      ]
      \textbf{Caption from Pretrained Model}\\[0.3em]
      The time series for Malawi's Total Factor Productivity index shows a generally upward trend from 2008 to 2017, with some fluctuations. The index starts at 85.99 in 2008 and reaches a peak of 110.47 in 2014. It then experiences a decline, dropping below 100 in 2015 and 2016. In 2017, the index showed a slight increase, reaching 100.06. Compared to global or regional norms, Malawi's productivity index appears to be lower than expected, suggesting a need for economic development and productivity enhancement strategies. 
    \end{tcolorbox}

    \begin{tcolorbox}[
        colback=red!5,
        colframe=green!40!black,
        arc=1mm,
        boxrule=0.3pt,
        width=\textwidth,
        valign=center,
        top=1pt,
        bottom=1pt
      ]
      \textbf{Caption from Finetuned Model}\\[0.3em]
      From 2008 to 2017, Malawi's Total Factor Productivity index (2015=100) exhibited considerable volatility, starting at 85.99 in 2008 and ending at 110.47 in 2017. The index experienced a notable increase from 2008 to 2010, rising from 85.99 to 95.76, followed by a dip to 91.75 in 2011, before generally trending upwards with some fluctuations. Compared to the historical mean of \textbf{94.84}, the mean of this specific time series is moderately higher at \textbf{99.88}, suggesting a period of relatively stronger productivity growth than the broader historical context.
    \end{tcolorbox}
  \end{tcolorbox}
  \caption{Finetuned model attempts to infer the mean, but it fails. Compare the \textbf{bolden} values.}
  \label{fig:failure_case_1}
\end{figure}

We observe that, while the pretrained model refrains from inferring the mean of the time series, displaying a form of "self-awareness" about its limitations, the finetuned model tends to make confident guesses. This behavior underscores a key limitation of the cross-entropy loss objective. Since ground-truth captions often include the time series mean, the model can reduce its loss by always attempting to predict them, as number tokens are treated like any text tokens. When the model generates captions with sentence structures resembling those in the ground truth, even erroneous guesses of these statistics incur less loss than omitting them entirely.

\paragraph{Case 2: Success} The following is a success case where the finetuned \texttt{Idefics 2} is able to infer the time series mean accurately with a negligible error. We compare the captions generated by the pretrained and finetuned versions of the model against the ground-truth.

Interestingly, the issue of statistical overconfidence appears to be model-specific, as different models exhibit varying behaviors after finetuning. In this case, the finetuned \texttt{Idefics 2} was able to infer both the mean and the standard deviation with reasonable accuracy, when even the ground-truth caption did not explicitly include the standard deviation. This signals that some models benefit more from finetuning on our training data.

\begin{figure}[ht]
  \centering
  \begin{tcolorbox}[
      colback=white,
      colframe=black!40,
      boxrule=0.4pt,
      arc=1mm,
      left=2pt,
      right=2pt,
      top=2pt,
      bottom=1pt,
      width=\columnwidth
    ]
    \fontsize{8pt}{9pt}\selectfont
    \centering

    \begin{tcolorbox}[
        colback=green!5,
        colframe=yellow!50!black,
        arc=1mm,
        boxrule=0.3pt,
        width=\textwidth,
        valign=top,
        top=2pt,
        bottom=2pt
      ]
      \textbf{Ground-Truth Caption}\\[0.3em]
      The death rate in Papua New Guinea from 2017 to 2021 shows a relatively stable trend, fluctuating around a mean of \textbf{6.52} deaths per 1000 people, with a slight dip to 6.4 in 2020 and a peak of 6.68 in 2021. Compared to the global average, which steadily increased from 7.53 in 2017 to 8.57 in 2021, Papua New Guinea's death rate is significantly lower. While the global death rate saw a notable increase, particularly between 2020 and 2021, Papua New Guinea's rate remained relatively consistent, indicating a different trajectory than the global trend.

    \end{tcolorbox}

    \begin{tcolorbox}[
        colback=blue!5,
        colframe=blue!40,
        arc=1mm,
        boxrule=0.3pt,
        width=\textwidth,
        valign=top,
        top=2pt,
        bottom=2pt
      ]
      \textbf{Caption from Pretrained Model}\\[0.3em]
      The yearly death rate in Papua New Guinea has been increasing since 2017. In 2017, the death rate was 6.51, and it has been increasing since then. In 2021, the death rate was 6.68, which is the highest value in the time series.
    \end{tcolorbox}

    \begin{tcolorbox}[
        colback=red!5,
        colframe=green!40!black,
        arc=1mm,
        boxrule=0.3pt,
        width=\textwidth,
        valign=center,
        top=1pt,
        bottom=1pt
      ]
      \textbf{Caption from Finetuned Model}\\[0.3em]
      From 2017 to 2021, the death rate in Papua New Guinea, a lower-middle-income country, fluctuated slightly, ranging from a low of 6.40 in 2020 to a high of 6.68 in 2021. The average death rate during this period was \textbf{6.49}, with a standard deviation of \textbf{0.1}, indicating relatively stable numbers. Compared to the global average death rate, which decreased from 7.57 to 7.16 over the same period, Papua New Guinea's death rate was significantly lower and showed no clear downward trend.
    \end{tcolorbox}
  \end{tcolorbox}
  \caption{Finetuned model successfully infers the mean and standard deviation with negligible error. Compare the \textbf{bolden} values.}
  \label{fig:success_case_1}
\end{figure}

\section{Editing and Review }
\label{sec:interface}

\subsection{Edits made in the captions}
During the editing process conducted by two of the authors, captions were systematically refined to ensure accuracy, clarity, and consistency across the dataset. The following key rules were applied:

\begin{enumerate}[leftmargin=*]
    \item \textbf{Removal of external speculation:} Captions were restricted strictly to information verifiable from the metadata, time series, and plot, avoiding any causal claims or conjecture not grounded in the time series values or provided metadata.
    \item \textbf{Variation in phrasing:} To reduce repetitiveness, sentence openings and phrases were varied rather than uniformly beginning and phrasing the same sentences. 
    \item \textbf{Pattern summarization:} When trends or unique structures (such as V-shaped or monotonic movements, etc.) were clearly visible, they were explicitly noted. 
    \item \textbf{Quantitative grounding:} Values such as maxima, minima, averages, and percentage changes were consistently included when relevant to ensure captions remained data-driven.
    \item \textbf{Consistency with variation:} While maintaining factual accuracy and grounding in the data, captions were intentionally varied in structure and style to avoid monotony and ensure more natural, human-like phrasing across the dataset.
\end{enumerate}

This systematic review process resulted in captions that were both faithful to the underlying data and stylistically coherent across the dataset.

\subsection{Semantic Drift Between Unedited and Edited Captions}

In this section, we show how much the captions were edited quantitatively. Table~\ref{tab:domain_edit_summary} quantifies the extent of human edits across domains using word-level divergence (WER), token-level change, and numeric overlap. The magnitude of rewriting varies substantially by domain. For example, \textit{border crossing}, \textit{diet}, and \textit{Walmart} captions exhibit very large edits, with mean WER values of $0.944$, $0.885$, and $0.939$, respectively, alongside sizeable average token increases of $+28.04$, $+26.88$, and $+21.24$, indicating extensive rewriting and content addition.

\begin{table}[ht]
\centering
\caption{
Domain-wise mean rewrite magnitude (WER), lexical overlap (ROUGE-L),
numeric faithfulness (Numeric Jaccard), and caption length change
($\Delta$ Tokens) between unedited and human-edited captions.
}
\label{tab:domain_edit_summary}
\small
\resizebox{\linewidth}{!}{%
\begin{tabular}{lcccc}
\toprule
\textbf{Domain} &
\textbf{WER} &
\textbf{ROUGE-L} &
\textbf{Numeric Jaccard} &
\textbf{$\Delta$ Tokens} \\
\midrule
Agriculture        & 0.663 & 0.541 & 0.892 & 7.63  \\
Border Crossing    & 0.944 & 0.287 & 0.509 & 28.04 \\
CO$_2$             & 0.733 & 0.452 & 0.674 & 18.76 \\
COVID              & 0.232 & 0.859 & 0.846 & -2.95 \\
Crime              & 0.593 & 0.555 & 0.854 & 0.88  \\
Demography         & 0.504 & 0.637 & 0.956 & 4.67  \\
Diet               & 0.885 & 0.365 & 0.618 & 26.88 \\
Online Retail      & 0.396 & 0.730 & 0.816 & -10.76 \\
Road Injuries      & 0.132 & 0.921 & 0.862 & 2.68  \\
Walmart            & 0.939 & 0.288 & 0.329 & 21.24 \\
\midrule
\textbf{Average} & \textbf{0.602} & \textbf{0.564} & \textbf{0.736} & \textbf{9.71} \\
\bottomrule
\end{tabular}
}
\end{table}

Other domains reflect more moderate editing. \textit{Agriculture} and \textit{CO$_2$} show mean WER values of $0.663$ and $0.733$, with average token increases of +7.63 and +18.76, suggesting partial restructuring rather than complete rewrites. Further, domains such as \textit{COVID}, \textit{road injuries}, and \textit{online retail} exhibit lower rewrite magnitudes, with WER values of $0.232$, $0.132$, and $0.396$, respectively, and small or negative mean token deltas, indicating that many edits involve localized refinements or concision rather than expansion.

Across domains, numeric Jaccard scores remain relatively high in several cases (e.g., $0.956$ for \textit{demography} and $0.892$ for \textit{agriculture}), showing that even when captions are substantially rewritten, numeric content is often preserved. Overall, human editing spans a wide spectrum of intervention levels, ranging from targeted phrasing adjustments to near-complete caption rewrites, depending on domain characteristics and the needs of individual captions.

\begin{figure}[ht]
    \centering
    {\setlength{\fboxsep}{0pt}
    \includegraphics[width=\columnwidth]{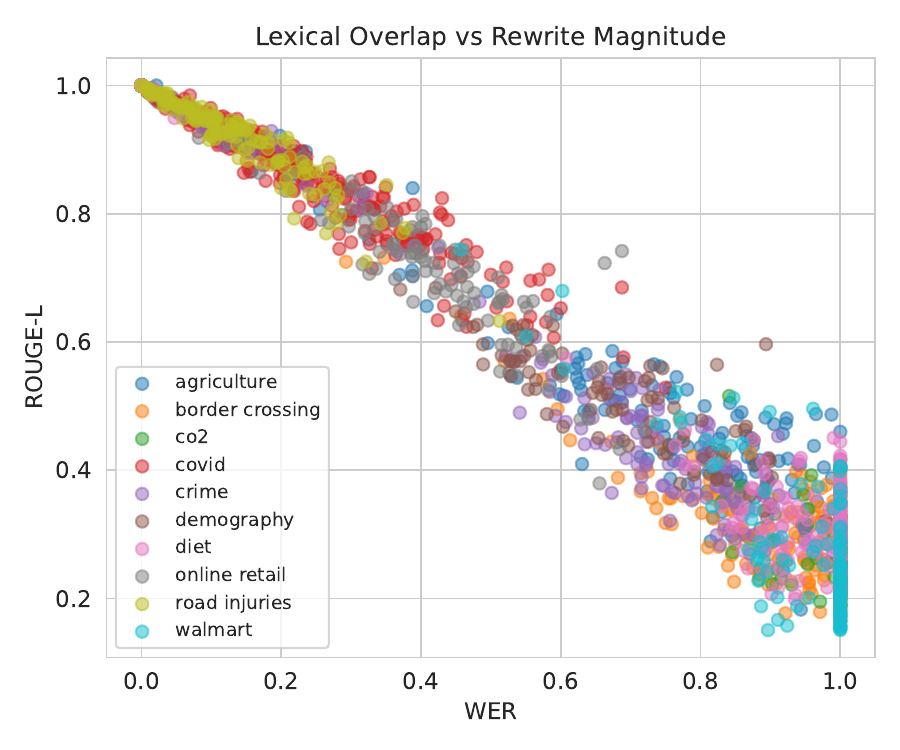}}
    \caption{Lexical overlap versus rewrite magnitude across domains.}
    \label{fig:overall_vs_wer}
\end{figure}

Figure~\ref{fig:overall_vs_wer} plots lexical overlap against rewrite magnitude, where each point corresponds to a single caption pair, with word error rate (WER) between the unedited and human-edited captions on the x-axis and ROUGE-L overlap on the y-axis. Colors denote different domains. We observe a strong inverse relationship between rewrite magnitude and lexical overlap: captions with low WER retain high ROUGE-L scores, while captions undergoing substantial rewriting (WER approaching $1.0$) exhibit markedly lower lexical overlap. This consistent trend across domains indicates that increased human intervention is primarily expressed through lexical and structural rephrasing. The plot further highlights how human edits progressively reduce surface-level overlap as rewrite magnitude increases, reflecting varying degrees of paraphrasing and content restructuring.

\begin{figure}[ht]
    \centering
    {\setlength{\fboxsep}{0pt}
    \includegraphics[width=\columnwidth]{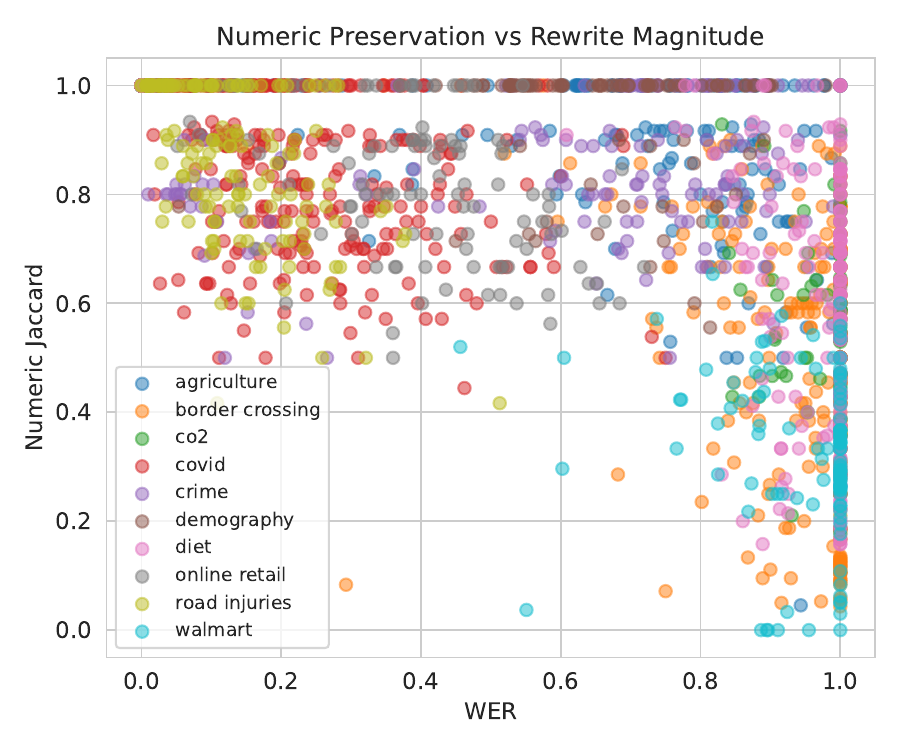}}
    \caption{Numeric preservation versus rewrite magnitude across domains.}
    \label{fig:num_vs_wer}
\end{figure}

Figure~\ref{fig:num_vs_wer} examines numeric preservation under similar conditions. Each point represents a caption pair, plotted by WER on the x-axis and numeric Jaccard similarity on the y-axis, with colors indicating domains. The figure shows that captions span a wide range of rewrite magnitudes, from minor edits (low WER) to near-complete rewrites (WER close to $1.0$). Despite substantial rewriting in many cases, numeric overlap often remains high, with numerous samples exhibiting numeric Jaccard scores above $0.8$ even at high WER values. At the same time, certain domains display greater dispersion at high rewrite magnitudes, reflecting more extensive numeric restructuring when captions are heavily rewritten.

\subsection{Interface}
\begin{figure}[ht]
    \centering
    {\setlength{\fboxsep}{0pt}
    \fbox{\includegraphics[width=\linewidth]{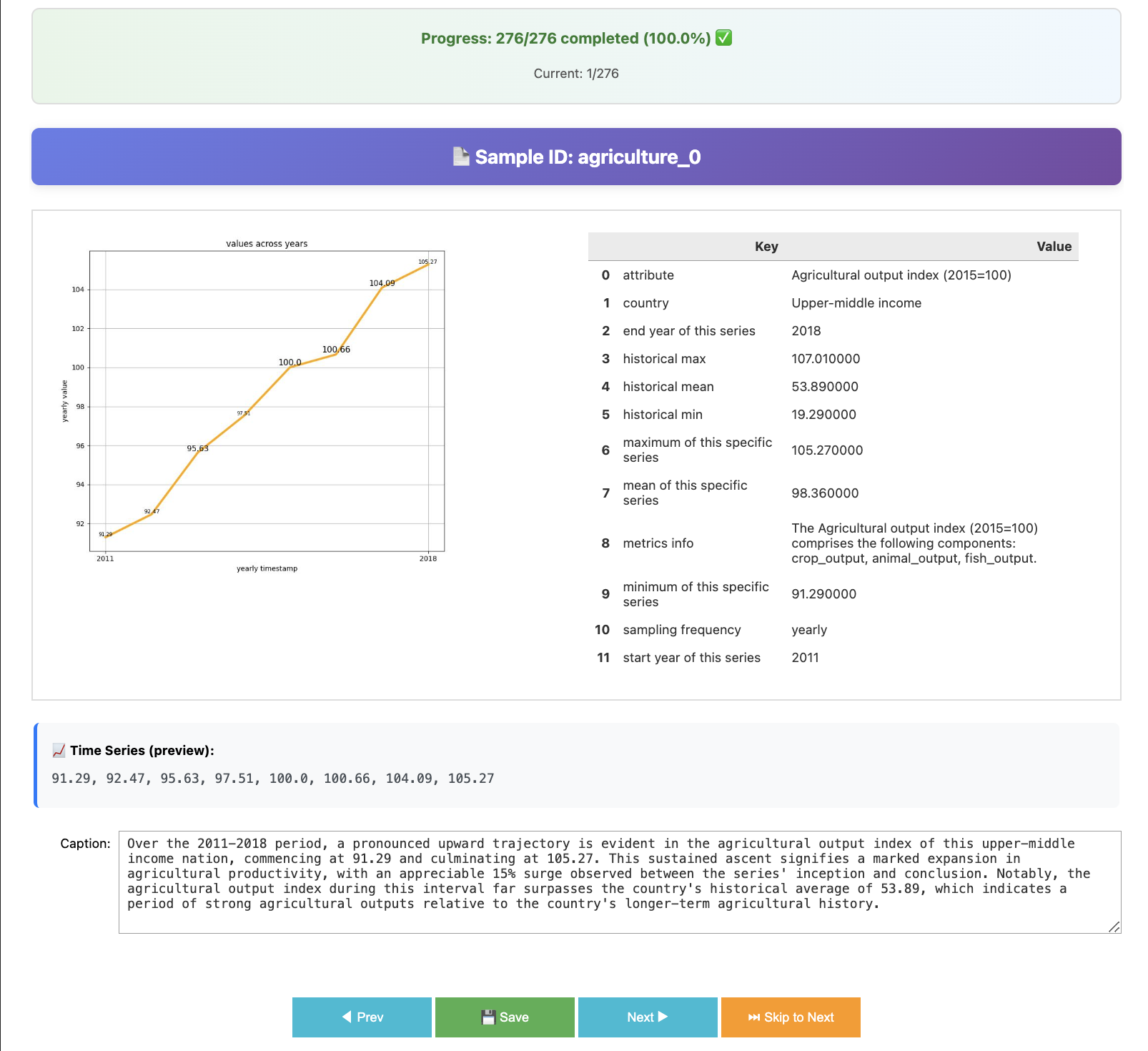}}}
    \caption{Interface used to edit and verify the captions.}
    \label{fig:edit_interface}
\end{figure}

In Figure~\ref{fig:edit_interface}, we provide a screenshot of the editing interface we used to edit the human-rewritten test set.

\section{Template-based Prompts} \label{prompt templates}
In this section, we illustrate the prompts used in our sample generation pipeline, evaluation, paraphrasing, PAL, and distractor generation in Q\&A. Angular brackets are used as placeholders for the actual values.

\subsection{Ground-Truth Synthetic Caption Generation Prompt}\label{gt generation prompt}
The following is an example of a prompt for generating the ground-truth caption from the source dataset \textit{Crime}.

\begin{lstlisting}
Here is a time series about the number of <sampling frequency> crimes in <town>, Los Angeles, from <start date> to <end date>:

<time series>

The all-time statistics of <town> until today are:  
Mean: <general mean of this town>  
Standard Deviation: <general standard deviation of this town>  
Minimum: <general minimum of this town>  
Maximum: <general maximum of this town>

And the statistics for this specific time series are:  
Mean: <mean of this specific series>  
Standard Deviation: <standard deviation of this specific series>  
Minimum: <minimum of this specific series>  
Maximum: <maximum of this specific series>

Describe this time series by focusing on trends and patterns. Discuss concrete numbers you see and pay attention to the dates.

For numeric values, ensure consistency with the provided time series. If making percentage comparisons, round to the nearest whole number. Report the dates when things happened.

Use the statistics I provided you for comparing this example to the normalcy.  
Do not add any extra information beyond what is given.  
Highlight significant spikes, dips, or patterns.

You don't have to explicitly report the numeric values of general statistics; you just use them for reference.  
Compare the trends in this time series to global or regional norms, explaining whether they are higher, lower, or follow expected seasonal patterns.  
When making comparisons, clearly state whether differences are minor, moderate, or significant. 

Use descriptive language to create engaging, natural-sounding text.  
Avoid repetitive phrasing and overused expressions.

Answer in a single paragraph of four sentences at most, without bullet points or any formatting.
\end{lstlisting}

\subsection{Baseline Caption Generation Prompt}\label{caption eval prompt}
When evaluating the baselines on our benchmark, we provide limited metadata, excluding the precomputed statistics of the time series, as the models are expected to infer them on their own. An example of the prompt is the following.

\begin{lstlisting}
Here is a time series about <sampling frequency> <measure> in the Indian city of <city name>. 

The starting time is <starting time> for the series. \n <time series> \n

Describe this time series following these rules:
    
    1. specify and include all contextual data that is needed.
    2. make statistical comparisons (max, min, mean, standard deviation) by providing numerical values.
    3. comment on patterns, shapes, and trends of the time series.
    4. explain how the period compares to historical or global norms.
    5. describe the nature of the movements, patterns, and anomalies if any.
    6. walk through the timeline and describe sub-periods, phases, and periodic patterns when present.
    7. Report dates when events happened.
      
**Not all information for the above rules might be present or needed**, focus on information that is actually evident in the data inputs.
          
Here is some historical context:
 - all-time maximum: <all-time  maximum>
 - all-time minimum: <all-time minimum>
 - all-time average value until today: <all-time average value until today>
 - all-time standard deviation until today: <all-time standard deviation until today>
    
Use descriptive language to create engaging, natural-sounding text.
Avoid repetitive phrasing and overused expressions.

Answer in a single paragraph of four to five sentences at most, without bullet points or any formatting.
\end{lstlisting}

\subsection{Caption Paraphrasing Prompt}\label{caption paraphrasing prompt}
To rephrase a caption into a different linguistic style while preserving semantic and numeric information, we feed the following prompt into a paraphraser model of choice.

\begin{lstlisting}
You are a helpful assistant. Your task is to rephrase the following paragraph that describes a time series. You MUST strictly follow these rules:

Preserve all factual information: All numeric values, statistics (min, max, mean, etc.), trends ('increased', 'peaked'), comparisons ('higher than'), and dates must remain exactly the same.

Change the style completely: Use different sentence structures, synonyms, and grammatical constructions. Alter the tone (e.g., make it more formal or more conversational). Do not use the same phrasing as the original.

Output only the rephrased paragraph, with no additional explanation.

Here is the paragraph to rephrase: 
<caption>
\end{lstlisting}

\subsection{PAL Prompt}\label{PAL_prompt}
Here we reproduce the full prompt used for the program-aided context:
\begin{lstlisting}
### Task
<caption_prompt>

### Instructions for the assistant
1. You are an expert coding assistant; think through the task **step-by-step**.
2. Write **Python 3.12** code (inside one ```python``` block) that computes the final answer.
   * Use only the Python Standard Library (e.g., you may use the `math`, `statistics` libraries).
   * Wrap everything in a `solve()` function that will be invoked to produce the final caption.
   * The code **must produce the caption string itself**. Any numeric values can be computed
     in Python and formatted into the caption string. Make sure to use any values you compute
     in the resulting caption string.
3. The `solve()` function you write will be invoked to produce the final caption.

### Output format (exactly; no extra text, explanations, or formatting)
```python
# code that defines solve() and any desired strings
solve()
```
\end{lstlisting}
The full TSC prompt from~\ref{caption eval prompt} is injected as the \texttt{caption\_prompt} string.

\subsection{Semantic Perturbation Prompt}\label{semantic perturbation prompt}
To perturb a caption so that its semantic meaning is altered while keeping numbers intact, we feed the following prompt into \texttt{Gemini 2.0 Flash}.

\begin{lstlisting}
Your task is to minimally modify a time series description so that its meaning is altered but the numbers are maintained. 

For example, you can switch "increase" with "decrease", "upward" to "downward", or something more sophisticated. Keep the description structurally identical to the original text; you don't have to alter too much information. Altering anywhere between 1 and 3 parts is enough. Do not edit the numbers.
    
Here's the description to modify:
<caption>
    
Give your answer in a paragraph of text as the given description, without any explanation or formatting.
\end{lstlisting}

\subsection{Numeric Perturbation Prompt}\label{numeric perturbation prompt}
To perturb a caption so that its numbers are altered while its semantic information is preserved, we feed the following prompt into \texttt{Gemini 2.0 Flash}.

\begin{lstlisting}
Your task is to slightly modify the numbers in a time series description so that its semantics remain the same, but the numbers are slightly altered. 

For example, you can replace "12" with "12.2", "45%" with "46%". Keep the description structurally and semantically identical to the original text; you don't have to alter all numbers but anywhere between 1 to 3 times is enough. Make sure that the altered number still makes sense and fits the scale of the phenomenon.
    
Here's the description to modify:

<caption>
    
Give your answer in a paragraph of text as the given description, without any explanation and formatting.
\end{lstlisting}

\begin{table}[t]
\centering
\caption{Inter-rater ranking correlation (Spearman $\rho$) across three extractors ($p < 0.001$ for all values).}
\label{tab:correlation_combined}
\renewcommand{\arraystretch}{1.2} 

\begin{subtable}{\columnwidth}
    \centering
    \caption{Minimum value inference accuracy}
    \begin{tabular}{l S[table-format=1.4] S[table-format=1.4] S[table-format=1.4]}
    \toprule
    \textbf{Extractor} & \textbf{GPT-4o} & \textbf{C3H} & \textbf{G2.0F} \\
    \midrule
    GPT-4o           & 1.0000 & 0.8643 & 0.9643 \\
    Claude 3 Haiku   & 0.8643 & 1.0000 & 0.8464 \\
    Gemini 2.0 Flash & 0.9643 & 0.8464 & 1.0000 \\
    \bottomrule
    \end{tabular}
\end{subtable}

\vspace{1em} 

\begin{subtable}{\columnwidth}
    \centering
    \caption{Maximum value inference accuracy}
    \begin{tabular}{l S[table-format=1.4] S[table-format=1.4] S[table-format=1.4]}
    \toprule
    \textbf{Extractor} & \textbf{GPT-4o} & \textbf{C3H} & \textbf{G2.0F} \\
    \midrule
    GPT-4o           & 1.0000 & 0.8500 & 0.9214 \\
    Claude 3 Haiku   & 0.8500 & 1.0000 & 0.8107 \\
    Gemini 2.0 Flash & 0.9214 & 0.8107 & 1.0000 \\
    \bottomrule
    \end{tabular}
\end{subtable}

\vspace{1em} 

\begin{subtable}{\columnwidth}
    \centering
    \caption{Mean value inference accuracy}
    \begin{tabular}{l S[table-format=1.4] S[table-format=1.4] S[table-format=1.4]}
    \toprule
    \textbf{Extractor} & \textbf{GPT-4o} & \textbf{C3H} & \textbf{G2.0F} \\
    \midrule
    GPT-4o           & 1.0000 & 0.9643 & 0.9893 \\
    Claude 3 Haiku   & 0.9643 & 1.0000 & 0.9750 \\
    Gemini 2.0 Flash & 0.9893 & 0.9750 & 1.0000 \\
    \bottomrule
    \end{tabular}
\end{subtable}

\vspace{1em} 

\begin{subtable}{\columnwidth}
    \centering
    \caption{Std. Deviation inference accuracy}
    \begin{tabular}{l S[table-format=1.4] S[table-format=1.4] S[table-format=1.4]}
    \toprule
    \textbf{Extractor} & \textbf{GPT-4o} & \textbf{C3H} & \textbf{G2.0F} \\
    \midrule
    GPT-4o           & 1.0000 & 0.9571 & 0.7536 \\
    Claude 3 Haiku   & 0.9571 & 1.0000 & 0.8286 \\
    Gemini 2.0 Flash & 0.7536 & 0.8286 & 1.0000 \\
    \bottomrule
    \end{tabular}
\end{subtable}

\vspace{0.5em}
\raggedright \footnotesize \textit{Note: C3H = Claude 3 Haiku, G2.0F = Gemini 2.0 Flash.}
\label{tabs:stat_inf_robustness}
\end{table}

\subsection{Numeric Parsing\label{parsing prompt}} 
To compute the statistical inference scores (min, max, mean, STD), we employ an LLM (\textit{Gemini 2.0 Flash}) to extract the statistical claims from the captions. The extracted values will be compared against ground-truth precomputed values from the metadata. A statistical value is correct if it's within 5\% error compared to the true value. The statistics extraction prompt is shown below.  

\begin{lstlisting}
I will provide you with a paragraph of text, which described a time series. Your job is to extract important numbers from it. Specifically, you have to extract minimum, maximum, mean, and standard deviation of this specific time series (not historical statistics) from the text if the text mentions them. You must ignore any statistic about the general history or all-time. For example, you should ignore statistics like "historical maximum", "all-time mean", or something like that. I only need you to extract statistics of the current specific time series.

    Here's the text:
    \n
    {caption}
    \n

    Provide your answer in a JSON format as follows:
   {{
      "minimum": A,
      "maximum": B,
      "mean": X,
      "std": Y

   }}

   IMPORTANT: For each statistic (minimum, maximum, mean, std):
   - If the text DOES NOT mention or attempt to report that statistic at all, set its value to null
   - If the text DOES mention or attempt to report that statistic (even if the value is uncertain or approximate), provide the numerical value

   The key distinction is whether the caption attempts to report the statistic or not. If it attempts to report it, extract the number. If it doesn't attempt to report it at all, use null.

   Provide your answer in a json format and do not say anything more and don't give any explanation. Start your answer directly by opening a brace.
  
  """
\end{lstlisting}

Furthermore, to prove that using a single LLM as the extractor is robust enough, we leveraged two additional LLMs, \textit{GPT-4o} and \textit{Claude 3 Haiku}, as extractors. We show that the choice of extraction model does not significantly impact the evaluation outcomes across all four statistical inference metrics. For pairwise ranking correlations, we observed strong to excellent agreement: the minimum metric averaged $\rho = 0.8917$ (range: $0.8464–0.9643$), the maximum metric averaged $\rho = 0.8607$ (range: $0.8107–0.9214$), the mean metric averaged $\rho = 0.9762$ (range: $0.9643–0.9893$), and the standard deviation metric averaged $\rho = 0.8464$ (range: $0.7536–0.9571$). All pairwise correlations were statistically significant ($p < 0.001$). Critically, when examining the mean metric, which measures accuracy in reporting the average value of time series, all three extractors exhibited $100\%$ agreement on the top-5 performing models (\textit{qwenvl\_pal}, \textit{OpenAI GPT-4o}, \textit{Gemini 2.0 Flash}, \textit{InternVL-38B}, and \textit{Claude 3 Haiku}). Similarly, for the standard deviation metric, all three extractors showed $100\%$ agreement on the top-3 models. These results validate the inter-rater reliability of our statistical inference evaluation methodology across all statistical claims and confirm that our findings are robust and not artifacts of extractor-specific biases. Specific correlations are illustrated in Tables \ref{tabs:stat_inf_robustness}.

\section{Human Baseline\label{human baseline}}

To establish a human performance baseline, we invited university students to complete four Q\&A tasks voluntarily. These tasks span a range of reasoning types, including fine-grained statistical comparisons, semantic interpretation, and multimodal alignment. Participants were recruited through academic networks and completed the tasks without the aid of external tools, ensuring a fair comparison with models operating under similar conditions. Participation was entirely voluntary, with no compensation, and individuals could withdraw at any time. 
Below, we present the instructions given to the volunteers for their participation.

\lstset{
  basicstyle=\ttfamily,
  breaklines=true,
  frame=single,
  columns=fullflexible,
}

\begin{lstlisting}[language=]
Participant Information and Consent Form for Time Series Q\&A Questionnaire

Thank you for considering participation in our study!

This questionnaire is part of a research project evaluating human performance on time series understanding tasks. Your responses will help us establish a baseline for comparing human performance to that of current language models. You will be given a Google Form consisting of 10 to 14 multiple-choice questions of the same type, and you should not use any external tools.

Please read the following information carefully before continuing:

Voluntary Participation: Your participation is entirely voluntary. You may choose not to participate or to withdraw at any time without any consequences.

Duration: The questionnaire is brief and is estimated to take between 3 and 6 minutes to complete.

Anonymity & Data Use: No personal information will be collected or stored. Your answers will remain anonymous and will be used solely for research purposes, such as evaluating and reporting model performance in academic publications.

No Compensation: There is no monetary or material compensation for participating in this study.

Confidentiality: All collected data will be handled securely. Only aggregated and anonymized results will be published.

By proceeding, you confirm that you understand the above terms and agree to participate in this research study.

Thank you for your collaboration and contribution to our research!

Date: __________________           Signature:  __________________
\end{lstlisting}

\subsection{Attempt Analysis in Numeric Metrics}
\label{sec:attempt-analysis}

We provide a detailed diagnostic analysis of model behavior across the four numeric inference tasks: mean, minimum (Min), maximum (Max), and standard deviation (STD). We analyze each statistic independently to expose distinct reasoning behaviors, robustness patterns, and failure modes. All observations are grounded in successes, attempts, and omissions (Null) of these statistics. 

Specifically, \textbf{success rate} is defined as the proportion of correctly reported statistical claims among attempts. To illustrate, if the model only attempted to report the time series mean once across the entire test set, and the mean is correct, the success rate would be $1$. To account for this flaw, the \textbf{attempt rate} is defined as the proportion of times where the model does make the statistical claim, whether it is correct or not. By multiplying the success rate by the attempt rate, we obtain the proportion of correct statistical claims over all evaluation instances. Finally, we count the statistical omissions (Null) where the model did not attempt the statistical claim whatsoever.

\begin{table}[ht]
\caption{Mean inference performance across domains and models.}
\centering
\resizebox{\columnwidth}{!}{
\begin{tabular}{l l c c c}
\toprule
\textbf{Domain} & \textbf{Model} & \textbf{Success} & \textbf{Attempt} & \textbf{Null} \\
\midrule

\multirow{4}{*}{Agriculture}
& Gemini 2.0 Flash & 0.9940 & 1.0000 & 0 \\
& GPT-4o & 0.9940 & 1.0000 & 0 \\
& QwenVL & 0.4850 & 0.9581 & 7 \\
& QwenVL PAL & 0.9701 & 0.9701 & 5 \\
\midrule

\multirow{4}{*}{\shortstack{Border\\Crossing}}
& Gemini 2.0 Flash & 0.7709 & 0.9907 & 2 \\
& GPT-4o & 0.8359 & 0.9938 & 1 \\
& QwenVL & 0.2879 & 0.9876 & 4 \\
& QwenVL PAL & 0.9659 & 0.9721 & 9 \\
\midrule

\multirow{4}{*}{COVID}
& Gemini 2.0 Flash & 0.5167 & 0.9700 & 9 \\
& GPT-4o & 0.5367 & 1.0000 & 0 \\
& QwenVL & 0.1200 & 0.9233 & 23 \\
& QwenVL PAL & 0.8400 & 0.9833 & 5 \\
\midrule

\multirow{4}{*}{Crime}
& Gemini 2.0 Flash & 0.8235 & 1.0000 & 0 \\
& GPT-4o & 0.9281 & 1.0000 & 0 \\
& QwenVL & 0.2092 & 0.9673 & 5 \\
& QwenVL PAL & 0.9804 & 0.9804 & 3 \\
\midrule

\multirow{4}{*}{Demography}
& Gemini 2.0 Flash & 0.9917 & 1.0000 & 0 \\
& GPT-4o & 1.0000 & 1.0000 & 0 \\
& QwenVL & 0.6000 & 0.6333 & 44 \\
& QwenVL PAL & 0.9500 & 0.9500 & 6 \\
\midrule

\multirow{4}{*}{Diet}
& Gemini 2.0 Flash & 0.9763 & 0.9858 & 3 \\
& GPT-4o & 0.9858 & 0.9858 & 3 \\
& QwenVL & 0.6066 & 0.9005 & 21 \\
& QwenVL PAL & 0.9242 & 0.9336 & 14 \\
\midrule

\multirow{4}{*}{\shortstack{Road\\Injuries}}
& Gemini 2.0 Flash & 0.9868 & 0.9868 & 2 \\
& GPT-4o & 1.0000 & 1.0000 & 0 \\
& QwenVL & 0.3816 & 0.9145 & 13 \\
& QwenVL PAL & 0.9474 & 0.9539 & 7 \\
\midrule

\multirow{4}{*}{Walmart}
& Gemini 2.0 Flash & 1.0000 & 1.0000 & 0 \\
& GPT-4o & 0.9817 & 1.0000 & 0 \\
& QwenVL & 0.4128 & 0.7431 & 28 \\
& QwenVL PAL & 0.9908 & 0.9908 & 1 \\

\bottomrule
\end{tabular}
}
\label{tab:mean_inference}
\end{table}

\begin{table}[ht]
\caption{Minimum inference performance across domains and models.}
\centering
\resizebox{\linewidth}{!}{
\begin{tabular}{l l c c c}
\toprule
\textbf{Domain} & \textbf{Model} & \textbf{Success} & \textbf{Attempt} & \textbf{Null} \\
\midrule

\multirow{4}{*}{Agriculture}
 & Gemini 2.0 Flash & 0.9641 & 0.9641 & 6 \\
 & GPT-4o           & 0.9940 & 0.9940 & 1 \\
 & QwenVL           & 0.3533 & 0.8503 & 25 \\
 & QwenVL PAL       & 0.9641 & 0.9641 & 6 \\
\midrule

\multirow{4}{*}{\shortstack{Border \\ Crossing}}
 & Gemini 2.0 Flash & 0.9721 & 0.9907 & 2 \\
 & GPT-4o           & 0.9876 & 0.9969 & 0 \\
 & QwenVL           & 0.3839 & 0.9505 & 16 \\
 & QwenVL PAL       & 0.9505 & 0.9876 & 4 \\
\midrule

\multirow{4}{*}{COVID}
 & Gemini 2.0 Flash & 0.9000 & 0.9600 & 12 \\
 & GPT-4o           & 0.9567 & 0.9933 & 2 \\
 & QwenVL           & 0.3533 & 0.8133 & 56 \\
 & QwenVL PAL       & 0.9633 & 0.9800 & 6 \\
\midrule

\multirow{4}{*}{Crime}
 & Gemini 2.0 Flash & 0.9869 & 0.9935 & 1 \\
 & GPT-4o           & 1.0000 & 1.0000 & 0 \\
 & QwenVL           & 0.1765 & 0.9804 & 3 \\
 & QwenVL PAL       & 0.9804 & 0.9869 & 2 \\
\midrule

\multirow{4}{*}{Demography}
 & Gemini 2.0 Flash & 0.9667 & 0.9667 & 4 \\
 & GPT-4o           & 0.9917 & 0.9917 & 1 \\
 & QwenVL           & 0.5083 & 0.5083 & 59 \\
 & QwenVL PAL       & 0.9333 & 0.9417 & 7 \\
\midrule

\multirow{4}{*}{Diet}
 & Gemini 2.0 Flash & 0.9668 & 0.9763 & 5 \\
 & GPT-4o           & 0.9763 & 0.9763 & 5 \\
 & QwenVL           & 0.5261 & 0.8863 & 24 \\
 & QwenVL PAL       & 0.9289 & 0.9526 & 10 \\
\midrule

\multirow{4}{*}{\shortstack{Road \\ Injuries}}
 & Gemini 2.0 Flash & 1.0000 & 1.0000 & 0 \\
 & GPT-4o           & 1.0000 & 1.0000 & 0 \\
 & QwenVL           & 0.6382 & 0.7829 & 33 \\
 & QwenVL PAL       & 0.9605 & 0.9605 & 6 \\
\midrule

\multirow{4}{*}{Walmart}
 & Gemini 2.0 Flash & 0.9817 & 0.9817 & 2 \\
 & GPT-4o           & 1.0000 & 1.0000 & 0 \\
 & QwenVL           & 0.6606 & 0.9908 & 1 \\
 & QwenVL PAL       & 0.9908 & 0.9908 & 1 \\

\bottomrule
\end{tabular}
}
\label{tab:min_inference}
\end{table}

\begin{table}[ht]
\caption{Maximum inference performance across domains and models.}
\centering
\resizebox{\linewidth}{!}{
\begin{tabular}{l l c c c}
\toprule
\textbf{Domain} & \textbf{Model} & \textbf{Success} & \textbf{Attempt} & \textbf{Null} \\
\midrule

\multirow{4}{*}{Agriculture}
& Gemini 2.0 Flash & 0.9760 & 0.9760 & 4 \\
& GPT-4o & 0.9940 & 0.9940 & 1 \\
& QwenVL & 0.9042 & 0.9880 & 2 \\
& QwenVL PAL & 0.9820 & 0.9940 & 1 \\
\midrule

\multirow{4}{*}{\shortstack{Border\\Crossing}}
& Gemini 2.0 Flash & 0.9783 & 0.9907 & 2 \\
& GPT-4o & 0.9907 & 0.9969 & 0 \\
& QwenVL & 0.7771 & 0.9783 & 7 \\
& QwenVL PAL & 0.9536 & 0.9938 & 2 \\
\midrule

\multirow{4}{*}{COVID}
& Gemini 2.0 Flash & 0.9667 & 0.9733 & 8 \\
& GPT-4o & 0.9900 & 0.9933 & 2 \\
& QwenVL & 0.5367 & 0.8933 & 32 \\
& QwenVL PAL & 0.9800 & 1.0000 & 0 \\
\midrule

\multirow{4}{*}{Crime}
& Gemini 2.0 Flash & 0.9935 & 0.9935 & 1 \\
& GPT-4o & 1.0000 & 1.0000 & 0 \\
& QwenVL & 0.7386 & 0.9935 & 1 \\
& QwenVL PAL & 0.9804 & 1.0000 & 0 \\
\midrule

\multirow{4}{*}{Demography}
& Gemini 2.0 Flash & 0.9667 & 0.9667 & 4 \\
& GPT-4o & 0.9917 & 0.9917 & 1 \\
& QwenVL & 0.6833 & 0.6917 & 37 \\
& QwenVL PAL & 0.9333 & 0.9750 & 3 \\
\midrule

\multirow{4}{*}{Diet}
& Gemini 2.0 Flash & 0.9810 & 0.9810 & 4 \\
& GPT-4o & 0.9810 & 0.9810 & 4 \\
& QwenVL & 0.8341 & 0.9479 & 11 \\
& QwenVL PAL & 0.9336 & 0.9953 & 1 \\
\midrule

\multirow{4}{*}{\shortstack{Road\\Injuries}}
& Gemini 2.0 Flash & 0.9737 & 0.9737 & 4 \\
& GPT-4o & 1.0000 & 1.0000 & 0 \\
& QwenVL & 0.8487 & 0.9079 & 14 \\
& QwenVL PAL & 0.9605 & 0.9934 & 1 \\
\midrule

\multirow{4}{*}{Walmart}
& Gemini 2.0 Flash & 0.9725 & 1.0000 & 0 \\
& GPT-4o & 1.0000 & 1.0000 & 0 \\
& QwenVL & 0.2569 & 0.9908 & 1 \\
& QwenVL PAL & 0.9908 & 1.0000 & 0 \\

\bottomrule
\end{tabular}
}
\label{tab:max_inference}
\end{table}

\begin{table}[ht]
\caption{Standard deviation inference performance across domains and models.}
\centering
\resizebox{\linewidth}{!}{
\begin{tabular}{l l c c c}
\toprule
\textbf{Domain} & \textbf{Model} & \textbf{Success} & \textbf{Attempt} & \textbf{Null} \\
\midrule

\multirow{4}{*}{Agriculture}
& Gemini 2.0 Flash & 0.0000 & 0.9940 & 167 \\
& GPT-4o & 0.0000 & 1.0000 & 167 \\
& QwenVL & 0.0000 & 0.7246 & 167 \\
& QwenVL PAL & 0.0000 & 0.9581 & 167 \\
\midrule

\multirow{4}{*}{\shortstack{Border\\Crossing}}
& Gemini 2.0 Flash & 0.4149 & 0.9412 & 18 \\
& GPT-4o & 0.4551 & 0.9907 & 2 \\
& QwenVL & 0.0248 & 0.9876 & 4 \\
& QwenVL PAL & 0.7678 & 0.9659 & 11 \\
\midrule

\multirow{4}{*}{COVID}
& Gemini 2.0 Flash & 0.3633 & 0.9700 & 9 \\
& GPT-4o & 0.3367 & 1.0000 & 0 \\
& QwenVL & 0.0100 & 0.9267 & 22 \\
& QwenVL PAL & 0.8000 & 0.9567 & 13 \\
\midrule

\multirow{4}{*}{Crime}
& Gemini 2.0 Flash & 0.4771 & 0.9935 & 1 \\
& GPT-4o & 0.4771 & 1.0000 & 0 \\
& QwenVL & 0.0458 & 0.9869 & 2 \\
& QwenVL PAL & 0.9608 & 0.9804 & 3 \\
\midrule

\multirow{4}{*}{Demography}
& Gemini 2.0 Flash & 0.5167 & 1.0000 & 0 \\
& GPT-4o & 0.2833 & 0.9917 & 1 \\
& QwenVL & 0.0417 & 0.8833 & 14 \\
& QwenVL PAL & 0.0333 & 0.9333 & 8 \\
\midrule

\multirow{4}{*}{Diet}
& Gemini 2.0 Flash & 0.0000 & 0.9716 & 211 \\
& GPT-4o & 0.0000 & 0.9953 & 211 \\
& QwenVL & 0.0000 & 0.7536 & 211 \\
& QwenVL PAL & 0.0000 & 0.9336 & 211 \\
\midrule

\multirow{4}{*}{\shortstack{Road\\Injuries}}
& Gemini 2.0 Flash & 0.2566 & 0.9605 & 6 \\
& GPT-4o & 0.1316 & 1.0000 & 0 \\
& QwenVL & 0.0395 & 0.9342 & 10 \\
& QwenVL PAL & 0.0197 & 0.9013 & 15 \\
\midrule

\multirow{4}{*}{Walmart}
& Gemini 2.0 Flash & 0.0000 & 0.9725 & 109 \\
& GPT-4o & 0.0000 & 0.9817 & 109 \\
& QwenVL & 0.0000 & 0.8073 & 109 \\
& QwenVL PAL & 0.0000 & 0.9908 & 109 \\

\bottomrule
\end{tabular}
}
\label{tab:std_inference}
\end{table}

\paragraph{Mean Inference.} Mean inference is the most stable and reliable numeric operation across all evaluated models. Strong models consistently achieve high success rates ($\geq 0.97$) with near-universal attempt rates ($\approx 1.00$) and negligible null counts across all settings. For example, \textit{Gemini2.0 Flash} and \textit{GPT-4o} attain success rates between $0.97$ and $1.00$ with zero or near-zero nulls throughout Table~\ref{tab:mean_inference}. Weaker models exhibit a distinct pattern: while attempt rates remain high (typically $0.90$–$0.98$), success drops substantially (e.g., $0.12$–$0.61$), accompanied by elevated null counts (often $>20$). This decoupling of attempt and success indicates that failures arise from imprecise aggregation rather than abstention. Importantly, mean inference performance does not correlate with time series length. Domains with long horizons show comparable success and null rates to short-horizon settings, confirming that temporal span is not a limiting factor.

\paragraph{Min Inference.} Minimum inference is substantially more fragile than mean inference. Although attempt rates remain high across models (typically $0.85$–$1.00$), success rates drop sharply for weaker models, and nulling increases significantly. For example, QwenVL exhibits success as low as $0.18$–$0.53$, with null counts exceeding $30$–$50$ in several settings (Table~\ref{tab:min_inference}). In contrast, stronger models maintain success rates above $0.95$ with minimal nulling. Crucially, this fragility does not align with time series length. Short sequences can yield failure rates comparable to or worse than long sequences. Qwen PAL variants consistently reduce nulling (e.g., null counts reduced by $2$–$4\times$ relative to base Qwen) and improve success, but still fail to match the robustness of stronger closed models.

\paragraph{Max Inference.} Maximum inference is the most robust extrema-related operation. Across nearly all models and settings, attempt rates are $\approx 1.00$, omission is minimal, and success rates are consistently high. Strong models achieve success rates of $0.97$–$1.00$, while even weaker models often exceed $0.73$–$0.90$ success (Table~\ref{tab:max_inference}). Notably, high success coincides with aggressive attempt behavior: failures are not driven by abstention. For instance, \textit{QwenVL} attempts maximum inference in $>97$ of cases but still exhibits reduced success in some settings, indicating semantic or conceptual errors rather than omission. Models exhibit a directional asymmetry in extrema reasoning: identifying peaks is substantially easier than identifying troughs, and high maximum scores are not artifacts of selective answering.

\paragraph{STD Inference.}
Standard deviation inference exposes a \textbf{fundamental capability gap} across all evaluated models. Unlike the other statistics, models frequently attempt STD inference (attempt rates often $0.90$–$1.00$), yet produce null or unscorable outputs in the majority of cases. This results in near-zero success across many settings. For example, several domains exhibit success $=0.00$ with null counts exceeding $100$–$200$ despite high attempt rates (Table~\ref{tab:std_inference}). This failure pattern is not driven by abstention, nor by sequence length. Even short-horizon settings show complete failure. PAL increases attempt rates but rarely improves success, demonstrating that the issue is not instruction following but the absence of internalized variance modeling. STD inference requires explicit reasoning about dispersion, which current models do not reliably internalize. Unlike extrema, variance cannot be approximated by semantic heuristics.

\section{Q\&A Tasks}

We present accuracy scores for VLMs on the Q\&A task in Table~\ref{tab:qa_eval}. An analysis of the highlighted statistics reveals a striking contrast between the finetuned and pretrained models. The finetuned model frequently produces highly confident yet sometimes incorrect predictions, whereas the pretrained model demonstrates more caution, acknowledging that the mean is lower than expected without attempting to estimate a specific value. 
Notably, certain proprietary models are now reaching, and at times even surpassing, human performance on specific subsets of tasks. While this signals exciting progress in the field, it also highlights the nuances of human cognitive performance, particularly under conditions where distraction might occur. It is vital to note, however, that no singular model has consistently achieved near-human proficiency across the entirety of the benchmark's demands. The plot retrieval task, in particular, stands out as a significant hurdle, robustly affirming the unparalleled human capacity for holistic visual-numeric integration, a critical frontier for time series understanding.

\begin{table}[ht]
\centering
\small
\setlength{\tabcolsep}{3pt}
\renewcommand{\arraystretch}{1.05}
\caption{Model accuracy for time series Q\&A tasks. \textbf{Bold} and \underline{underlined} scores denote first and second places (excluding human). Cap, Plt, and TS refer to caption, plot, and time series matching, whereas the comparative reasoning task is split into Amp, Peak, Mean, and Var, denoting amplitude, peak-earlier, mean, and variance comparison.}
\label{tab:qa_eval}

\resizebox{\columnwidth}{!}{
\begin{tabular}{l|ccc|cccc}
\toprule
\rowcolor[gray]{0.95}
\textbf{Model} & \textbf{Cap} & \textbf{Plt} & \textbf{TS} & \textbf{Amp} & \textbf{Peak} & \textbf{Mean} & \textbf{Var} \\
\midrule
\textbf{Proprietary} & & & & & & & \\
Gemini2 F
  & \underline{0.78} & 0.30 & \underline{0.61} & 0.80 & 0.42 & \underline{0.70} & 0.62 \\
Gemini2.5 
  & 0.66 & 0.30 & 0.31 & \textbf{1.00} & \textbf{1.00} & \textbf{1.00} & \textbf{0.85} \\
Claude3 H
  & 0.68 & 0.29 & 0.57 & 0.65 & 0.40 & 0.53 & 0.33 \\
GPT-4o
  & \textbf{0.96} & \underline{0.31} & \textbf{0.77} & 0.83 & 0.73 & \underline{0.70} & 0.63 \\
\midrule
\textbf{Pretrained} & & & & & & & \\
InternVL 2.5
  & 0.55 & 0.17 & 0.49 & 0.60 & 0.47 & 0.45 & 0.40 \\
LLaVA v1.6 
  & 0.39 & 0.27 & 0.32 & 0.45 & 0.45 & 0.42 & 0.45 \\
Phi-4 M.I.
  & 0.62 & 0.29 & 0.45 & 0.70 & 0.82 & 0.68 & \underline{0.70} \\
Idefics 2
  & 0.49 & 0.25 & 0.29 & 0.35 & 0.40 & 0.40 & 0.50 \\
SmolVLM
  & 0.26 & \textbf{0.34} & 0.28 & 0.40 & 0.48 & 0.44 & 0.60 \\
QwenVL
  & 0.68 & 0.27 & \underline{0.61} & 0.70 & 0.50 & 0.60 & 0.40 \\
Llama 3.2
  & 0.66 & 0.24 & 0.27 & 0.45 & 0.63 & 0.43 & 0.30 \\
\midrule
\textit{Human}
  & 0.81 & 0.95 & 0.83 & 0.93 & 0.85 & 0.95 & 0.90 \\
\bottomrule
\end{tabular}
}
\end{table}

\subsection{Qwen-Based Filtering} \label{qwen filter}
To show that questions erroneously answered by \texttt{Qwen 2.5 Omni} are indeed harder, we evaluated a subset of models on both an easy set of $600$ questions and the hard set generated by \texttt{Qwen 2.5 Omni}. The questions in the easy set are randomly sampled from those correctly answered by \texttt{Qwen 2.5 Omni}. Table~\ref{tab:qwen-hard-vs-easy} depicts the comparison. All models, regardless of architecture, show a consistent performance gain on the “easy” subset, demonstrating that Qwen-filtered questions are broadly harder, not uniquely harder for Qwen. To ensure a balanced benchmark, we release both the full set ($38.4$k questions) and the hard subset ($7$k questions), enabling evaluation and training across the entire difficulty spectrum.

\begin{table}[ht]
\centering
\small
\setlength{\tabcolsep}{6pt}
\renewcommand{\arraystretch}{1.1}
\caption{Performance on easy vs. hard questions and corresponding lift.}
\vspace{-0.5em}
\begin{tabular}{l|cc|c}
\toprule
\rowcolor[gray]{0.95}
\textbf{Model} & \textbf{Easy} & \textbf{Hard} & \textbf{Lift} \\
\midrule
Idefics 2      & 65\% & 46\% & +19\% \\
InternVL 2.5   & 72\% & 43\% & +29\% \\
Phi-4          & 61\% & 46\% & +15\% \\
SmolVLM        & 53\% & 48\% & +5\%  \\
Llama-3.2      & 54\% & 49\% & +5\%  \\
\bottomrule
\end{tabular}
\label{tab:qwen-hard-vs-easy}
\vspace{-0.5em}
\end{table}

\subsection{Distractor Generation in Q\&A Tasks} \label{artificial distractors}

To increase task difficulty, artificial perturbations are applied in the Time Series Matching and Caption Matching tasks. As shown in the table below, these perturbations significantly impacted model performance in Time Series Matching, increasing task difficulty and forcing models to reason over trends rather than relying on superficial cues. 

To illustrate the rationale of this design choice, assume the correct time series option is \texttt{[1,2,3]}. Having distractors like \texttt{[101, 99, 102]} makes the question trivial due to its totally different scale and nature. Our distractors are generated by perturbations on the correct time series, resulting in the following distractors.

\begin{enumerate}[leftmargin=*]
    \item \textbf{Shuffled}: \texttt{[2,1,3]}, avoids answering correctly by exploiting numeric lookup without temporal order awareness.
    \item \textbf{Reversed}: \texttt{[3,2,1]}, avoids reasoning without trend awareness.
    \item \textbf{Gaussian-noised}: \texttt{[1.03, 1.99, 3.002]}, forces precise numeric reasoning instead of superficial numeric and trend similarity.
\end{enumerate}

\begin{table}[H]
\centering
\small
\setlength{\tabcolsep}{6pt}
\renewcommand{\arraystretch}{1.1}
\caption{Qwen-2.5-Omni-7B accuracy by task and distractor type.}
\begin{tabular}{l|l|c}
\toprule
\rowcolor[gray]{0.95}
\textbf{Question Type} & \textbf{Distractor Type} & \textbf{Accuracy} \\
\midrule
TS Matching & Cross-domain             & 0.9803 \\
TS Matching & Same-domain              & 0.9586 \\
TS Matching & Artificially Perturbed   & \textbf{0.6864} \\
Caption Matching     & Cross-domain             & 0.8325 \\
Caption Matching     & Same-domain              & \textbf{0.8250} \\
Caption Matching     & Artificially Perturbed   & 0.8532 \\
\bottomrule
\end{tabular}
\label{tab:accuracy_results}
\end{table}

Table~\ref{tab:accuracy_results} shows that these perturbations indeed augment question difficulty. For Caption Matching, we do not find a significant difference between artificially perturbed distractors compared to original negative examples drawn from the dataset. We release the set of Q\&A questions without artificial perturbations as well. Perturbations are not applied to the other tasks; negative samples are instead drawn directly from other time series in the dataset.

\subsection{Temporal Matching Task}
\label{sec:temp-match-appendix}

Figure~\ref{fig:temporal_buckets} illustrates the temporal bucket structure used across our temporal matching diagnostic questions.

\begin{figure}[ht]
\centering
\fontsize{8pt}{9pt}\selectfont



\begin{tabular}{p{0.9\columnwidth} p{0.9\columnwidth} p{0.9\columnwidth}}

\begin{tcolorbox}[colback=white,colframe=black!40,arc=1mm,boxrule=0.3pt]

\textbf{Hourly Time Series }\\
Start time: \texttt{2022-07-12 22:00:00}\\
\small
[19.59, 19.48, 20.19, 18.95, 19.53, 19.41, 19.53, 18.54, 19.09, 16.71,
18.26, 19.19, 19.01, 19.38, 19.60, 19.63, 17.98, 17.71, 18.03, 18.43,
18.51, 18.30, 18.40, 18.16, 18.16, 18.42, 19.49, 17.91, 19.19, 18.20,
18.80, 18.20, 18.26, 17.87, 18.03, 18.29, 17.84, 18.73, 18.88, 18.27,
17.97, 18.09, 17.96, 17.37, 18.05, 18.42, 18.00, 18.73, 17.54, 18.56,
17.58, 17.73, 18.42, 18.02, 17.94, 17.75, 17.72, 19.24, 18.87, 18.74,
18.01, 17.88]
\end{tcolorbox}

\begin{tcolorbox}[colback=cyan!5,colframe=black!40,arc=1mm,boxrule=0.3pt]
\textbf{Early Bucket} \quad Query: \texttt{2022-07-13 15:00:00}

Options:

\begin{tabular}{@{}llll@{}}
A. 17.71 \quad &
B. 19.38 \quad &
C. 17.98 \quad &
D. 18.16
\end{tabular}

\textbf{Answer:} A
\end{tcolorbox}

\begin{tcolorbox}[colback=cyan!5,colframe=black!40,arc=1mm,boxrule=0.3pt]
\textbf{Mid-1 Bucket}\quad
Query: \texttt{2022-07-14 00:00:00}

Options:

\begin{tabular}{@{}llll@{}}
A. 18.80 \quad &
B. 18.26 \quad &
C. 18.40 \quad &
D. 19.49
\end{tabular}

\textbf{Answer:} D
\end{tcolorbox}

\begin{tcolorbox}[colback=cyan!5,colframe=black!40,arc=1mm,boxrule=0.3pt]
\textbf{Mid-2 Bucket}\quad
Query: \texttt{2022-07-14 04:00:00}

Options: \\
\begin{tabular}{@{}llll@{}}
A. 17.84 \quad &
B. 18.80 \quad &
C. 18.03 \quad &
D. 18.42
\end{tabular}

\textbf{Answer:} B
\end{tcolorbox}

\begin{tcolorbox}[colback=cyan!5,colframe=black!40,arc=1mm,boxrule=0.3pt]
\textbf{Mid-3 Bucket}\quad
Query: \texttt{2022-07-14 18:00:00}

Options:\\
\begin{tabular}{@{}llll@{}}
A. 18.56 \quad &
B. 18.42 \quad &
C. 18.05 \quad &
D. 18.73
\end{tabular}

\textbf{Answer:} C
\end{tcolorbox}

\begin{tcolorbox}[colback=cyan!5,colframe=black!40,arc=1mm,boxrule=0.3pt]
\textbf{Late Bucket}\quad
Query: \texttt{2022-07-15 04:00:00}

Options: \\
\begin{tabular}{@{}llll@{}}
A. 17.94 \quad &
B. 18.01 \quad &
C. 18.56 \quad &
D. 18.02
\end{tabular}

\textbf{Answer:} A
\end{tcolorbox}

&
&

\end{tabular}

\caption{
Illustration of temporal buckets used in our diagnostic MCQ task.
All questions query the same hourly time series, with increasing temporal
distance from the start timestamp, isolating temporal indexing difficulty.
}
\label{fig:temporal_buckets}
\end{figure}

\paragraph{Distractor Generation for Temporal Matching}
Distractor options are generated using a temporal locality--driven sampling strategy.
Given a time series $\mathbf{x} = \{x_1, \dots, x_N\}$ and a correct index $i^\ast$, we first collect candidate values from a symmetric temporal window centered at $i^\ast$.
If the number of unique candidates is insufficient, the window is expanded to a larger radius, with a final fallback to sampling from the entire series. This ensures distractors are temporally plausible while avoiding trivial elimination strategies (Algorithm~\ref{alg:tlds}).

\begin{algorithm}
\caption{Temporal Locality--Driven Distractor Sampling (TLDS)}
\label{alg:tlds}
\begin{algorithmic}[1]
\REQUIRE Time series $\mathbf{x}$, correct index $i^\ast$, window size $w$, maximum window $W$, number of options $K$
\ENSURE $K-1$ distractor values

\STATE Initialize empty candidate set $\mathcal{C}$

\FOR{$o = -w$ to $w$}
    \STATE $i \leftarrow i^\ast + o$
    \IF{$0 \le i < N$ \AND $i \neq i^\ast$}
        \IF{$x_i \neq x_{i^\ast}$}
            \STATE Add $x_i$ to $\mathcal{C}$
        \ENDIF
    \ENDIF
\ENDFOR

\IF{$|\mathcal{C}| < K-1$}
    \STATE Repeat the above procedure using window size $W$
\ENDIF

\IF{$|\mathcal{C}| < K-1$}
    \STATE $\mathcal{C} \leftarrow \{x_1, \dots, x_N\} \setminus \{x_{i^\ast}\}$
\ENDIF

\STATE Sample $K-1$ distinct values uniformly from $\mathcal{C}$
\RETURN sampled distractors
\end{algorithmic}
\end{algorithm}

\paragraph{Index Lookup Baseline.} We implement an index-based lookup baseline (Algorithm~\ref{alg:strict_lookup}) that deterministically retrieves the ground-truth value for a queried timestamp when exact temporal alignment is possible. Given a start time and query time, the baseline computes the hour offset using strict datetime parsing and requires the offset to be an exact integer. If the resulting index is within bounds, the corresponding time series value is retrieved and matched against the multiple-choice options using exact numeric equality, with no rounding or tolerance. The baseline returns an answer only when a unique exact match exists; otherwise, it abstains. This baseline serves as a correctness upper bound for the task and isolates model failures from ambiguities in data representation or parsing.

\begin{algorithm}
\caption{Temporal Index Lookup Baseline}
\label{alg:strict_lookup}
\begin{algorithmic}[1]
\REQUIRE Time series $\mathbf{x} = [x_0, \dots, x_{N-1}]$, start time $t_0$, query time $t_q$, option set $\mathcal{O}$
\ENSURE Predicted option label or failure

\STATE Parse $t_0$ and $t_q$ using ISO datetime parsing
\STATE Compute time offset $\Delta t \leftarrow t_q - t_0$
\STATE Convert offset to hours: $h \leftarrow \Delta t / 3600$

\IF{$h$ is not an integer}
    \RETURN failure
\ENDIF

\STATE $i \leftarrow h$

\IF{$i < 0$ OR $i \geq |\mathbf{x}|$}
    \RETURN failure
\ENDIF

\STATE $v^\ast \leftarrow x_i$
\STATE Parse option values using  numeric parsing
\STATE $\mathcal{M} \leftarrow \{ k \mid \mathcal{O}[k] = v^\ast \}$

\IF{$|\mathcal{M}| = 1$}
    \RETURN the unique label in $\mathcal{M}$
\ELSE
    \RETURN failure
\ENDIF

\end{algorithmic}
\end{algorithm}

\subsection{Sample Q\&A Questions \label{qa-samples}}
\begin{figure}[ht]
  \centering
  \begin{tcolorbox}[
      colback=white,
      colframe=black!40,
      boxrule=0.4pt,
      arc=1mm,
      left=3pt,
      right=3pt,
      top=3pt,
      bottom=2pt,
      width=0.48\textwidth
    ]
    \fontsize{8pt}{9pt}\selectfont
    \centering


    \vspace{0.6em}

    \begin{tcolorbox}[
        colback=cyan!5,
        colframe=cyan!40,
        arc=1mm,
        boxrule=0.3pt,
        width=\textwidth,
        top=2pt,
        bottom=2pt
      ]
      \textbf{Question}

      \vspace{0.3em}
      
      \includegraphics[width=0.5\textwidth]{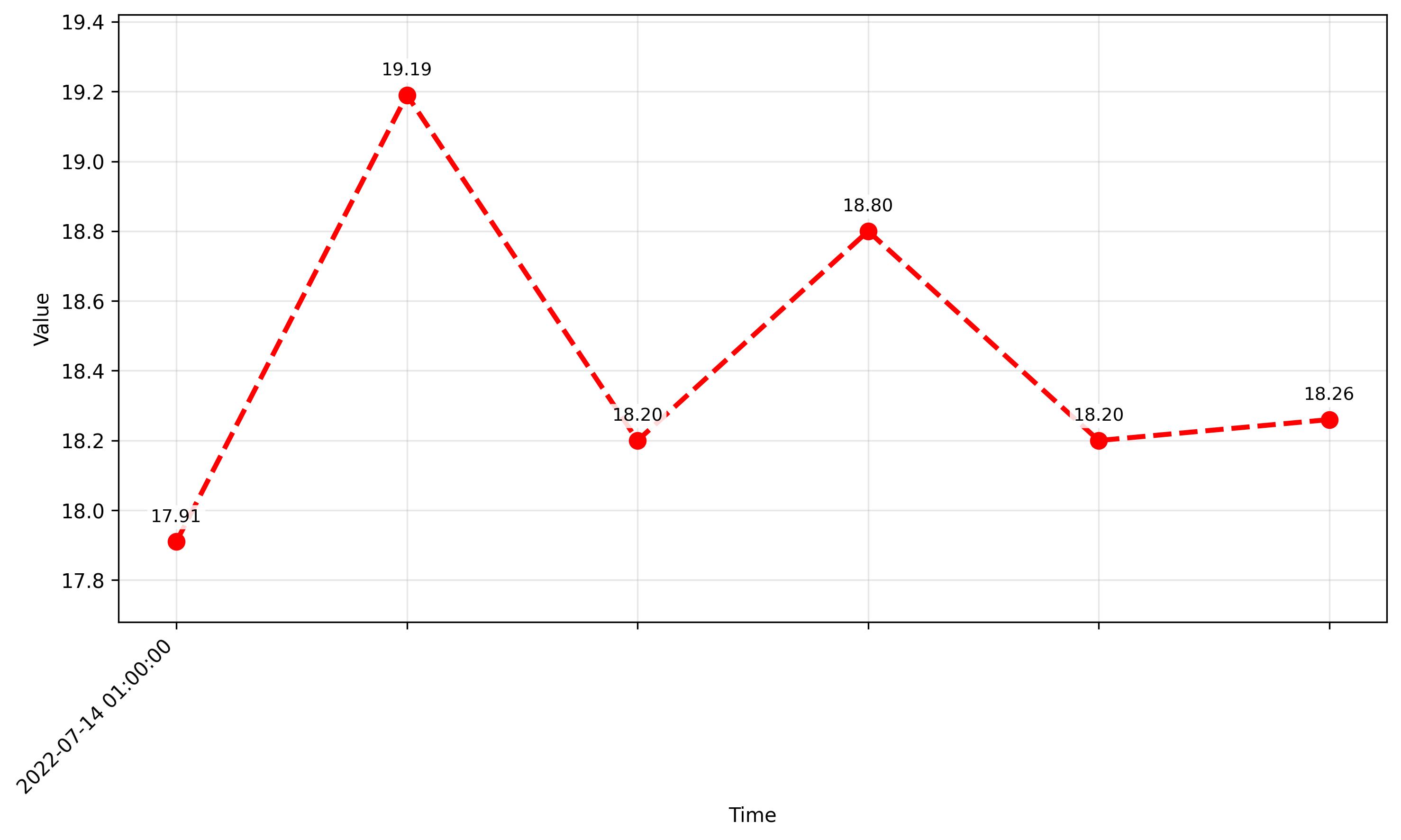}

      You are given an hourly time series shown as a plot. Only the starting time is labeled on the x-axis. Each point represents one hour.\\

      Start time:
      2022-07-14 01:00:00\\

      Question:
      What is the value of the time series at the following time?

      2022-07-14 04:00:00\\

        Options: \\
        A. 18.26 \quad
        B. 18.8 \quad
        C. 19.19 \quad
        D. 18.2\\

      Answer with exactly one capital letter from \{A, B, C, D\}. Do not explain anything. Just answer in one capital letter.
      Answer:
    \end{tcolorbox}

    \begin{tcolorbox}[
        colback=lime!5,
        colframe=lime!40!black,
        arc=1mm,
        boxrule=0.3pt,
        width=\textwidth,
        top=1pt,
        bottom=1pt
      ]
      \textbf{Answer}\\[0.3em]
      \texttt{B}
    \end{tcolorbox}

  \end{tcolorbox}
  \caption{Example of a \textit{plot-based temporal matching} question.}
  \label{fig:time-matching-plot}
\end{figure}

\begin{figure}[H]
  \centering
  \begin{tcolorbox}[
      colback=white,
      colframe=black!40,
      boxrule=0.4pt,
      arc=1mm,
      left=2pt,
      right=2pt,
      top=2pt,      
      bottom=1pt,   
      width=0.48\textwidth
    ]
    \fontsize{8pt}{9pt}\selectfont  
    \centering

    \begin{tcolorbox}[
        colback=cyan!5,
        colframe=cyan!40,
        arc=1mm,
        boxrule=0.3pt,
        width=\textwidth,
        valign=top,
        top=2pt,       
        bottom=2pt
      ]
      \textbf{Question}\\[0.3em]  
      Here is a time series:\\ 
      186.57, 186.57, 186.57, 186.57, 186.57, 150.29, 150.29, 150.29, 150.29, 150.29, 150.29, 150.29, 103.14, 103.14, 103.14, 103.14, 103.14, 103.14, 103.14, 77.00, 77.00, 77.00, 77.00, 77.00, 77.00, 77.00, 52.71, 52.71, 52.71, 52.71, 52.71, 52.71, 52.71, 41.71, 41.71, 41.71, 41.71, 41.71, 41.71, 41.71, 39.71, 39.71, 39.71, 39.71, 39.71, 39.71, 39.71, 29.86, 29.86, 29.86, 29.86, 29.86, 29.86, 29.86, 27.43, 27.43, 27.43, 27.43, 27.43, 27.43, 27.43, 22.57, 22.57, 22.57, 22.57, 22.57, 22.57, 22.57, 15.14, 15.14, 15.14, 15.14, 15.14, 15.14, 15.14, 18.71, 18.71, 18.71, 18.71, 18.71, 18.71, 18.71, 22.71, 22.71, 22.71, 22.71, 22.71, 22.71, 22.71, 23.14, 23.14, 23.14, 23.14, 23.14, 23.14, 23.14, 21.00, 21.00, 21.00, 21.00, 21.00, 21.00, 21.00, 30.57, 30.57, 30.57, 30.57, 30.57, 30.57, 30.57, 30.57, 30.57, 30.57, 30.57, 30.57, 30.57, 30.57, 36.29, 36.29, 36.29, 36.29, 36.29, 36.29, 36.29, 59.71, 59.71, 59.71, 59.71, 59.71, 59.71, 59.71, 93.71, 93.71, 93.71, 93.71, 93.71, 93.71, 93.71, 140.86, 140.86, 140.86, 140.86, 140.86, 140.86, 140.86\\
      
      Here are four plots of different time series:\\
      (A)~\includegraphics[width=0.5\textwidth]{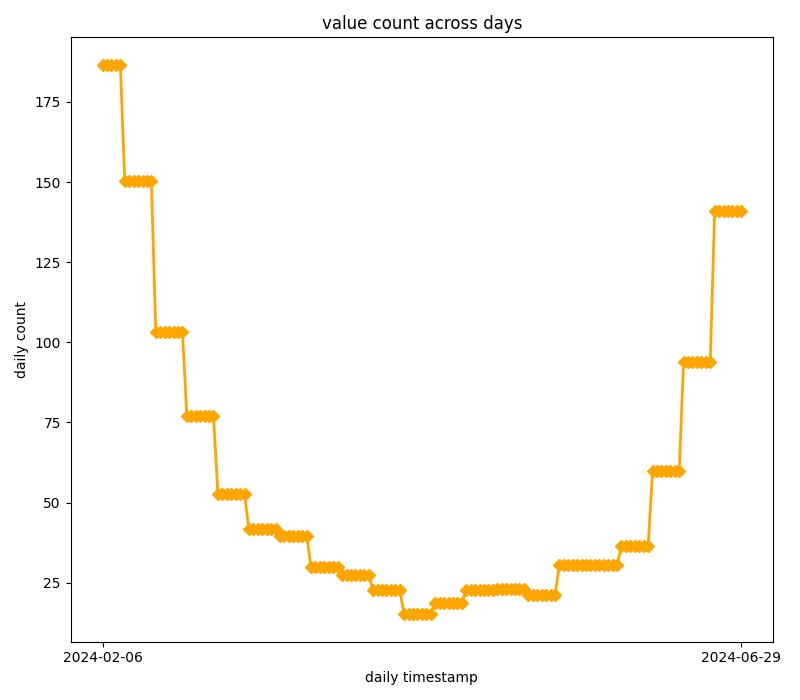} \\
      (B)~\includegraphics[width=0.6\textwidth]{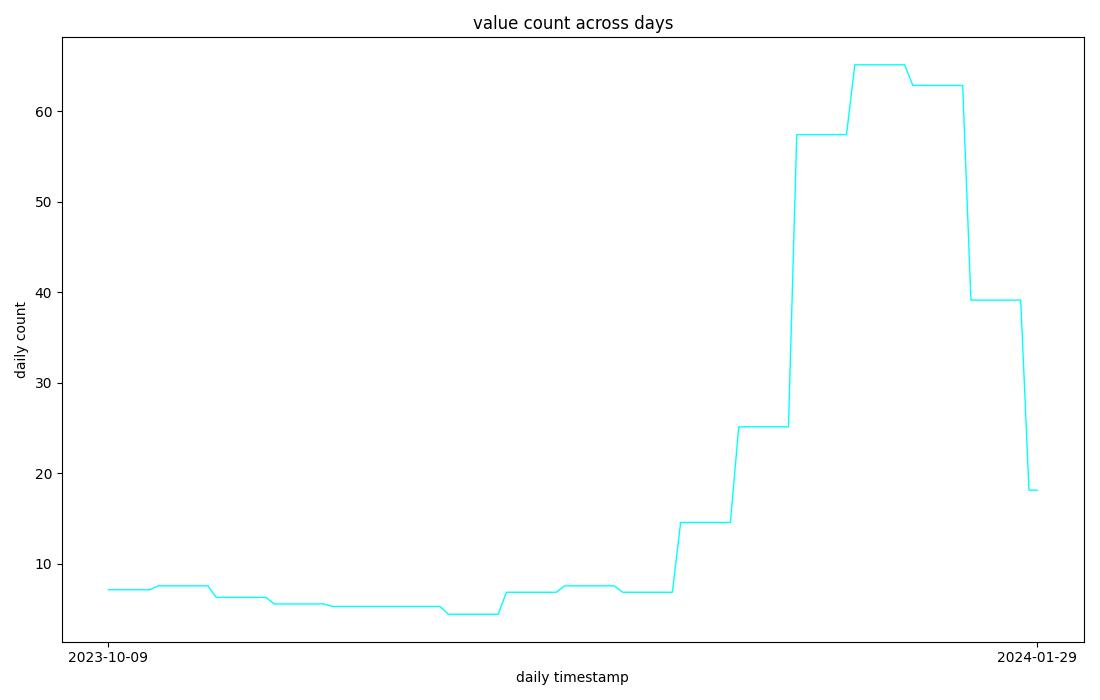} \\
      (C)~\includegraphics[width=0.6\textwidth]{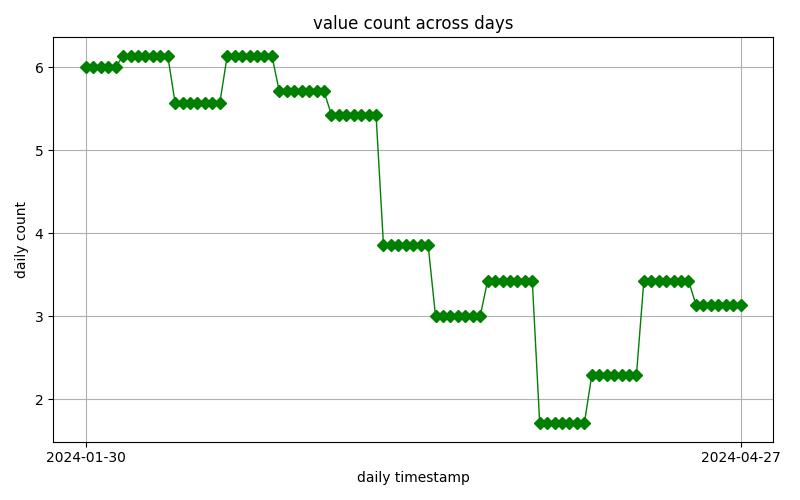} \\(D)~\includegraphics[width=0.6\textwidth]{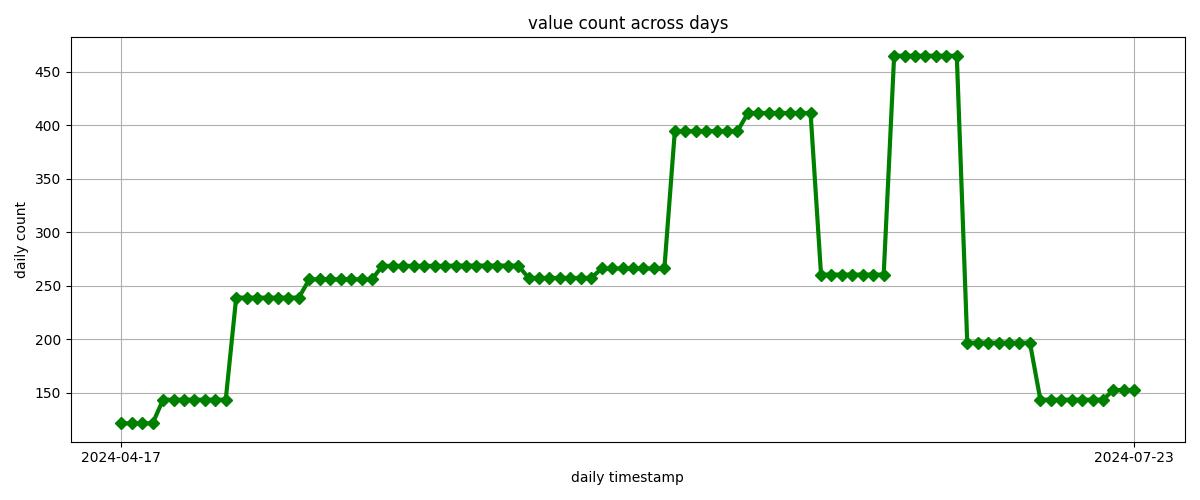} \\
      
      \vspace{0.3em}
      Which plot corresponds to the time series provided above?
      
      \vspace{0.3em}
      You must respond only with valid JSON, and no extra text or markdown.\\ \\
      The JSON schema is: \\
      \{\texttt{"answer": <string>}\} \\
      \texttt{<string>} must be an answer string containing only A, B, C, or D.\\
      Ensure your output parses as JSON with exactly one top-level object containing the answer field.
    \end{tcolorbox}

    \begin{tcolorbox}[
        colback=lime!5,
        colframe=lime!40!black,
        arc=1mm,
        boxrule=0.3pt,
        width=\textwidth,
        valign=center,
        top=1pt,
        bottom=1pt
      ]
      \textbf{Answer}\\[0.3em]
      \texttt{"answer": "A"}
    \end{tcolorbox}
  \end{tcolorbox}
  \caption{Example of a \textit{plot matching} question.}
  \label{fig:plot matching}
\end{figure}

We provide examples of Q\&A questions in Figures~\ref{fig:amplitude},~\ref{fig:peak},~\ref{fig:mean},~\ref{fig:variance},~\ref{fig:caption matching},~\ref{fig:plot matching}, ~\ref{fig:ts matching}, ~\ref{fig:time-matching}, and~\ref{fig:time-matching-plot}, covering one example per question type. 

\begin{figure}[H]
  \centering
  \begin{tcolorbox}[
      colback=white,
      colframe=black!40,
      boxrule=0.4pt,
      arc=1mm,
      left=2pt,
      right=2pt,
      top=2pt,      
      bottom=1pt,   
      width=0.48\textwidth
    ]
    \fontsize{8pt}{9pt}\selectfont  
    \centering

    \begin{tcolorbox}[
        colback=cyan!5,
        colframe=cyan!40,
        arc=1mm,
        boxrule=0.3pt,
        width=\textwidth,
        valign=top,
        top=2pt,       
        bottom=2pt
      ]
      \textbf{Question}\\[0.3em]  
      Given two time series A and B, detect which one has a higher amplitude defined as the maximum - minimum.
      \vspace{0.3em}
      
      A: [1.15, 0.92, 0.85, 0.75, 0.57, 0.62, 0.6, 0.5, 0.68, 0.72, 0.8, 0.67, 0.8, 0.55, 0.55, 0.7, 0.88]\\
      B: [87.0, 83.0, 77.0, 74.0, 84.0] \\
      
      \vspace{0.3em}
      You must respond only with valid JSON, and no extra text or markdown.\\ \\
      The JSON schema is: \\
      \{\texttt{"answer": <string>}\} \\
      \texttt{<string>} must be an answer string containing only A, B.\\
      Ensure your output parses as JSON with exactly one top-level object containing the answer field.
    \end{tcolorbox}

    \begin{tcolorbox}[
        colback=lime!5,
        colframe=lime!40!black,
        arc=1mm,
        boxrule=0.3pt,
        width=\textwidth,
        valign=center,
        top=1pt,
        bottom=1pt
      ]
      \textbf{Answer}\\[0.3em]
      \texttt{"answer": "B"}
    \end{tcolorbox}
  \end{tcolorbox}
  \caption{Example of a \textit{time series amplitude comparison} question.}
  \label{fig:amplitude}
\end{figure}

\begin{figure}[ht]
  \centering
  \begin{tcolorbox}[
      colback=white,
      colframe=black!40,
      boxrule=0.4pt,
      arc=1mm,
      left=2pt,
      right=2pt,
      top=2pt,      
      bottom=1pt,   
      width=0.48\textwidth
    ]
    \fontsize{8pt}{9pt}\selectfont  
    \centering

    \begin{tcolorbox}[
        colback=cyan!5,
        colframe=cyan!40,
        arc=1mm,
        boxrule=0.3pt,
        width=\textwidth,
        valign=top,
        top=2pt,       
        bottom=2pt
      ]
      \textbf{Question}\\[0.3em]  
      Given two time series A and B, detect which one reaches its maximum earlier.
      \vspace{0.3em}
      
      A: [66.76, 83.06, 85.77, 90.77, 98.81, 90.62, 80.05, 91.36, 89.59, 76.4, 80.1, 85.6, 84.41]\\
      B: [949.0, 689.0, 561.0, 552.0, 563.0] \\
      
      \vspace{0.3em}
      You must respond only with valid JSON, and no extra text or markdown.\\ \\
      The JSON schema is: \\
      \{\texttt{"answer": <string>}\} \\
      \texttt{<string>} must be an answer string containing only A, B.\\
      Ensure your output parses as JSON with exactly one top-level object containing the answer field.
    \end{tcolorbox}

    \begin{tcolorbox}[
        colback=lime!5,
        colframe=lime!40!black,
        arc=1mm,
        boxrule=0.3pt,
        width=\textwidth,
        valign=center,
        top=1pt,
        bottom=1pt
      ]
      \textbf{Answer}\\[0.3em]
      \texttt{"answer": "B"}
    \end{tcolorbox}
  \end{tcolorbox}
  \caption{Example of a \textit{time series peak comparison} question.}
  \label{fig:peak}
\end{figure}

\begin{figure}[H]
  \centering
  \begin{tcolorbox}[
      colback=white,
      colframe=black!40,
      boxrule=0.4pt,
      arc=1mm,
      left=2pt,
      right=2pt,
      top=2pt,      
      bottom=1pt,   
      width=0.48\textwidth
    ]
    \fontsize{8pt}{9pt}\selectfont  
    \centering

    \begin{tcolorbox}[
        colback=cyan!5,
        colframe=cyan!40,
        arc=1mm,
        boxrule=0.3pt,
        width=\textwidth,
        valign=top,
        top=2pt,       
        bottom=2pt
      ]
      \textbf{Question}\\[0.3em]  
      Here is a time series caption:\\
      From 2014 to 2019, Bulgaria's Agricultural output index (2015=100) generally increased, starting at 103.4 in 2014 and reaching a peak of 109.23 in 2019, with a slight dip to 100.0 in 2015. The average output index during this period was 105.4, notably lower than the historical mean of 126.73, suggesting a period of relatively lower agricultural productivity compared to Bulgaria's long-term performance. The increase from 2015 to 2019 indicates a moderate recovery and growth phase-t within this specific timeframe.

      \vspace{0.3em}  
      What time series is best described by this caption?\\
      (A)~[109.23, 107.24, 108.45, 104.1, 100.0, 103.4]\\
      (B)~[108.45, 100.0, 104.1, 107.24, 103.4, 109.23]\\
      (C)~[103.9, 99.8, 104.1, 109.2, 106.8, 109.23]\\
      (D)~[103.4, 100.0, 104.1, 108.45, 107.24, 109.23]
      
      \vspace{0.3em}
      You must respond only with valid JSON, and no extra text or markdown.\\ \\
      The JSON schema is: \\
      \{\texttt{"answer": <string>}\} \\
      \texttt{<string>} must be an answer string containing only A, B, C, or D.\\
      Ensure your output parses as JSON with exactly one top-level object containing the answer field.
    \end{tcolorbox}

    \begin{tcolorbox}[
        colback=lime!5,
        colframe=lime!40!black,
        arc=1mm,
        boxrule=0.3pt,
        width=\textwidth,
        valign=center,
        top=1pt,
        bottom=1pt
      ]
      \textbf{Answer}\\[0.3em]
      \texttt{"answer": "D"}
    \end{tcolorbox}
  \end{tcolorbox}
  \caption{Example of a \textit{time series matching} question.}
  \label{fig:ts matching}
\end{figure}

\begin{figure}[H]
  \centering
  \begin{tcolorbox}[
      colback=white,
      colframe=black!40,
      boxrule=0.4pt,
      arc=1mm,
      left=2pt,
      right=2pt,
      top=2pt,      
      bottom=1pt,   
      width=0.48\textwidth
    ]
    \fontsize{8pt}{9pt}\selectfont  
    \centering

    \begin{tcolorbox}[
        colback=cyan!5,
        colframe=cyan!40,
        arc=1mm,
        boxrule=0.3pt,
        width=\textwidth,
        valign=top,
        top=2pt,       
        bottom=2pt
      ]
      \textbf{Question}\\[0.3em]  
      Given the following two time series A and B, please identify which one has higher overall values.
  
      \vspace{0.3em}
      
      A: [65.0, 65.0, 64.0, 37.0, 55.0, 51.0]\\
      B: [6.29, 6.29, 6.29, 7.0, 7.0, 7.0, 7.0, 6.71, 6.71, 6.71, 6.71, 6.717, 7.57, 7.57, 7.14, 7.14, 7.14, 7.14, 7.43] \\
      
      \vspace{0.3em}
      You must respond only with valid JSON, and no extra text or markdown.\\ \\
      The JSON schema is: \\
      \{\texttt{"answer": <string>}\} \\
      \texttt{<string>} must be an answer string containing only A, B.\\
      Ensure your output parses as JSON with exactly one top-level object containing the answer field.
    \end{tcolorbox}

    \begin{tcolorbox}[
        colback=lime!5,
        colframe=lime!40!black,
        arc=1mm,
        boxrule=0.3pt,
        width=\textwidth,
        valign=center,
        top=1pt,
        bottom=1pt
      ]
      \textbf{Answer}\\[0.3em]
      \texttt{"answer": "A"}
    \end{tcolorbox}
  \end{tcolorbox}
  \caption{Example of a \textit{time series mean comparison} question.}
  \label{fig:mean}
\end{figure}

\begin{figure}[H]
  \centering
  \begin{tcolorbox}[
      colback=white,
      colframe=black!40,
      boxrule=0.4pt,
      arc=1mm,
      left=2pt,
      right=2pt,
      top=2pt,      
      bottom=1pt,   
      width=0.48\textwidth
    ]
    \fontsize{8pt}{9pt}\selectfont  
    \centering

    \begin{tcolorbox}[
        colback=cyan!5,
        colframe=cyan!40,
        arc=1mm,
        boxrule=0.3pt,
        width=\textwidth,
        valign=top,
        top=2pt,       
        bottom=2pt
      ]
      \textbf{Question}\\[0.3em]  
      Given the following two time series A and B, please identify which one has higher volatility. 
  
      \vspace{0.3em}
      
      A: [0.14, 0.14, 0.14, 0.29, 0.29, 0.29, 0.29, 0.29, 0.29, 0.29, 0.57, 0.57, 0.57, 0.57, 0.57, 0.57]\\
      B: [0.21, 0.33, 0.41, 0.39, 0.44, 0.35, 0.35, 0.43, 0.51, 0.65, 0.69, 0.74] \\
      
      \vspace{0.3em}
      You must respond only with valid JSON, and no extra text or markdown.\\ \\
      The JSON schema is: \\
      \{\texttt{"answer": <string>}\} \\
      \texttt{<string>} must be an answer string containing only A, B.\\
      Ensure your output parses as JSON with exactly one top-level object containing the answer field.
    \end{tcolorbox}

    \begin{tcolorbox}[
        colback=lime!5,
        colframe=lime!40!black,
        arc=1mm,
        boxrule=0.3pt,
        width=\textwidth,
        valign=center,
        top=1pt,
        bottom=1pt
      ]
      \textbf{Answer}\\[0.3em]
      \texttt{"answer": "A"}
    \end{tcolorbox}
  \end{tcolorbox}
  \caption{Example of a \textit{time series variance comparison} question.}
  \label{fig:variance}
\end{figure}

\begin{figure}[H]
  \centering
  \begin{tcolorbox}[
      colback=white,
      colframe=black!40,
      boxrule=0.4pt,
      arc=1mm,
      left=2pt,
      right=2pt,
      top=2pt,
      bottom=1pt,
      width=0.48\textwidth
    ]
    \fontsize{8pt}{9pt}\selectfont
    \centering

    \begin{tcolorbox}[
        colback=cyan!5,
        colframe=cyan!40,
        arc=1mm,
        boxrule=0.3pt,
        width=\textwidth,
        valign=top,
        top=2pt,
        bottom=2pt
      ]
      \textbf{Question}\\[0.3em]

You are given an hourly time series. \\ 
\par\noindent
Time series:
\par\noindent
[19.59, 19.48, 20.19, 18.95, 19.53, 19.41, 19.53, 18.54, 19.09, 16.71, 18.26, 19.19, 19.01, 19.38, 19.6, 19.63, 17.98, 17.71, 18.03, 18.43, 18.51, 18.3, 18.4, 18.16, 18.16, 18.42, 19.49, 17.91, 19.19, 18.2, 18.8, 18.2, 18.26, 17.87, 18.03, 18.29, 17.84, 18.73, 18.88, 18.27, 17.97, 18.09, 17.96, 17.37, 18.05, 18.42, 18.0, 18.73, 17.54, 18.56, 17.58, 17.73, 18.42, 18.02, 17.94, 17.75, 17.72, 19.24, 18.87, 18.74, 18.01, 17.88]\\
\par\noindent
The series starts at:
\par\noindent
2022-07-12 22:00:00\\

\par\noindent
Question:
\par\noindent
What is the value of the time series at the following time?

\par\noindent
2022-07-13 15:00:00\\

\par\noindent
Options: \\
A. 17.71 \quad
B. 19.38 \quad
C. 17.98 \quad
D. 18.16\\

\par\noindent
You MUST answer using exactly one capital letter.
\par\noindent
Final answer (one letter only):

    \end{tcolorbox}

    \begin{tcolorbox}[
        colback=lime!5,
        colframe=lime!40!black,
        arc=1mm,
        boxrule=0.3pt,
        width=\textwidth,
        valign=center,
        top=1pt,
        bottom=1pt
      ]
      \textbf{Answer}\\[0.3em]
      \texttt{A}
    \end{tcolorbox}

  \end{tcolorbox}
  \caption{Example of a \textit{time series temporal matching} question.}
  \label{fig:time-matching}
\end{figure}

\begin{figure}[H]
  \centering
  \begin{tcolorbox}[
      colback=white,
      colframe=black!40,
      boxrule=0.4pt,
      arc=1mm,
      left=2pt,
      right=2pt,
      top=2pt,      
      bottom=1pt,   
      width=0.48\textwidth
    ]
    \fontsize{8pt}{9pt}\selectfont  
    \centering

    \begin{tcolorbox}[
        colback=cyan!5,
        colframe=cyan!40,
        arc=1mm,
        boxrule=0.3pt,
        width=\textwidth,
        valign=top,
        top=2pt,       
        bottom=2pt
      ]
      \textbf{Question}\\[0.3em]  
      Here is a time series:\\
      37.00, 37.00, 37.00, 37.00, 37.00, 40.57, 40.57, 40.57, 40.57, 40.57, 40.57\\

      \vspace{0.3em}
      What caption best describes this time series?\\
      
      (A)~From May 1st to July 26th, 2024, the daily COVID-19 deaths in China show a fluctuating pattern, with values generally ranging between 0.29 and 2.86. There are periods of relative stability, such as the initial days of May with a consistent 0.86, interspersed with occasional spikes to 2.86, and dips to 0.29 towards the end of July. Compared to the general daily death statistics for China, where the mean is 73.0 and the maximum reaches 6812.0, this specific time series indicates a period of significantly lower daily deaths, suggesting a substantial improvement in the COVID-19 situation during this timeframe.\\ \\
      (B)~From October 24, 2023, to November 3, 2023, the daily COVID-19 cases in Luxembourg show a relatively stable pattern, beginning at 37.3 cases and rising to 40.57 cases by October 29, 2023, where it remains for the rest of the period. Compared to the country's general statistics, where the mean is 236, the daily cases during this period are significantly lower, suggesting a period of reduced viral transmission. This trend does not follow any expected seasonal patterns, as COVID-19 case numbers are known to fluctuate unpredictably.\\ \\
      (C)~From October 24, 2023, to November 3, 2023, the daily COVID-19 cases in Luxembourg show a relatively stable pattern, beginning at 37 cases and rising to 40.57 cases by October 29, 2023, where it remains for the rest of the period. Compared to the country's general statistics, where the mean is 236.0, the daily cases during this period are significantly lower, suggesting a period of reduced viral transmission. This trend does not follow any expected seasonal patterns, as COVID-19 case numbers are known to fluctuate unpredictably.\\ \\
      (D)~From October 24, 2023, to November 3, 2023, the daily COVID-19 cases in Luxembourg show a relatively unstable pattern, beginning at 37 cases and decreasing to 40.57 cases by October 29, 2023, where it remains for the rest of the period. Compared to the country's general statistics, where the mean is 236.0, the daily cases during this period are significantly lower, suggesting a period of reduced viral transmission. This trend does follow expected seasonal patterns, as COVID-19 case numbers are known to fluctuate unpredictably.
      
      \vspace{0.3em}
      You must respond only with valid JSON, and no extra text or markdown.\\ \\
      The JSON schema is: \\
      \{\texttt{"answer": <string>}\} \\
      \texttt{<string>} must be an answer string containing only A, B, C, or D.\\
      Ensure your output parses as JSON with exactly one top-level object containing the answer field.
    \end{tcolorbox}

    \begin{tcolorbox}[
        colback=lime!5,
        colframe=lime!40!black,
        arc=1mm,
        boxrule=0.3pt,
        width=\textwidth,
        valign=center,
        top=1pt,
        bottom=1pt
      ]
      \textbf{Answer}\\[0.3em]
      \texttt{"answer": "C"}
    \end{tcolorbox}
  \end{tcolorbox}
  \caption{Example of a \textit{caption matching} question.}
  \label{fig:caption matching}
\end{figure}

\section{CaTS-Bench Samples \label{example samples}}

\begin{figure*}[ht]
\centering
\begin{tcolorbox}[colback=white, colframe=black!40, boxrule=0.4pt, arc=1mm, left=3pt, right=3pt, top=3pt, bottom=3pt, width=\linewidth]
\begin{minipage}[t]{0.495\textwidth}
\begin{tcolorbox}[colback=blue!5, colframe=blue!40, arc=1mm, boxrule=0.3pt, height=4.5cm]
\textbf{Time Series Segment}\\[0.3em]
\tiny\ttfamily\justifying 7.0, 7.0, 7.0, 7.0, 7.0, 7.0, 7.0, 8.0, 25.5, 42.0, 
163.33, 258.0, 214.5, 322.5, 354.75, 402.0, 182.33, 141.25, 
69.25, 47.0, 12.5, 7.0, 7.0, 7.0, 7.0, 7.0, 7.0, 7.0, 
7.0, 7.0, 7.0, 11.25, 84.0, 194.5, 338.75, 374.75, 427.25, 
272.75, 332.67, 377.75, 232.33, 111.67, 113.25, 23.5, 10.0, 
7.0, 7.0, 7.0, 7.0, 7.0, 7.0, 7.0, 7.0, 7.0, 7.0, 
9.0, 72.25, 91.0, 54.5, 213.0, 299.25, 233.75, 259.0, 
390.75, 367.75, 285.25, 207.75, 63.25, 9.5, 7.0, 7.0, 
6.75, 6.75, 7.0, 7.0
\end{tcolorbox}
\end{minipage}
\hfill
\begin{minipage}[t]{0.495\textwidth}
\begin{tcolorbox}[colback=teal!5, colframe=teal!100, arc=1mm, boxrule=0.3pt, height=4.5cm]
\textbf{Metadata JSON}\\[0.3em]
\tiny\ttfamily\justifying
\{
    "all-time  maximum": 730.0,
    "all-time average value until today": 127.62,
    "all-time minimum": 0.0,
    "all-time standard deviation until today": 175.79,
    "average value in this time series": 105.78,
    "city": "Visakhapatnam",
    "maximum value in this time series": 427.25,
    "measure": "SR (W/mt2)",
    "minimum value in this time series": 6.75,
    "sampling frequency": "hourly",
    "standard deviation in this time series": 133.47,
    "start\_month": "July",
    "start\_year": 2017,
    "starting time": "2022-06-27 22:00:00",
    "state": "Andhra Pradesh",
    "station\_location": "GVM Corporation, Visakhapatnam "
\}
\end{tcolorbox}
\end{minipage}

\vspace{0.3em}

\begin{minipage}[t]{0.495\textwidth}
\begin{tcolorbox}[colback=orange!5, colframe=orange!100, arc=1mm, boxrule=0.3pt, height=4cm]
\textbf{Line Plot Image}\\[0.3em]
\includegraphics[width=0.9\linewidth,height=0.12\textheight]{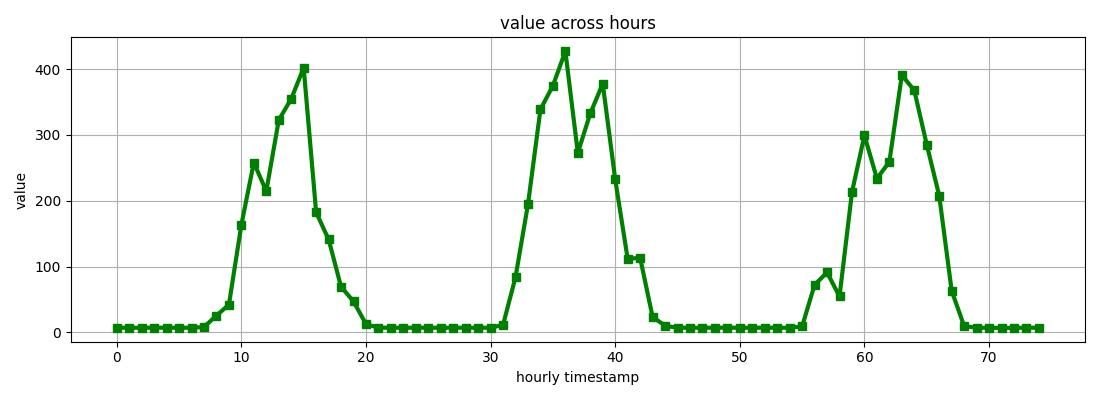}
\end{tcolorbox}
\end{minipage}
\hfill
\begin{minipage}[t]{0.495\textwidth}
\begin{tcolorbox}[colback=purple!5, colframe=purple!100, arc=1mm, boxrule=0.3pt, height=4cm]
\textbf{Caption}\\[0.3em]
\scriptsize
The hourly solar radiation (sr) data from Visakhapatnam, starting on June 27, 2022, exhibits a clear daily pattern of low values around 7 w/mt² during the night and early morning, sharply increasing to peaks during daylight hours, with a maximum value of 427.25 w/mt². compared to the city’s all-time average of 127.62 w/mt², the average value in this time series is 105.78 w/mt². The data follows a consistent diurnal cycle, with repeated peaks during the day and low values at night, showing a stable daily pattern.
\end{tcolorbox}
\end{minipage}
\end{tcolorbox}
\caption{Sample 1 showing time series data, metadata, plot image, and reference caption.}
\label{fig:sample}
\end{figure*}

\begin{figure*}[ht]
\centering
\begin{tcolorbox}[colback=white, colframe=black!40, boxrule=0.4pt, arc=1mm, left=2pt, right=2pt, top=3pt, bottom=2pt, width=\linewidth]

\begin{minipage}[t]{0.495\textwidth}
\begin{tcolorbox}[colback=blue!5, colframe=blue!100, arc=1mm, boxrule=0.3pt, height=4.5cm]
\textbf{Time Series Segment}\\[0.3em]
\scriptsize\ttfamily 57, 75, 100, 93, 48, 71, 38, 40, 51

\end{tcolorbox}
\end{minipage}
\hfill
\begin{minipage}[t]{0.495\textwidth}
\begin{tcolorbox}[colback=teal!5, colframe=teal!100, arc=1mm, boxrule=0.3pt, height=4.5cm]
\textbf{Metadata JSON}\\[0.3em]
\tiny\ttfamily
\{
    "border": "US-Canada Border",
    "end date of the series": "2020-02-01",
    "general maximum in the history of this port": 1922,
    "general mean in the history of this port": 70.79,
    "general minimum in the history of this port": 0,
    "general standard deviation in the history of this port": 108.73,
    "maximum in this specific series": 100,
    "mean of this specific series": 63.67,
    "means": "Trucks",
    "minimum in this specific series": 38,
    "port": "Del Bonita",
    "sampling frequency": "monthly",
    "standard deviation of this specific series": 21.17,
    "start date of the series": "2019-05-01",
    "state": "Montana"
\}
\end{tcolorbox}
\end{minipage}

\vspace{0.5em}

\begin{minipage}[t]{0.495\textwidth}
\begin{tcolorbox}[colback=orange!5, colframe=orange!100, arc=1mm, boxrule=0.3pt, height=4cm]
\textbf{Line Plot Image}\\[0.3em]
\includegraphics[width=\linewidth,height=0.12\textheight]{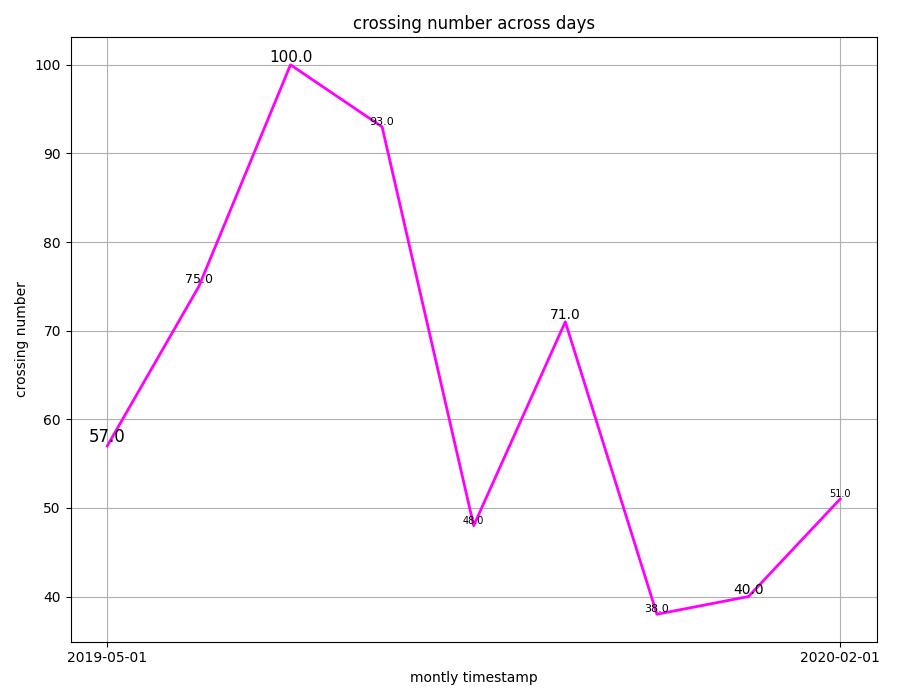}
\end{tcolorbox}
\end{minipage}
\hfill
\begin{minipage}[t]{0.495\textwidth}
\begin{tcolorbox}[colback=purple!5, colframe=purple!100, arc=1mm, boxrule=0.3pt, height=4cm]
\textbf{Caption}\\[0.3em]
\scriptsize
From May 2019 to February 2020, the Del Bonita port recorded noticeable variability in truck crossings. The series begins with 57 crossings in May 2019, rises to a maximum of 100 in July 2019, and declines to 38 in November 2019. The mean of 63.67 for this period remains below the historical average of 70.79, indicating slightly reduced throughput relative to the port’s long-term behavior. The standard deviation of 21.17, substantially lower than the historical variability of 108.73, reflects constrained fluctuation levels. Overall, this period represents a phase of subdued yet stable truck movement with limited deviations from month to month and a high peak at the start.
\end{tcolorbox}
\end{minipage}

\end{tcolorbox}
\caption{Sample 2 showing time series data, metadata, plot image, and reference caption.}
\label{fig:sample3}
\end{figure*}

\begin{figure*}[ht]
\centering
\begin{tcolorbox}[colback=white, colframe=black!40, boxrule=0.4pt, arc=1mm, left=3pt, right=3pt, top=3pt, bottom=3pt, width=\linewidth]

\begin{minipage}[t]{0.495\textwidth}
\begin{tcolorbox}[colback=blue!5, colframe=blue!40, arc=1mm, boxrule=0.3pt, height=4.5cm]
\textbf{Time Series Segment}\\[0.3em]
\scriptsize\ttfamily 0.52, 0.32, 0.30, 0.38, 0.41, 0.51, 0.43, 0.41, 0.47
\end{tcolorbox}
\end{minipage}
\hfill
\begin{minipage}[t]{0.495\textwidth}
\begin{tcolorbox}[colback=teal!5, colframe=teal!100, arc=1mm, boxrule=0.3pt, height=4.5cm]
\textbf{Metadata JSON}\\[0.3em]
\tiny\ttfamily
\{
    "attribute": "co2\_emissions",
    "country": "Djibouti",
    "end year of this series": 2018,
    "maximum of this specific series": 0.52,
    "mean of this specific series": 0.42,
    "minimum of this specific series": 0.30,
    "population at the end year": 1071886.0,
    "population at the start year": 930251.0,
    "region": "Middle East \& North Africa",
    "sampling frequency": "yearly",
    "standard deviation of this specific series": 0.07,
    "start year of this series": 2010
\}
\end{tcolorbox}
\end{minipage}

\vspace{0.5em}

\begin{minipage}[t]{0.495\textwidth}
\begin{tcolorbox}[colback=orange!5, colframe=orange!100, arc=1mm, boxrule=0.3pt, height=3.8cm]
\textbf{Line Plot Image}\\[0.3em]
\includegraphics[width=0.9\linewidth,height=0.12\textheight]{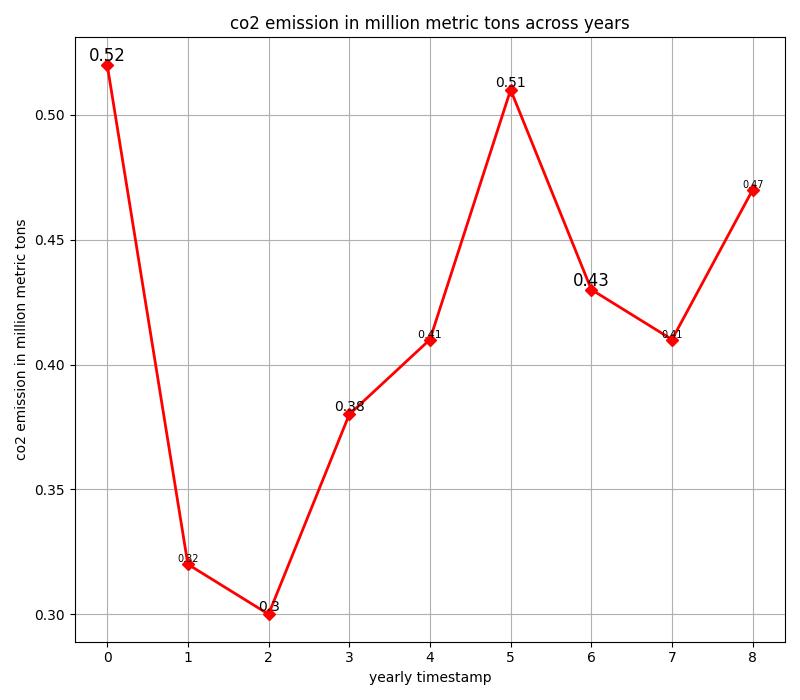}
\end{tcolorbox}
\end{minipage}
\hfill
\begin{minipage}[t]{0.495\textwidth}
\begin{tcolorbox}[colback=purple!5, colframe=purple!100, arc=1mm, boxrule=0.3pt, height=3.8cm]
\textbf{Caption}\\[0.3em]
\scriptsize
Djibouti's CO2 emissions from 2010 to 2018 show fluctuations around the average of 0.42 million metric tons, with a standard deviation of 0.07. Starting at 0.52 million metric tons in 2010, emissions generally decreased to a low of 0.3 million metric tons by 2012, before experiencing some increases and decreases, ending the period at 0.47 million metric tons in 2018.
\end{tcolorbox}
\end{minipage}

\end{tcolorbox}
\caption{Sample 3 showing time series data, metadata, plot image, and reference caption.}
\label{fig:sample5}
\end{figure*}

\begin{figure*}[ht]
\centering
\begin{tcolorbox}[colback=white, colframe=black!40, boxrule=0.4pt, arc=1mm, left=2pt, right=2pt, top=3pt, bottom=2pt, width=\linewidth]

\begin{minipage}[t]{0.495\textwidth}
\begin{tcolorbox}[colback=blue!5, colframe=blue!100, arc=1mm, boxrule=0.3pt, height=4.7cm]
\textbf{Time Series Segment}\\[0.3em]
\scriptsize\ttfamily 
  90.0, 90.0, 104.0, 104.0, 104.0, 104.0, 104.0, 104.0, 104.0, 110.57, 110.57, 110.57, 110.57, 110.57, 110.57, 110.57, 121.29, 121.29, 121.29, ..., 268.86, 268.86, 257.29, 257.29, 257.29, 257.29, 257.29, 257.29, 257.29, 266.14, 266.14, 266.14, 266.14, 266.14, 266.14, 266.14, 394.57, 394.57, 394.57, 394.57, 394.57, 394.57, 394.57, 411.57, 411.57, 411.57, 411.57, ..., 196.71, 196.71, 196.71, 143.43, 143.43, 143.43, 143.43, 143.43, 143.43, 143.43, 152.43
\end{tcolorbox}
\end{minipage}
\hfill
\begin{minipage}[t]{0.495\textwidth}
\begin{tcolorbox}[colback=teal!5, colframe=teal!100, arc=1mm, boxrule=0.3pt, height=4.7cm]
\textbf{Metadata JSON}\\[0.3em]
\tiny\ttfamily
\{
    "attribute": "cases",
    "country": "Thailand",
    "end date of this series": "2024-07-21",
    "historical maximum in this country": 26073.0,
    "historical mean in this country": 2877.0,
    "historical minimum in this country": 0.0,
    "historical standard deviation in this country": 5778.0,
    "income group": "Low \& Middle Income",
    "maximum of this specific series": 465.14,
    "mean of this specific series": 240.7,
    "minimum of this specific series": 90.0,
    "population": 71697024,
    "region": "East Asia \& Pacific",
    "sampling frequency": "daily",
    "standard deviation of this specific series": 106.03,
    "start date of this series": "2024-03-29"
\}
\end{tcolorbox}
\end{minipage}

\vspace{0.5em}

\begin{minipage}[t]{0.495\textwidth}
\begin{tcolorbox}[colback=orange!5, colframe=orange!100, arc=1mm, boxrule=0.3pt, height=4cm]
\textbf{Line Plot Image}\\[0.3em]
\includegraphics[width=\linewidth,height=0.12\textheight]{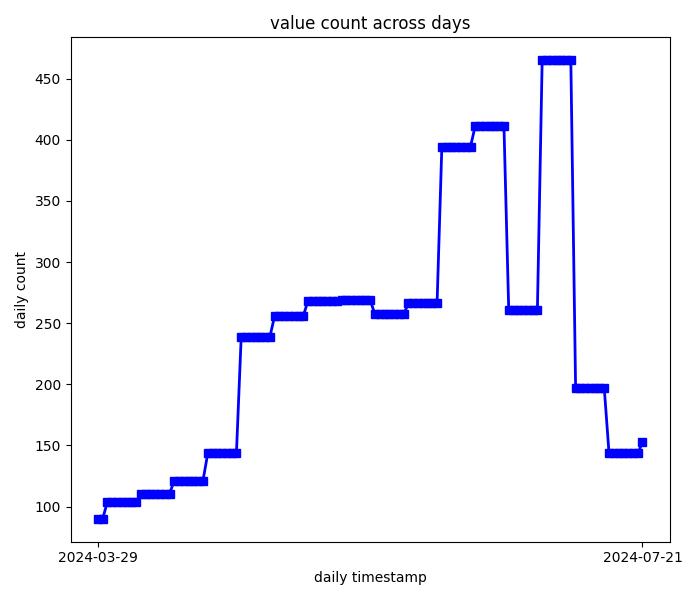}
\end{tcolorbox}
\end{minipage}
\hfill
\begin{minipage}[t]{0.495\textwidth}
\begin{tcolorbox}[colback=purple!5, colframe=purple!100, arc=1mm, boxrule=0.3pt, height=4cm]
\textbf{Caption}\\[0.3em]
\scriptsize
From March 29th to July 21st, 2024, the daily COVID-19 cases in Thailand fluctuated, starting at 90.0 and peaking at 465.14 around mid-June. This peak is significantly lower than the general maximum of 26073.0 daily cases observed in Thailand, indicating a substantial decrease in case numbers during this period. The time series shows an overall pattern of initial stability, followed by increases and decreases, with a final value of 152.43, suggesting a moderate decline towards the end of the observed period.
\end{tcolorbox}
\end{minipage}

\end{tcolorbox}
\caption{Sample 4 showing time series data, metadata, plot image, and reference caption.}
\label{fig:sample1}
\end{figure*}

\begin{figure*}[ht]
\centering
\begin{tcolorbox}[colback=white, colframe=black!40, boxrule=0.4pt, arc=1mm, left=2pt, right=2pt, top=3pt, bottom=2pt, width=\linewidth]

\begin{minipage}[t]{0.495\textwidth}
\begin{tcolorbox}[colback=blue!5, colframe=blue!100, arc=1mm, boxrule=0.3pt, height=3.8cm]
\textbf{Time Series Segment}\\[0.3em]
\scriptsize\ttfamily 83.65, 81.28, 86.98, 93.55, 96.89, 100.21, 101.88, 100.0, 103.96, 109.67

\end{tcolorbox}
\end{minipage}
\hfill
\begin{minipage}[t]{0.495\textwidth}
\begin{tcolorbox}[colback=teal!5, colframe=teal!100, arc=1mm, boxrule=0.3pt, height=3.8cm]
\textbf{Metadata JSON}\\[0.3em]
\tiny\ttfamily
\{
    "attribute": "Total Factor Productivity index (2015=100)",
    "country": "India",
    "end year of this series": 2017,
    "historical max": 115.04,
    "historical mean": 68.79,
    "historical min": 46.19,
    "maximum of this specific series": 109.67,
    "mean of this specific series": 95.81,
    "metrics info": "The Total Factor Productivity index (2015=100) is computed as the Output quantity divided by aggregated inputs.",
    "minimum of this specific series": 81.28,
    "sampling frequency": "yearly",
    "start year of this series": 2008
\}
\end{tcolorbox}
\end{minipage}

\vspace{0.5em}

\begin{minipage}[t]{0.495\textwidth}
\begin{tcolorbox}[colback=orange!5, colframe=orange!100, arc=1mm, boxrule=0.3pt, height=4cm]
\textbf{Line Plot Image}\\[0.3em]
\includegraphics[width=\linewidth,height=0.12\textheight]{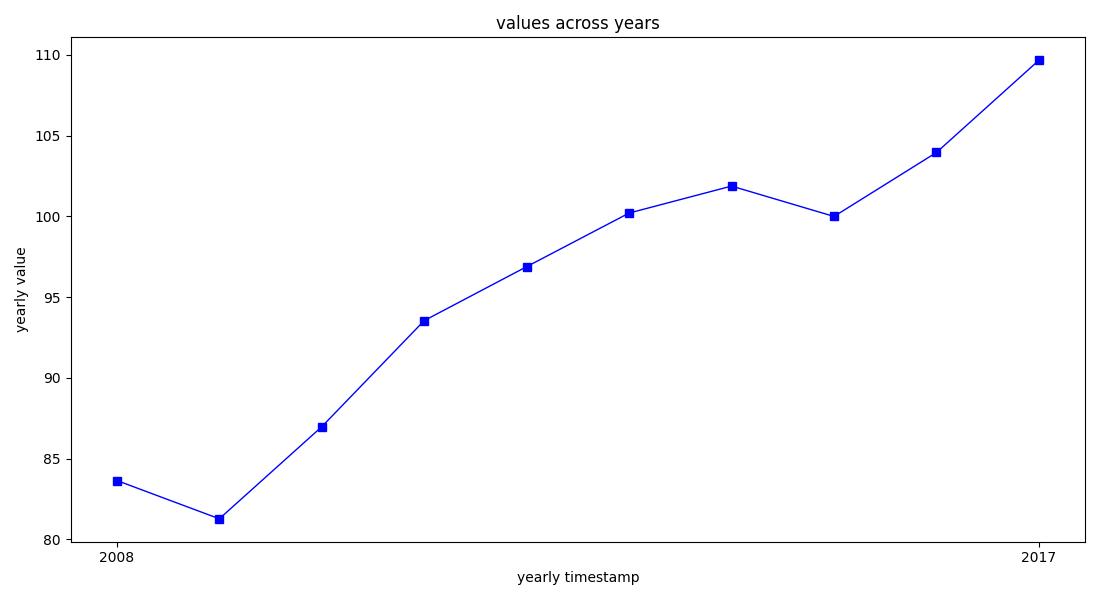}
\end{tcolorbox}
\end{minipage}
\hfill
\begin{minipage}[t]{0.495\textwidth}
\begin{tcolorbox}[colback=purple!5, colframe=purple!100, arc=1mm, boxrule=0.3pt, height=4cm]
\textbf{Caption}\\[0.3em]
\scriptsize
India's Total Factor Productivity index demonstrates consistent upward momentum from 2008 to 2017. Starting at 83.65 in 2008, the index experiences an initial dip to 81.28 in 2009 before steadily climbing to a peak of 109.67 in 2017, a 35\% increase from the initial value. This growth trajectory, with a mean of 95.81, suggests a period of increasing efficiency in India's production processes during this decade, contrasting with the lower historical mean of 68.79 and remaining well above the historical minimum of 46.19. The overall pattern reveals sustained growth with minimal volatility following the initial 2009 recovery.
\end{tcolorbox}
\end{minipage}

\end{tcolorbox}
\caption{Sample 5 showing time series data, metadata, plot image, and reference caption.}
\label{fig:sample2}
\end{figure*}

Representative CaTS-Bench samples are shown in Figures~\ref{fig:sample}, \ref{fig:sample3}, \ref{fig:sample5}, \ref{fig:sample1}, and \ref{fig:sample2}.

\end{document}